\documentclass[10pt,twocolumn,letterpaper]{article}

\usepackage{cvpr}              

\usepackage{graphicx}
\usepackage{comment}
\usepackage{amsmath}
\usepackage{amssymb}
\usepackage{amsfonts}
\usepackage{pifont}
\usepackage{booktabs}
\usepackage{algorithmic}
\usepackage{textcomp}
\usepackage{cite}
\usepackage{xcolor,soul}
\usepackage[normalem]{ulem}
\usepackage[cjk]{kotex}
\usepackage{multirow, array}
\usepackage{url}
\usepackage{footnote}
\makesavenoteenv{tabular}
\usepackage{relsize}
\usepackage[linesnumberedhidden,ruled,vlined]{algorithm2e}
\usepackage{adjustbox}
\usepackage{caption}
\usepackage{stfloats}
\usepackage[accsupp]{axessibility}
\usepackage[symbol]{footmisc}

\usepackage[pagebackref,breaklinks,colorlinks]{hyperref}

\usepackage[capitalize]{cleveref}
\crefname{section}{Sec.}{Secs.}
\Crefname{section}{Section}{Sections}
\Crefname{table}{Table}{Tables}
\crefname{table}{Tab.}{Tabs.}

\def\cvprPaperID{*****} 
\def\confName{CVPR}
\def\confYear{2023}

\begin{document}

\title{Balanced Energy Regularization Loss \\ for Out-of-distribution Detection 
}

\author{Hyunjun Choi$^1,^{2\textsuperscript{*}}$ \quad
Hawook Jeong$^2$ \quad
Jin Young Choi$^1$\\
{\small$^1$ ASRI, ECE., Seoul National University} \quad
{\small$^2$ RideFlux Inc.} \\
{\tt\footnotesize numb7315@snu.ac.kr} \quad
{\tt\footnotesize hawook@rideflux.com} \quad
{\tt\footnotesize jychoi@snu.ac.kr}
}
\maketitle

\begin{abstract}
In the field of out-of-distribution (OOD) detection, a previous method that use auxiliary data as OOD data has shown promising performance.
However, the method provides an equal loss to all auxiliary data to differentiate them from inliers.
However, based on our observation, in various tasks, there is a general imbalance in the distribution of the auxiliary OOD data across classes.
We propose a balanced energy regularization loss that is simple but generally effective for a variety of tasks.
Our balanced energy regularization loss utilizes class-wise different prior probabilities for auxiliary data to address the class imbalance in OOD data.
The main concept is to regularize auxiliary samples from majority classes, more heavily than those from minority classes.
Our approach performs better for OOD detection in semantic segmentation, long-tailed image classification, and image classification than the prior energy regularization loss.
Furthermore, our approach achieves state-of-the-art performance in two tasks: OOD detection in semantic segmentation and long-tailed image classification. Code is available at \url{https://github.com/hyunjunChhoi/Balanced_Energy}
\end{abstract}
\footnotetext{\textsuperscript{*}Work done as an intern at RideFlux.}
\vspace{-0.4cm}
\section{Introduction}
\label{sec:intro}

\vspace{-0.1cm}
\label{section_introduction}

\begin{figure*}[ht!]
\captionsetup{font=footnotesize}
\captionsetup[subfloat]{position=top}   
\centering
\subfloat[]{\includegraphics[width=1.4\columnwidth]{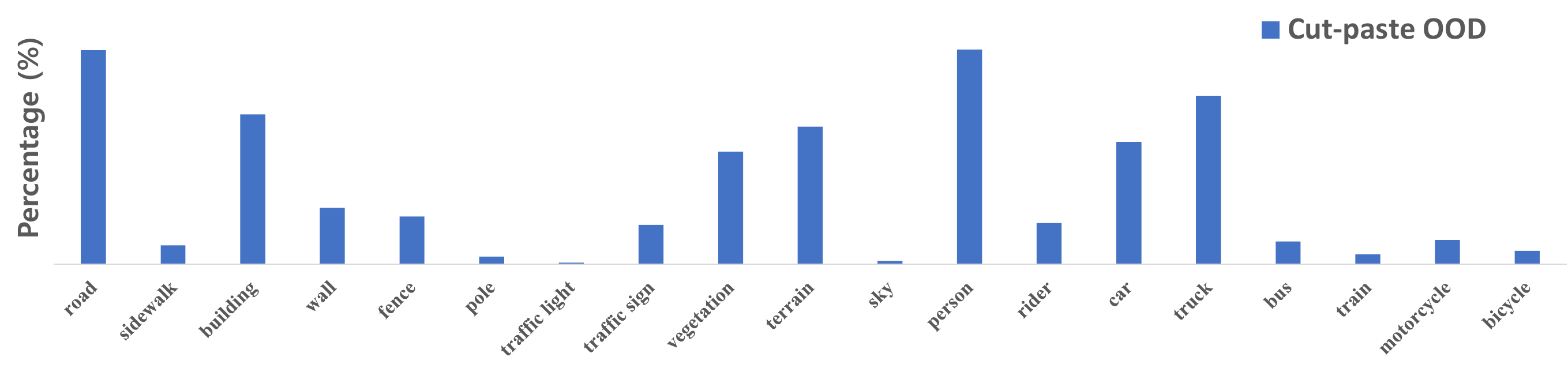}
\label{intro_fig_top}

}

\vfill
\vspace{-0.2cm}
\centering
\subfloat[]{\includegraphics[width=1.8\columnwidth]{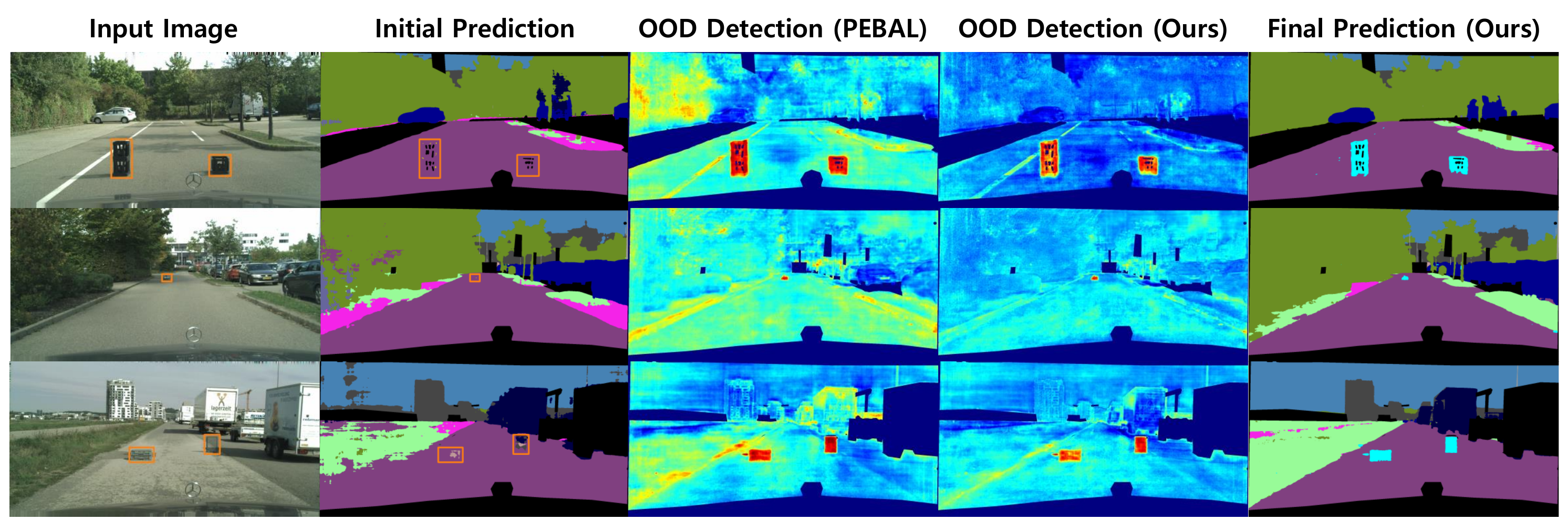}

\label{intro_fig_down}

}
\vspace{-0.2cm}
\caption{Overview of our approach in semantic segmentation task (a): Class distribution of cut-pasted OOD pixels collected from 10000 synthesized scene images ; (b): OOD detection result in Fishyscapes validation sets. 
Our balanced energy PEBAL(Ours) is the method that substitutes the energy regularization loss in PEBAL~\cite{tian2021pixel} with our balanced energy regularization loss.}
\end{figure*}

    Deep neural networks are used in a variety of fields such as image classification~\cite{krizhevsky2017imagenet} and semantic segmentation~\cite{farabet2012learning}.
    However, there is a challenge in the practical use of deep neural networks in areas where safety is crucial, such as autonomous driving and medical diagnosis~\cite{kendall2017uncertainties,leibig2017leveraging}.
    In particular, deep neural networks have the issue of providing high confidence to out-of-distribution (OOD) samples that are not used for training~\cite{hein2019relu}.
    As a result, Maximum softmax probability (MSP) score has been proposed to identify these OOD samples~\cite{hendrycks2016baseline}.
    Based on the score, OOD detection performance is evaluated by metrics (e.g. AUROC, FPR).
    Both in image classification~\cite{liang2017enhancing,lee2018simple,sastry2020detecting,hendrycks2018deep,liu2020energy,wang2022partial,yang2021semantically,liu2019large,papadopoulos2021outlier}(including long-tailed image classification~\cite{wang2022partial}) and semantic segmentation~\cite{hendrycks2019scaling,blum2021fishyscapes,bevandic2018discriminative,bevandic2019simultaneous,lis2019detecting,mukhoti2018evaluating,malinin2018predictive,di2021pixel,jung2021standardized,tian2021pixel,grcic2022densehybrid,chan2021entropy}, different approaches have been suggested to enhance the OOD detection performance. 
    Among them, we concentrate on the methods using auxiliary data as OOD data which indicate superior OOD detection performance to the previous methods that only use in-distribution samples.
   
    Outlier Exposure (OE) utilizes an auxiliary dataset of outliers to improve OOD detection performance~\cite{hendrycks2018deep}. The auxiliary data is consist of classes that do not overlap with the in-distribution data and the test OOD data.
    OE leverages the cross-entropy loss for the existing training data and the regularization loss for the auxiliary data.  
    The cross-entropy loss that results from giving the auxiliary data a uniform label is the regularization loss of OE.
    Meanwhile, a new energy score has been introduced in Energy-based OOD detection (EnergyOE) which replaces the MSP score~\cite{liu2020energy}.
    Furthermore, EnergyOE suggests an energy regularization loss that differs from that of OE to enhance performance.
    The squared hinge loss for energy with every existing (in-distribution) piece of data and every auxiliary (OOD) piece of data is added to create the energy regularization loss.
    Similarly, in semantic segmentation, the OOD detection performance is enhanced by using the auxiliary dataset of the outlier.   
    Meta-OOD~\cite{chan2021entropy} organized the auxiliary dataset of the outlier by scenes of the COCO dataset~\cite{lin2014microsoft}.
    Although the process of creating the auxiliary data is different from image classification, the training loss is comparable.
    Meta-OOD adopts the regularization loss proposed by OE.
    Recently, PEBAL~\cite{tian2021pixel} also adopts energy regularization loss proposed by EnergyOE. 
    
    However, when regularizing auxiliary data, the existing methods for OOD detection do not take into account variations between auxiliary data samples. The variations are severe especially on real data such as semantic segmentation for autonomous driving.
    As seen in Figure~\ref{intro_fig_top}, for the pre-trained model, the class distribution of the auxiliary OOD data  is not uniform across classes, i.e., imbalanced.
    To address the imbalanced problem, we regularize the auxiliary data differently for each sample.
    To achieve this, we propose a balanced energy regularization loss to apply higher regularization to majority classes than minority classes  
    in auxiliary data.
    In other words, auxiliary samples of majority classes receive a larger energy constraint than samples of minority classes. 
    We introduce the term $Z$, which indicates whether a sample belongs to the majority or minority of a class.
    $Z$ is the weighted sum of the softmax output of the classification model for a sample (i.e., the posterior probability of a class for a given sample), where the weight is the prior probability for the class.
    Unlike the existing energy regularization loss, our balanced energy regularization loss adjusts to the value of 
    $Z$ for an auxiliary data sample.
    Two adaptive loss components make up our loss: loss margin and loss weight.
    The adaptive loss margin provides an additional $Z$-proportional margin in the squared hinge loss for auxiliary data.
    The adaptive loss weight gives a weight proportional to $Z$ to the squared hinge loss.

    We confirm our novel loss in three tasks: semantic segmentation, long-tailed image classification, and image classification.
    The proposed loss is simple but generally effective for various tasks.
    Figure~\ref{intro_fig_down} illustrates how our method outperforms the previous state-of-the-art (SOTA) algorithm PEBAL in the semantic segmentation task by replacing the energy regularization loss with our loss.
    OOD detection performance is also enhanced when using our loss compared to the baseline (EnergyOE) which use only energy regularization loss.
    In all image classification tasks, we evaluate our method on semantically coherent OOD detection (SC-OOD) benchmark~\cite{yang2021semantically}.
    In long-tailed image classification task, our approach reveals superior OOD performance compared to both OE and EnergyOE methods which use auxiliary data.
    In addition, our approach outperforms the previous SOTA method PASCL~\cite{wang2022partial}, 
    Similarly, in the image classification task, we demonstrate the superiority of our loss by outperforming both OE and EnergyOE, which make use of auxiliary data. The contributions are summarized as:
    \vspace{-0.2cm}
    \begin{itemize}
    \item By making inferences based on previously trained models, we explain the imbalanced distribution of auxiliary OOD data.
    \vspace{-0.2cm}
    \item We suggest a novel balanced energy regularization loss to address the class imbalance in auxiliary OOD data.
     \vspace{-0.2cm}
    \item The proposed balanced loss performs better for OOD detection than the previous energy regularization loss.
    \vspace{-0.6cm}
    \item The SOTA performance for OOD detection in two tasks is achieved by our OOD detection method.
    \end{itemize}

    \vspace{-0.5cm}

\section{Related Work}
\label{section_related_works}
\vspace{-0.1cm}

\subsection{OOD in Image Classification}
\vspace{-0.1cm}

    In the image classification task, there are two main approaches for OOD detection in deep neural networks. The first is a method to attain prediction uncertainty for a pre-trained model
    ~\cite{liang2017enhancing,sastry2020detecting,liu2020energy,lee2018simple}.
    The second approach modifies the architecture or loss based on new training~\cite{hendrycks2018deep,liu2020energy,hein2019relu,papadopoulos2021outlier,devries2018learning}.
    The first method primarily suggests new measures to boost performance.
    Proposed measures include the baseline MSP~\cite{hendrycks2016baseline}, Mahalanobis distance~\cite{lee2018simple}, the distance from the distribution of the Gram matrix~\cite{sastry2020detecting}, and free energy whose value is computed from the logit rather than the probability~\cite{liu2020energy}.

    The second method mainly uses auxiliary data as OOD data to enhance performance.
    Representatively, there are OE~\cite{hendrycks2018deep} and EnergyOE~\cite{liu2020energy} as methods of using auxiliary data for learning.
    OECC~\cite{papadopoulos2021outlier} replaced OE's cross entropy loss with total variance loss and included calibration loss.
    In UDG~\cite{yang2021semantically}, auxiliary data is once again divided into OOD samples and in-distribution samples through unsupervised dual grouping to improve performance. The more challenging SC-OOD benchmark is also proposed by UDG.
\vspace{-0.1cm}
    
\subsection{OOD in Semantic Segmentation}
\vspace{-0.1cm}

    In general, semantic segmentation's OOD detection benchmarks~\cite{blum2021fishyscapes,lis2019detecting,hendrycks2019scaling} resolve the detection problem of OOD that appears in complex urban driving scenes~\cite{cordts2016cityscapes}.
    The assessment method follows the same criteria as image classification~\cite{hendrycks2016baseline}, but instead of evaluating images as a whole, it does so pixel-by-pixel.
    Similarly, there are numerous ways to enhance OOD detection by putting forth new measures such as MSP~\cite{hendrycks2019scaling}, Entropy~\cite{hendrycks2016baseline}, Mahalanobis~\cite{lee2018simple}, and Energy~\cite{liu2020energy}.
    Newly proposed methods include Max Logit~\cite{hendrycks2019scaling}, which is the maximum value of logit, and Standardized Max Logit (SML)~\cite{jung2021standardized}, which is improved by reducing the deviation of logit by class.
    On the other hand, Bayesian Deeplab~\cite{mukhoti2018evaluating} measures uncertainty through a dropout layer based on Bayesian estimation.
    Image Resynthesis~\cite{lis2019detecting} and Synboost~\cite{di2021pixel} are algorithms that reconstruct and use new data from existing data through a generative model.

    Similar to the task of classifying images, the majority of the top techniques use auxiliary data as OOD data.
    The OOD data is synthesized by cut-pasting the object's mask from auxiliary dataset, or is raw pixels from auxiliary dataset.
    These approaches get the mask or pixels from Imagenet~\cite{bevandic2018discriminative,bevandic2019simultaneous}, ADE 20k ~\cite{grcic2022densehybrid}, and COCO~\cite{tian2021pixel,chan2021entropy}.
    By adopting the regularization loss suggested by OE and using additional post-processing, Meta-OOD~\cite{chan2021entropy} enhances performance.
    PEBAL~\cite{tian2021pixel} also adopts the energy regularization loss proposed by EnergyOE and boost its performance through abstention learning and some additional regularization losses.
    DenseHybrid~\cite{grcic2022densehybrid} utilizes a hybrid model of discriminative and generative classifiers.
    The following approach~\cite{bevandic2019simultaneous,bevandic2018discrimainative} are  based on a binary classifier.
    \vspace{-0.1cm}

\subsection{OOD in Long-tailed Image Classification}
\vspace{-0.1cm}

    Real-world data frequently exhibits a long-tail distribution, and learning from such imbalanced data has been questioned~\cite{he2009learning}.
    Deep neural networks demonstrate the degradation of performance when training on data with class imbalance~\cite{cao2019learning}.
    There are primarily two approaches resolving the issue of class imbalance.
    The first is a technique for readjusting the weights for each sample in the training loss~\cite{cao2019learning, park2021influence}.
    The second approach is a method to train multiple expert models to ensemble~\cite{wang2020long}.

    PASCL~\cite{wang2022partial} tackles the OOD detection problem and finds difficulty in the long-tailed training set.
    Similarly, ~\cite{liu2019large} deals with the open set classification challenge in the long-tailed training set.
    Particularly, PASCL assesses the performance of MSP~\cite{hendrycks2016baseline}, OE~\cite{hendrycks2018deep}, EnergyOE~\cite{liu2020energy}, SOFL~\cite{mohseni2020self}, OECC~\cite{papadopoulos2021outlier}, and NTOM~\cite{chen2021atom} in the SC-OOD benchmark~\cite{yang2021semantically} as a baseline for the OOD detection problem in long-tailed image classification.
    By incorporating partiality and asymmetry to the existing supervised contrastive learning to accommodate the long-tailed situation, PASCL achieves SOTA performance.

\section{Proposed Method}
\label{proposed_method}
\vspace{-0.1cm}

\subsection{Preliminary}
\vspace{-0.1cm}

\label{prelim}
    We can formalize a discriminative neural classifier as $f(\mathbf{x}):\mathbb{R}^D\to\mathbb{R}^K$
    ,which maps an input image $\mathbf{x}$ with $D$ dimension to a real-valued vector (logit) with $K$ dimension which is a number of classes.
    Probability vector $F(\mathbf{x})$ is computed as $Softmax(f(\mathbf{x}))$, which satisfies $\mathbf{1}^{T}F(\mathbf{x})=1$ and $F(\mathbf{x}) \geq 0$. 
    $y \in \{1,2,\dots, K\}$ represents the class label. 
    $f_{y}(\mathbf{x})$ and $F_{y}(\mathbf{x})$ indicates the $y$ th index of $f(\mathbf{x})$ and $F(\mathbf{x})$, respectively .
    
    Outlier Exposure(OE)~\cite{hendrycks2018deep} leverages the cross-entropy loss for the existing training (in-distribution) data and the regularization loss for the auxiliary (OOD) data.
    The minimization goal for the maximum softmax probability baseline detector is as follows:
    \begin{equation} 
    \small
    \label{equation:OE}
    \min\limits_{\theta}\quad  \mathbb{E}_{(\mathbf{x},y)\sim{D_{in}^{train}}}[-\log F_{y}(\mathbf{x})]+\lambda L_{OE},   
\end{equation}
    where $L_{OE}=\mathbb{E}_{\mathbf{x}_{out}\sim{D_{out}^{train}}}[H(\mathbf{u};F(\mathbf{x}))]$, 
     whereas $D_{in}$ and $D_{out}$ denote the in-distribution(ID) training set and the OOD training set, respectively. $\mathbf{u}$ is uniform distribution and $H$ is cross entropy loss.
     $L_{OE}$ is a regularization loss for OE
     
Energy-based OOD detection (EnergyOE) also leverages the cross-entropy loss for the  training (ID) data and the regularization loss for the auxiliary (OOD) data. However, EnergyOE proposes energy regularization loss $L_{energy}$ different from that of $L_{OE}$, which is given by
\begin{equation} 
\label{equation:Energy}
\small
\begin{split}
    L_{energy}&=L_{in,hinge}+L_{out,hinge}
    \\ &=\mathbb{E}_{(\mathbf{x}_{in},y)\sim{D_{in}^{train}}}[(\max(0,E(\mathbf{x})-m_{in}))^{2}]
    \\ &+\mathbb{E}_{\mathbf{x}_{out}\sim{D_{out}^{train}}}[(\max(0,E(\mathbf{x})-m_{out}))^{2}],
\end{split}
\end{equation}
where $E(\mathbf{x};f)=-T \cdot \log(\sum_{j=1}^{K}e^{f_{j}(\mathbf{x}))/T})$.
Energy function $E(\mathbf{x};f)$ is computed as LogSumExp of logit with temperature scaling,  In most cases,  temperature $T$=$1$. Energy regularization loss is the sum of squared hinge losses for energy with each of the existing (ID) data and the auxiliary (OOD) data.   
\vspace{-0.1cm}

\subsection{Balanced Energy regularization loss}
\label{section:regularization_loss}
\vspace{-0.1cm}

    Given the property of an OOD sample, our balanced energy regularization loss performs various regularizations for the OOD training data.
    The property of the OOD sample is modeled in the new term $Z$, which measures whether a sample belongs to a majority class or a minority class.
    The MAIN idea of our loss is to use a larger regularization to OOD samples of majority classes compared to OOD samples of minority classes.

    For the OOD data, we created the Z term to measure whether a sample is of the majority or minority class. 
    We require the prior probability of the OOD distribution to determine which class is the majority.
    Through inference on the pre-trained model of OOD data represented as auxiliary data, we obtain $N_{i}$, which is the number of samples that are classified as class $i$. Next, the prior probability of the OOD distribution is estimated by
    \begin{equation} 
    \small
    \label{equation:Prior}
    P(y=i|o)=\frac{N_{i}}{N_{1}+N_{2}+\dots+N_{K}}.
    \end{equation}    
    Using the discriminative neural classifier $f$, the posterior probability of the $i$-th class for a given image $\mathbf{x}$ is obtained by the softmax on the output of $f$, that is,
    \begin{equation} 
    \small
    \label{equation:softmaxprobability}
    P(y=i|\mathbf{x},o)=\frac{e^{f_{i}(\mathbf{x})}}{\sum_{j=1}^{K}  e^{f_{j}(\mathbf{x})}}.
    \end{equation} 
    The higher the posterior probability of $i$-th class for $\mathbf{x}$, the higher the probability that $\mathbf{x}$ belongs to $i$-th class. And the higher the prior probability of $i$-th class, the higher the probability that $i$-th class is a majority class.
    Hence the higher the product of $P(y=i|o)$ and  $P(y=i|\mathbf{x},o)$, the higher the possibility that $\mathbf{x}$ belongs to a majority class $i$. 
    From this result, a metric $Z$ to measure a possibility that $\mathbf{x}$ belongs to majority classes, is defined by 
    \begin{equation} 
    \small
    \label{equation:Zinitial}
    Z=\sum_{j=1}^{K} P(y=j|\mathbf{x},o)P(y=j|o).
    \end{equation} 
    In addition, we model additional generalized prior probability using hyperparameter $\gamma$.
    The degree of prior difference between classes is controlled by the hyperparameter $\gamma$.
    Finally, the generalized version $Z_{\gamma}$ is defined by
    \begin{equation} 
    \small
    \label{equation:Zfinal}
    Z_{\gamma}=\sum_{j=1}^{K} P(y=j|\mathbf{x},o)P_{\gamma}(y=j|o), 
 \end{equation} 
where  $P_{\gamma}(y=i|o)=L^{1}norm\{P^{\gamma}(y=i|o)\}$.
    For numerical stability, we apply L1-normalization after multiplying prior probability $P(y=i|o)$ by itself $\gamma$ times.
    If $\gamma$=$0$, then we model uniform prior probability and $Z_{\gamma}$ becomes constant value $\frac{1}{K}$. 
    If $\gamma$ is negative, we model the inverse distribution of prior probability.
    As $\gamma$ increases, the difference among prior probabilities of classes  increases.
    Based on the $Z_{\gamma}$ term, we design our balanced Energy regularization loss as follows.
    \begin{equation} 
    \small
    \label{equation:balancedenergy}
    \begin{split}
    &L_{energy,bal}=L_{in,hinge}+L_{out,bal}
    \\ &=\mathbb{E}_{(\mathbf{x}_{in},y)\sim{D_{in}^{train}}}[(\max(0,E(\mathbf{x})-m_{in}))^{2}]
    \\ &+\mathbb{E}_{\mathbf{x} \sim{D_{out}^{train}}}[(\max(0,E(\mathbf{x})-m_{out}-\alpha Z_{\gamma}))^{2}Z_{\gamma}],
\end{split}
\end{equation}
where $E(\mathbf{x};f)=-T \cdot \log(\sum_{j=1}^{K}e^{f_{j}(\mathbf{x}))/T})$.
    Our loss $L_{energy,bal}$ is the sum of  $L_{in,hinge}$ and $L_{out,bal}$.
    $L_{in,hinge}$ is squared hinge loss for in-distribution data, which is same as in Eq.~(\ref{equation:Energy}).
    $L_{out,bal}$ is our novel loss with two adaptive loss components that depend on $Z_{\gamma}$. The margin is the first component and the weight is the second component.
    As $Z_{\gamma}$ of a training sample increases, the loss margin and loss of weight increase, thus increasing the overall loss.
    First, the adaptive loss margin provides an additional $Z_{\gamma}$-proportional margin in the squared hinge loss for auxiliary data. 
    As a result, our adaptive loss margin is defined by $\alpha$$\cdot$$Z_{\gamma}$ which is $Z_{\gamma}$ multiplied by hyperparameter $\alpha$.
    Second, the adaptive loss weight gives a weight proportional to $Z_{\gamma}$ for the squared hinge loss. 
    Thus, the squared hinge loss is multiplied by our adaptive loss weight $Z_{\gamma}$ at the end.

\vspace{-0.2cm}
\subsection{Training Procedure}
\vspace{-0.3cm}
\SetKwInput{KwInput}{Input}                
\SetKwInput{KwOutput}{Output}              
\SetKwInput{Kwonestep}{Step1}
\SetKwInput{Kwtwostep}{Step2}
\begin{algorithm}[h]
\algsetup{linenosize=\tiny}
\footnotesize %
    \KwInput{$f$:Pre-trained model}
    \KwData{
    $D_{in}$:in-distribution training set,\\
    \quad\quad\quad  $D_{out}$:OOD training set
    }
    
    \Kwonestep{
    \textbf{Inference on OOD training set }}
    Load the weight of pre-trained model $f$; \\
    $ N_{j} \longleftarrow 0$, for all $j$=$1$ to $K$\\
    \For{$t=1$ to $T_{1}$}{
    Sample a mini batch $D_{mini,o}$ from $D_{out}$ \\
    Inference on the mini batch $f(D_{mini,o})$ \\
    \For{$j=1$ to $K$}{
    $n_{j}\longleftarrow$ count($\max\limits_{i}f(D_{mini,o})$, $j$) \\
    $N_{j}\longleftarrow N_{j}+n_{j}$
    }
    }
    Compute prior probability of OOD as Eq.~(\ref{equation:Prior}).
    
    \Kwtwostep{
    \textbf{Fine-tuning the pre-trained model}}
    
    \For{$t=T_{1}+1$ to $T_{2}$}
    {
       Sample mini-batches $D_{mini,i}$ and $D_{mini,o}$\\ from $D_{in}$ and  $D_{out}$, respectively. \\
       Update unfrozen classification layers of $f$ \\
       by minimizing Eq.~(\ref{equation:balancedenergylearning}).
    }
\caption{Balanced Energy Learning}
\label{algo}
\end{algorithm}
\vspace{-0.1cm}
\vspace{-0.2cm}

Our method leverages the cross-entropy loss for the existing training (ID) data and the regularization loss for the auxiliary (OOD) data as Outlier Exposure (OE).
Therefore, our minimizing objective is as follows:
    \begin{equation} 
    \small
    \label{equation:balancedenergylearning}
    \begin{split}
    &\min\limits_{\theta}\quad  \mathbb{E}_{(\mathbf{x},y)\sim{D_{in}^{train}}}[-\log F_{y}(\mathbf{x})]+\lambda L_{energy,bal}.   
    \end{split}
    \end{equation}
Balanced energy regularization loss $L_{energy,bal}$ is  defined in  Section~\ref{section:regularization_loss}.

Next, we summarize our balanced energy learning process in Algorithm.~\ref{algo}. 
Our approach is predicated on the idea that we have a model that has already been trained following standard neural network training (ST). Therefore, the input is a pre-trained neural classifier $f$ by ST process.
In the image classification task, $D_{in}$ is an original training image set, $D_{out}$ is an unlabeled image set of auxiliary data. 
In the semantic segmentation task, $D_{in}$ is an original training pixel set, $D_{out}$ is a pixel set that is synthesized by a cut-pasted OOD mask from auxiliary data.
Finally, our approach entails two steps. 
The first step is to determine $N_{i}$ by concluding model $f$, after which the prior probability of OOD is calculated.
The process of fine-tuning using our balanced energy regularization loss is the second step.

\section{Experiments}
\vspace{-0.1cm}

\subsection{Experiment Settings}
\vspace{-0.1cm}

\subsubsection{Dataset}
\vspace{-0.1cm}

{\bf Semantic segmentation task:} We use Cityscapes~\cite{cordts2016cityscapes} dataset as ID data. We use the object mask of COCO~\cite{lin2014microsoft} dataset as auxiliary data. 
For the OOD test data, we use Fishyscapes~\cite{blum2021fishyscapes} dataset and Road Anomaly~\cite{lis2019detecting} dataset.\\
\vspace{-0.3cm}

\noindent
{\bf Long-tailed image classification task:} We use two long-tailed image classification datasets CIFAR 10-LT~\cite{cao2019learning} and CIFAR 100-LT~\cite{cao2019learning} as ID data. Imbalance ratio $\rho$=$100$ following~\cite{wang2022partial}. we utilized TinyImages 80M~\cite{torralba200880} dataset as auxiliary data. 
For the OOD test data, we use six datasets (CIFAR~\cite{krizhevsky2009learning}, Texture~\cite{cimpoi2014describing}, SVHN~\cite{netzer2011reading}, LSUN~\cite{yu2015lsun}, Places365~\cite{zhou2017places}, and TinyImagenet~\cite{le2015tiny}) introduced in the SC-OOD benchmark~\cite{yang2021semantically}.\\
\vspace{-0.3cm}

\noindent
{\bf Image classification task:}  We use CIFAR10~\cite{krizhevsky2009learning}and CIFAR100~\cite{krizhevsky2009learning} as ID data, and the rest are the same as in the case of long-tailed.
\vspace{-0.4cm}
\subsubsection{Model}
\vspace{-0.1cm}

Following~\cite{tian2021pixel}, we use the semantic segmentation model of Deeplabv3$+$ like architecture with a WideResNet38~\cite{zhou2019semantic}. In long-tailed image classification and image classification task, we use ResNet18~\cite{he2016deep} model as in~\cite{wang2022partial}. To confirm generality in the long-tailed image classification task, We also use the WideResNet (WRN-40-2)~\cite{zagoruyko2016wide} model.
\vspace{-0.4cm}

\subsubsection{Implementation Details}
\vspace{-0.2cm}

\label{section:Implementation}
In semantic segmentation task, we employ a similar method as PEBAL~\cite{tian2021pixel}. 
we load the semantic segmentation pre-trained model by NVIDIA~\cite{zhou2019semantic} on the Cityscapes dataset. 
As in PEBAL, we build the auxiliary data by cutting and pasting the mask from the COCO data.
The prior probability is then drived from OOD pixels for random sample of 10000 scene images. We use the same training configuration as PEBAL for fine-tuning, with 20 epochs, Adam as the optimizer, and a learning rate of 0.00001. The distinction is that batch size is configured to be 8.

In both long-tailed and normal image classification, we employ a similar method as EnergyOE~\cite{liu2020energy}.
By using the ST method, we can obtain a pre-trained model following the setting of OE~\cite{hendrycks2018deep}. Our auxiliary dataset is a subset of TinyImages80M with 300K images. Next, 300K images are used to extract the prior probability. We only use 30K subset images for training following PASCAL~\cite{wang2022partial}.
For fine-tuning, we use an almost identical training setting as EnergyOE, where the initial learning rate is 0.001 with cosine decay~\cite{loshchilov2016sgdr} and the batch size is 128 for in-distribution data and 256 for unlabeled OOD training data. 
We summarize our hyperparameter setting for all tasks in Table~\ref{table_hyper}.

\begin{table}[t!]
\centering
\captionsetup{font=footnotesize}
\caption{Hyperparameter(hyper.P) setting in all tasks: Semantic segmentation (Seg), Long-tailed image classifcation(Long-tailed Cls), and image classification(Cls) }
\label{table_hyper}
\scriptsize
\begin{tabular}{|c|c|cc|cc|}
\hline
TASK      & \multirow{2}{*}{Seg} & \multicolumn{2}{c|}{Long-tailed Cls}    & \multicolumn{2}{c|}{Cls}                \\ \cline{1-1} \cline{3-6} 
hyper.P   &                      & \multicolumn{1}{c|}{CIF-10} & CIF-100 & \multicolumn{1}{c|}{CIF-10} & CIF-100 \\ \hline
class num $K$ & 19                   & \multicolumn{1}{c|}{10}      & 100      & \multicolumn{1}{c|}{10}      & 100      \\ \hline
temp $T$ & 1                   & \multicolumn{1}{c|}{1}      & 1      & \multicolumn{1}{c|}{1}      & 1      \\ \hline

$\lambda$ & 0.1                   & \multicolumn{1}{c|}{0.1}      & 0.1     & \multicolumn{1}{c|}{0.1}      & 0.1      \\ \hline
$\alpha$     & 5                    & \multicolumn{1}{c|}{10}      & 100      & \multicolumn{1}{c|}{10}      & 100      \\ \hline
$\gamma$     & 3.0                  & \multicolumn{1}{c|}{0.75}    & 0.75     & \multicolumn{1}{c|} {0.25}
&
0.025    \\ \hline
$m_{in}$      & -12                  & \multicolumn{1}{c|}{-23}     & -27      & \multicolumn{1}{c|}{-23}     & -27      \\ \hline
$m_{out}$     & -6                   & \multicolumn{1}{c|}{-5}      & -5       & \multicolumn{1}{c|}{-5}      & -5       \\ \hline
\end{tabular}
\end{table}
    
\begin{table}[t!]
\centering
\scriptsize
\captionsetup{font=footnotesize}
\caption{Evaluation result on Fishyscapes test sets (Lost\&Found, Static) : OOD detection performance with AP and FPR. Compared Methods are: MSP; 
En$^\dagger$ (Entropy);
kNN$^\dagger$ (kNN Embedding - density)
SML;               BD$^\dagger$ (Bayesian Deeplab);
DSN$^\dagger$ (Density Single-layer NLL); 
DMN$^\dagger$ (Density Minimum NLL); 
IR$^\dagger$ (Image Resynthesis);
DLR$^\dagger$ (Density Logistic Regression);
SB$^\dagger$ (SynBoost); DODH$^\dagger$ (Discriminative 
Outlier Detection Head); 
OTVC$^\dagger$ (OoD Training - Void Class); DD$^\dagger$ (Dirichlet Deeplab); 
DH$^\dagger$ (DenseHybrid); 
PEBAL;  
\textbf{Ours: (Balanced Energy PEBAL)}
}
\label{table_seg_test}
\noindent
$R^{\dagger}$: Re-training, $E^{\dagger}$: Extra Network, 
$O^{\dagger}$: OoD Data. 
\begin{tabular}{c|c|c|c|cc|cc}
\hline
\multirow{2}{*}{Method} &
  \multirow{2}{*}{\begin{tabular}[c]{@{}c@{}}$R^{\dagger}$\end{tabular}} &
  \multirow{2}{*}{\begin{tabular}[c]{@{}c@{}}$E^{\dagger}$\end{tabular}} &
  \multirow{2}{*}{$O^{\dagger}$} &
  \multicolumn{2}{c|}{FS Lost \& Found} &
  \multicolumn{2}{c}{FS Static} \\ \cline{5-8} 
                                      &   &   &   & \multicolumn{1}{c|}{AP$\uparrow$}    & FPR$\downarrow$   & \multicolumn{1}{c|}{AP$\uparrow$}    & FPR$\downarrow$   \\ \hline
MSP ~\cite{hendrycks2019scaling}                                   & \ding{55} & \ding{55} & \ding{55} & \multicolumn{1}{c|}{1.77}  & 44.85 & \multicolumn{1}{c|}{12.88} & 39.83 \\
En$^\dagger$~\cite{hendrycks2016baseline}                               & \ding{55} & \ding{55} & \ding{55} & \multicolumn{1}{c|}{2.93}  & 44.83 & \multicolumn{1}{c|}{15.41} & 39.75 \\
kNN$^\dagger$ ~\cite{blum2021fishyscapes}               & \ding{55} & \ding{55} & \ding{55} & \multicolumn{1}{c|}{3.55}  & 30.02 & \multicolumn{1}{c|}{44.03} & 20.25 \\
SML~\cite{jung2021standardized}                                   & \ding{55} & \ding{55} & \ding{55} & \multicolumn{1}{c|}{31.05} & 21.52 & \multicolumn{1}{c|}{53.11} & 19.64 
 \\ \hline BD$^\dagger$~\cite{mukhoti2018evaluating}                      & \ding{51} & \ding{55} & \ding{55} & \multicolumn{1}{c|}{9.81}  & 38.46 & \multicolumn{1}{c|}{48.70} & 15.05
 \\ \hline
DSN$^\dagger$~\cite{blum2021fishyscapes}              & \ding{55} & \ding{51} & \ding{55} & \multicolumn{1}{c|}{3.01}  & 32.90 & \multicolumn{1}{c|}{40.86} & 21.29 \\
DMN$^\dagger$~\cite{blum2021fishyscapes}                   & \ding{55} & \ding{51} & \ding{55} & \multicolumn{1}{c|}{4.25}  & 47.15 & \multicolumn{1}{c|}{62.14} & 17.43 \\
IR$^\dagger$~\cite{lis2019detecting}                    & \ding{55} & \ding{51} & \ding{55} & \multicolumn{1}{c|}{5.70}  & 48.05 & \multicolumn{1}{c|}{29.60} & 27.13 \\ \hline
DLR$^\dagger$~\cite{blum2021fishyscapes}           & \ding{55} & \ding{51} & \ding{51} & \multicolumn{1}{c|}{4.65}  & 24.36 & \multicolumn{1}{c|}{57.16} & 13.39 \\
SB$^\dagger$~\cite{di2021pixel}                              & \ding{55} & \ding{51} & \ding{51} & \multicolumn{1}{c|}{43.22} & 15.79 & \multicolumn{1}{c|}{72.59} & 18.75 \\ \hline
DODH$^\dagger$~\cite{bevandic2019simultaneous} & \ding{51} & \ding{51} & \ding{51} & \multicolumn{1}{c|}{31.31} & 19.02 & \multicolumn{1}{c|}{96.76} & 0.29  \\ \hline
OTVC$^\dagger$             & \ding{51} & \ding{55} & \ding{51} & \multicolumn{1}{c|}{10.29} & 22.11 & \multicolumn{1}{c|}{45.00} & 19.40 \\
DD$^\dagger$~\cite{malinin2018predictive}                     & \ding{51} & \ding{55} & \ding{51} & \multicolumn{1}{c|}{34.28} & 47.43 & \multicolumn{1}{c|}{31.30} & 84.60 \\
DH$^\dagger$~\cite{grcic2022densehybrid}                     & \ding{51} & \ding{55} & \ding{51} & \multicolumn{1}{c|}{47.06} & 3.97 & \multicolumn{1}{c|}{80.23} & 5.95 \\
PEBAL~\cite{tian2021pixel}                                 & \ding{51} & \ding{55} & \ding{51} & \multicolumn{1}{c|}{44.17} & 7.58  & \multicolumn{1}{c|}{92.38} & 1.73  \\ 
\textbf{Ours}          & \ding{51} & \ding{55} & \ding{51} & \multicolumn{1}{c|}{\textbf{51.83}}      & \textbf{3.76}      & \multicolumn{1}{c|}{\textbf{94.62}}      &   \textbf{0.99}   \\ \hline
\end{tabular}
\end{table}

\begin{table*}[ht!]
\centering
\scriptsize
\captionsetup{font=footnotesize}
\caption{Evaluation result on Fishyscapes validation sets and Road Anomaly test set : OOD detection performance with AUROC, AP, and FPR}
\label{table_seg_val}
\begin{tabular}{c|ccc|ccc|ccc}
\hline
\multirow{2}{*}{Method} &
  \multicolumn{3}{c|}{FS Lost \& Found} &
  \multicolumn{3}{c|}{FS Static} &
  \multicolumn{3}{c}{Road Anomaly} \\ \cline{2-10} 
 &
  \multicolumn{1}{c|}{AUC$\uparrow$} &
  \multicolumn{1}{c|}{AP$\uparrow$} &
  FPR$\downarrow$ &
  \multicolumn{1}{c|}{AUC$\uparrow$} &
  \multicolumn{1}{c|}{AP$\uparrow$} &
  FPR$\downarrow$ &
  \multicolumn{1}{c|}{AUC$\uparrow$} &
  \multicolumn{1}{c|}{AP$\uparrow$} &
  FPR$\downarrow$ \\ \hline
MSP~\cite{hendrycks2019scaling} &
  \multicolumn{1}{c|}{89.29} &
  \multicolumn{1}{c|}{4.59} &
  40.59 &
  \multicolumn{1}{c|}{92.36} &
  \multicolumn{1}{c|}{19.09} &
  23.99 &
  \multicolumn{1}{c|}{67.53} &
  \multicolumn{1}{c|}{15.72} &
  71.38 \\
Max Logit~\cite{hendrycks2019scaling} &
  \multicolumn{1}{c|}{93.41} &
  \multicolumn{1}{c|}{14.59} &
  42.21 &
  \multicolumn{1}{c|}{95.66} &
  \multicolumn{1}{c|}{38.64} &
  18.26 &
  \multicolumn{1}{c|}{72.78} &
  \multicolumn{1}{c|}{18.98} &
  70.48 \\
Entropy~\cite{hendrycks2016baseline} &
  \multicolumn{1}{c|}{90.82} &
  \multicolumn{1}{c|}{10.36} &
  40.34 &
  \multicolumn{1}{c|}{93.14} &
  \multicolumn{1}{c|}{26.77} &
  23.31 &
  \multicolumn{1}{c|}{68.80} &
  \multicolumn{1}{c|}{16.97} &
  71.10 \\
Energy~\cite{liu2020energy} &
  \multicolumn{1}{c|}{93.72} &
  \multicolumn{1}{c|}{16.05} &
  41.78 &
  \multicolumn{1}{c|}{95.90} &
  \multicolumn{1}{c|}{41.68} &
  17.78 &
  \multicolumn{1}{c|}{73.35} &
  \multicolumn{1}{c|}{19.54} &
  70.17 \\
Mahalanobis~\cite{lee2018simple} &
  \multicolumn{1}{c|}{96.75} &
  \multicolumn{1}{c|}{56.57} &
  11.24 &
  \multicolumn{1}{c|}{96.76} &
  \multicolumn{1}{c|}{27.37} &
  11.7 &
  \multicolumn{1}{c|}{62.85} &
  \multicolumn{1}{c|}{14.37} &
  81.09 \\
Meta-OOD~\cite{chan2021entropy} &
  \multicolumn{1}{c|}{93.06} &
  \multicolumn{1}{c|}{41.31} &
  37.69 &
  \multicolumn{1}{c|}{97.56} &
  \multicolumn{1}{c|}{72.91} &
  13.57 &
  \multicolumn{1}{c|}{-} &
  \multicolumn{1}{c|}{-} &
  - \\
Synboost~\cite{di2021pixel} &
  \multicolumn{1}{c|}{96.21} &
  \multicolumn{1}{c|}{60.58} &
  31.02 &
  \multicolumn{1}{c|}{95.87} &
  \multicolumn{1}{c|}{66.44} &
  25.59 &
  \multicolumn{1}{c|}{81.91} &
  \multicolumn{1}{c|}{38.21} &
  64.75 \\
SML~\cite{jung2021standardized} &
  \multicolumn{1}{c|}{94.97} &
  \multicolumn{1}{c|}{22.74} &
  33.49 &
  \multicolumn{1}{c|}{97.25} &
  \multicolumn{1}{c|}{66.72} &
  12.14 &
  \multicolumn{1}{c|}{75.16} &
  \multicolumn{1}{c|}{17.52} &
  70.70 \\
Deep Gambler~\cite{liu2019deep} &
  \multicolumn{1}{c|}{97.82} &
  \multicolumn{1}{c|}{31.34} &
  10.16 &
  \multicolumn{1}{c|}{98.88} &
  \multicolumn{1}{c|}{84.57} &
  3.39 &
  \multicolumn{1}{c|}{78.29} &
  \multicolumn{1}{c|}{23.26} &
  65.12 \\ \hline
PEBAL~\cite{tian2021pixel} &
  \multicolumn{1}{c|}{98.96} &
  \multicolumn{1}{c|}{58.81} &
  4.76 &
  \multicolumn{1}{c|}{\textbf{99.61}} &
  \multicolumn{1}{c|}{92.08} &
  1.52 &
  \multicolumn{1}{c|}{87.63} &
  \multicolumn{1}{c|}{\textbf{45.10}} &
  44.58 \\
\textbf{Balanced Energy PEBAL (Ours)} &
  \multicolumn{1}{c|}{\textbf{99.03}} &
  \multicolumn{1}{c|}{\textbf{67.07}} &
  \textbf{2.93} &
  \multicolumn{1}{c|}{99.55} &
  \multicolumn{1}{c|}{\textbf{92.49}} &
  \textbf{1.17} &
  \multicolumn{1}{c|}{\textbf{88.36}} &
  \multicolumn{1}{c|}{43.58} &
  \textbf{41.54} \\ \hline
EnergyOE~\cite{liu2020energy} &
  \multicolumn{1}{c|}{98.14} &
  \multicolumn{1}{c|}{45.61} &
  8.21 &
  \multicolumn{1}{c|}{99.32} &
  \multicolumn{1}{c|}{89.12} &
  2.62 &
  \multicolumn{1}{c|}{83.32} &
  \multicolumn{1}{c|}{32.59} &
  53.01 \\
\textbf{Balanced EnergyOE (Ours)} &
  \multicolumn{1}{c|}{\textbf{98.42}} &
  \multicolumn{1}{c|}{\textbf{54.58}} &
  \textbf{6.70} &
  \multicolumn{1}{c|}{\textbf{99.43}} &
  \multicolumn{1}{c|}{\textbf{91.77}} &
  \textbf{1.63} &
  \multicolumn{1}{c|}{\textbf{85.50}} &
  \multicolumn{1}{c|}{\textbf{34.90}} &
  \textbf{46.60} \\ \hline
\end{tabular}
\end{table*}
 

\vspace{-0.1cm}

\subsection{Semantic Segmentation}
\vspace{-0.1cm}

Table~\ref{table_seg_test} shows the results of our approach on the Fishyscapes leaderboard.
The technique that replaces the energy regularization loss in PEBAL with our balanced energy regularization loss is known as our balanced energy PEBAL. Our approach outperforms PEBAL and achieves SOTA in a methodology that utilize OOD data and require no extra network.

Table~\ref{table_seg_val} presents the results of our method on the Fishyscapes validation sets and Road Anomaly test set. Here, we compare our method with not only PEBAL, but also EnergyOE, which is a baseline that use only energy regularization loss. We show that our loss enhances both compared to using the original energy regularization loss. Furthermore, balanced energy PEBAL has superior performance compared to other baselines.
\vspace{-0.1cm}

\subsection{Long-Tailed Image Classification}
\vspace{-0.1cm}

\begin{table}[t!]
\captionsetup{font=footnotesize}
\caption{Evaluation result on CIFAR10-LT using ResNet18; (a): OOD detection performance with AUROC,AP and FPR; Mean over six random runs are reported(OE,EnergyOE,Ours). (b): Comparison result with other methods; average (over 6 datasets) OOD detection performance (AUROC,AP, FPR) and classification accuracy (ACC).}
\vspace{-0.2cm}
\begin{subtable}{1\linewidth}
\centering
\scriptsize
\caption{}
\label{cifar10_table_top}

\begin{tabular}{c|c|c|c|c}
\hline

Dataset                    & Method          & AUC$\uparrow$    & AP$\uparrow$      & FPR$\downarrow$            \\ \hline
\multirow{3}{*}{Texture}   & OE (tune)       & 87.98          & 80.05          & 45.54          \\
                           & EnergyOE (tune) & 95.53          & \textbf{92.93} & 23.26          \\ \cline{2-5} 
                           & \textbf{Ours}            & \textbf{95.69} & 92.38          & \textbf{21.26} \\ \hline
\multirow{3}{*}{SVHN}      & OE (tune)       & 92.10          & 95.52          & 27.37          \\
                           & EnergyOE (tune) & 96.63          & 98.46          & 14.52          \\ \cline{2-5} 
                           & Ours            & \textbf{97.74} & \textbf{98.89} & \textbf{9.87}  \\ \hline
\multirow{3}{*}{CIFAR100}  & OE (tune)       & 78.24          & 76.35          & 65.28          \\
                           & EnergyOE (tune) & 84.44          & 84.63          & 59.92          \\ \cline{2-5} 
                           & \textbf{Ours}             & \textbf{85.20} & \textbf{84.98} & \textbf{57.95} \\ \hline
\multirow{3}{*}{\begin{tabular}[c]{@{}c@{}}Tiny\\ ImageNet\end{tabular}} & OE (tune) & 81.47 & 75.79 & 58.68 \\
                           & EnergyOE (tune) & 88.40          & 84.95          & 45.17          \\ \cline{2-5} 
                           & \textbf{Ours}             & \textbf{88.92} & \textbf{84.98} & \textbf{42.38} \\ \hline
\multirow{3}{*}{LSUN}      & OE (tune)       & 86.19          & 85.85          & 54.49          \\
                           & EnergyOE (tune) & 94.00          & \textbf{93.70} & 26.96          \\ \cline{2-5} 
                           & \textbf{Ours}             & \textbf{94.48} & 93.15          & \textbf{23.88} \\ \hline
\multirow{3}{*}{Places365} & OE (tune)       & 84.27          & 93.84          & 59.08          \\
                           & EnergyOE (tune) & 92.51          & 97.14          & 32.88          \\ \cline{2-5} 
                           & \textbf{Ours}            & \textbf{93.35} & \textbf{97.23} & \textbf{28.25} \\ \hline
\multirow{3}{*}{Average}   & OE (tune)       & 85.04          & 84.57          & 51.74          \\
                           & EnergyOE (tune) & 91.92          & \textbf{91.97} & 33.79          \\ \cline{2-5} 
                           & \textbf{Ours}             & \textbf{92.56} & 91.94          & \textbf{30.60} \\ \hline
\end{tabular}

\caption{}
\begin{tabular}{c|c|c|c|c|c}
\hline

Dataset                  & Method              & AUC$\uparrow$    & AP$\uparrow$      & FPR$\downarrow$     & ACC$\uparrow$            \\ \hline
\multirow{9}{*}{Average} & MSP~\cite{hendrycks2016baseline}(ST)           & 74.55
& 72.26
& 61.30
& 73.28
\\
                         & Energy~\cite{liu2020energy}(ST)        & 80.33
                         & 77.02
                         & 53.82 
                         & 73.28
                         \\
                         & OECC~\cite{papadopoulos2021outlier}               & 87.28          & 86.29          & 45.24          & 60.16          \\
                         & EnergyOE~\cite{liu2020energy}(scratch) & 89.31          & 88.92          & 40.88          & 74.68          \\
                         & OE~\cite{hendrycks2018deep}(scratch)       & 89.77          & 87.25          & 34.65          & 73.84          \\
                         & PASCL~\cite{wang2022partial}             & 90.99          & 89.24          & 33.36          & 77.08 \\
                         & Open-Sampling~\cite{wei2022open}             & 90.24          & 85.44          & 31.00          & 77.06 \\
                         & OE~\cite{hendrycks2018deep}(tune)          & 85.04          & 84.57          & 51.74          & 69.79          \\
                         & EnergyOE~\cite{liu2020energy}(tune)    & 91.92          & \textbf{91.97} & 33.79          & 74.53          \\ \cline{2-6} 
                         & \textbf{Ours}               & \textbf{92.56} & 91.94          & \textbf{30.60} & 76.22          \\
                         &  \textbf{Ours+AdjLogit~\cite{menon2020long}}               &\textbf{92.56} & 91.94 & \textbf{30.60} & \textbf{81.37}          \\                            
                         \hline
\end{tabular}
\label{cifar10_table_down}
\end{subtable}
\end{table}

\begin{table}[t!]
\captionsetup{font=footnotesize}
\caption{Evaluation result on CIFAR100-LT using ResNet18; (a): OOD detection performance with AUROC,AP and FPR; Mean over six random runs are reported(OE,EnergyOE,Ours). (b): Comparison result with other methods; average (over 6 datasets) OOD detection performance (AUROC,AP, FPR) and classification accuracy (ACC).}
\vspace{-0.2cm}
\begin{subtable}{1\linewidth}
\centering
\scriptsize
\caption{}
\label{cifar100_table_top}

\begin{tabular}{c|c|c|c|c}
\hline
Dataset                    & Method           & AUC$\uparrow$    & AP$\uparrow$      & FPR$\downarrow$            \\ \hline
\multirow{3}{*}{Texture}   & OE (tune)       & 66.29          & 51.98          & 84.04          \\
                           & EnergyOE (tune) & 79.56          & 70.88          & 68.60          \\ \cline{2-5} 
                           & \textbf{Ours}             & \textbf{82.10} & \textbf{73.09} & \textbf{64.19} \\ \hline
\multirow{3}{*}{SVHN}      & OE (tune)       & 74.93          & 85.41          & 63.94          \\
                           & EnergyOE (tune) & 86.19          & 91.74          & 42.27          \\ \cline{2-5} 
                           & \textbf{Ours}             & \textbf{88.66} & \textbf{92.88} & \textbf{33.79} \\ \hline
\multirow{3}{*}{CIFAR10}   & OE (tune)       & 59.44          & 56.34          & 84.70          \\
                                                                         & EnergyOE (tune) & \textbf{61.15} & \textbf{56.66} & \textbf{82.60} \\ \cline{2-5} 
                           & \textbf{Ours}             & 59.40          & 54.97          & 85.16          \\ \hline
\multirow{3}{*}{\begin{tabular}[c]{@{}c@{}}Tiny\\ ImageNet\end{tabular}} & OE (tune)       & 66.24          & 51.07          & 80.04          \\
                           & EnergyOE (tune) & 70.78          & 55.90          & 74.43          \\ \cline{2-5} 
                           & \textbf{Ours}             & \textbf{71.42} & \textbf{56.52} & \textbf{74.22} \\ \hline
\multirow{3}{*}{LSUN}      & OE (tune)       & 73.46          & 59.07          & 73.05          \\
                           & EnergyOE (tune) & 81.61          & 69.16          & 57.37          \\ \cline{2-5} 
                           & \textbf{Ours}             & \textbf{83.83} & \textbf{71.23} & \textbf{52.04} \\ \hline
\multirow{3}{*}{Places365} & OE (tune)       & 71.70          & 85.08          & 74.62          \\
                           & EnergyOE (tune) & 79.12          & 89.09          & 61.96          \\ \cline{2-5} 
                           & \textbf{Ours}             & \textbf{81.10} & \textbf{89.94} & \textbf{57.52} \\ \hline
\multirow{3}{*}{Average}   & OE (tune)       & 68.68          & 64.83          & 76.73          \\
                           & EnergyOE (tune) & 76.40          & 72.24          & 64.54          \\ \cline{2-5} 
                           & \textbf{Ours}            & \textbf{77.75} & \textbf{73.10} & \textbf{61.15} \\ \hline
\end{tabular}

\caption{}
\begin{tabular}{c|c|c|c|c|c}
\hline
Dataset                  & Method             & AUC$\uparrow$    & AP$\uparrow$      & FPR$\downarrow$        & ACC$\uparrow$            \\ \hline
\multirow{9}{*}{Average} & MSP~\cite{hendrycks2016baseline}(ST)           & 61.17
& 58.10
& 83.30
& 40.22
\\
                         & Energy~\cite{liu2020energy}(ST)        & 64.08
                         & 59.86
                         & 80.59
                         & 40.22
                         \\
                         & OECC~\cite{papadopoulos2021outlier}               & 70.38          & 66.87          & 73.15          & 32.93          \\
                         & EnergyOE~\cite{liu2020energy}(scratch) & 71.10          & 67.23          & 71.78          & 39.05          \\
                         & OE~\cite{hendrycks2018deep}(scratch)       & 72.91          & 67.16          & 68.89          & 39.04          \\
                         & PASCL~\cite{wang2022partial}             & 73.32          & 67.18          & 67.44          & 43.10 \\
                         & Open-Sampling~\cite{wei2022open}             & 74.46          & 69.49          & 66.82         & 39.86 \\                         
                         
                         & OE~\cite{hendrycks2018deep}(tune)          & 68.68          & 64.83          & 76.73          & 38.93          \\
                         & EnergyOE~\cite{liu2020energy}(tune)    & 76.40          & 72.24          & 64.54          & 40.65          \\ \cline{2-6} 
                         &  \textbf{Ours}               & \textbf{77.75} & \textbf{73.10} & \textbf{61.15} & 41.05          \\
                         &  \textbf{Ours+AdjLogit~\cite{menon2020long}}               & \textbf{77.75} & \textbf{73.10} & \textbf{61.15} &  \textbf{45.66}          \\                         
                         \hline
\end{tabular}
\label{cifar100_table_down}
\end{subtable}
\end{table}

Table~\ref{cifar10_table_top} shows the CIFAR10-LT experiment results of comparison with the existing baselines (OE, EnergyOE) in detail for each data. We present the average of all over six random runs (OE, EnergyOE, Ours). 
16 out of 18 show better performance than baseline in AUROC, AP, and FPR. 
Table~\ref{cifar10_table_down} depicts the summary of the CIFAR10-LT experiment results which is the average performance over 6 datasets compared to the other methods.
When compared to the baseline results for the model before fine-tuning, our approach performs better.
The point to note is that accuracy also improves compared to using the energy regularization loss. When evaluating the accuracy, PASCL has stage 2 which uses the loss proposed by AdjLogit~\cite{menon2020long} to improve long-tailed classification accuracy. For a fair comparison with PASCL, we also report our accuracy after going through stage 2 used by PASCL. 
Our method outperforms the SOTA algorithm, PASCL.

Table~\ref{cifar100_table_top} details the comparison between the CIFAR100-LT experiment results and the current baselines (OE, EnergyOE). We present the average of all six random runs (OE, EnergyOE, Ours).15 out of 18 show better performance than baseline in AUROC, AP, and FPR. 
Table~\ref{cifar100_table_down} presents the summary of CIFAR100-LT experiment results which is the average performance over 6 datasets compared to the other methods. Similarly, as CIFAR10-LT, accuracy also improves compared to using the energy regularization loss. Our method outperforms the SOTA algorithm, PASCL.
\vspace{-0.1cm}

\subsection{Image Classification}
\vspace{-0.1cm}

\begin{table}[t!]
\captionsetup{font=footnotesize}
\caption{ Evaluation result on CIFAR using ResNet18; average (over 6 datasets) OOD detection performance (AUROC,AP, FPR) and classification accuracy (ACC) (a): Result on CIFAR10 (b): Result on CIFAR100}
\label{balanced_merge_table}
\vspace{-0.2cm}
\begin{subtable}{1\linewidth}
\centering
\scriptsize
\caption{}
\label{balanced_cifar10}
\begin{tabular}{c|c|c|c|c|c}
\hline
Dataset                  & Method         & AUC$\uparrow$    & AP$\uparrow$      & FPR$\downarrow$        & ACC$\uparrow$           \\ \hline
\multirow{5}{*}{Average} & MSP~\cite{hendrycks2016baseline}(ST)         & 89.25          & 86.63          & 31.32         & \textbf{93.69}          \\
                         & Energy~\cite{liu2020energy}(ST)      & 91.55          & 89.88          & 29.07         & \textbf{93.69}          \\
                         & OECC~\cite{papadopoulos2021outlier}      & 96.33           &    95.38      &   \textbf{14.36}       &      91.57     \\
                         & OE~\cite{hendrycks2018deep}(tune)       & 95.68          & 95.36          & 18.20         & 93.37          \\
                         & EnergyOE~\cite{liu2020energy}(tune) & 96.77          & \textbf{96.72}          & 14.82          & 93.30 \\ \cline{2-6} 
                         &  \textbf{Ours}           & \textbf{96.83} & 96.70 & 14.51 & 93.00          \\ \hline
\end{tabular}

\caption{}
\begin{tabular}{c|c|c|c|c|c}
\hline
Dataset                  & Method         & AUC$\uparrow$    & AP$\uparrow$      & FPR$\downarrow$        & ACC$\uparrow$               \\ \hline
\multirow{5}{*}{Average} & MSP~\cite{hendrycks2016baseline}(ST)        & 76.14          & 71.29          & 62.78          & \textbf{75.70}          \\
                         & Energy~\cite{liu2020energy}(ST)     & 79.78          & 73.31          & 57.59          & \textbf{75.70}          \\
                         & OECC~\cite{papadopoulos2021outlier}      &    84.03       &   77.94       &    45.26      &    69.55      \\
                         & OE~\cite{hendrycks2018deep}(tune)       & 82.76          & 77.93          & 51.72          & 74.33          \\
                         & EnergyOE~\cite{liu2020energy}(tune) & 85.84          & \textbf{80.99}          & 43.02          & 74.95 \\ \cline{2-6} 
                         &  \textbf{Ours}            & \textbf{85.85} & 80.91 & \textbf{42.93} & 74.83          \\ \hline
\end{tabular}
\label{balanced_cifar100}
\end{subtable}
\end{table}

Similar to the long-tailed image classification task, we compare our approach with the existing baselines (OE, EnergyOE). Table~\ref{balanced_cifar10} depicts the summary of CIFAR10 experiment results which is the average performance over 6 datasets compared to the other methods.
The accuracy performs marginally worse than using energy regularization loss, which is different from long-tailed case. However, the OOD performance is still improved compared to using energy regularization loss.
Table~\ref{balanced_cifar100} illustrates the summary of CIFAR100 experiment results which are the average performance over 6 datasets compared to the other methods.
Similar to the CIFAR10 experiment, OOD performance improves when using our losses compared to baseline.

\section{Discussion}
\subsection{Empirical Analysis}
\vspace{-0.1cm}
\begin{figure}[ht!]
\captionsetup{font=footnotesize}
\captionsetup[subfloat]{position=top}   
\centering
\subfloat[]{\includegraphics[width=0.75\columnwidth]{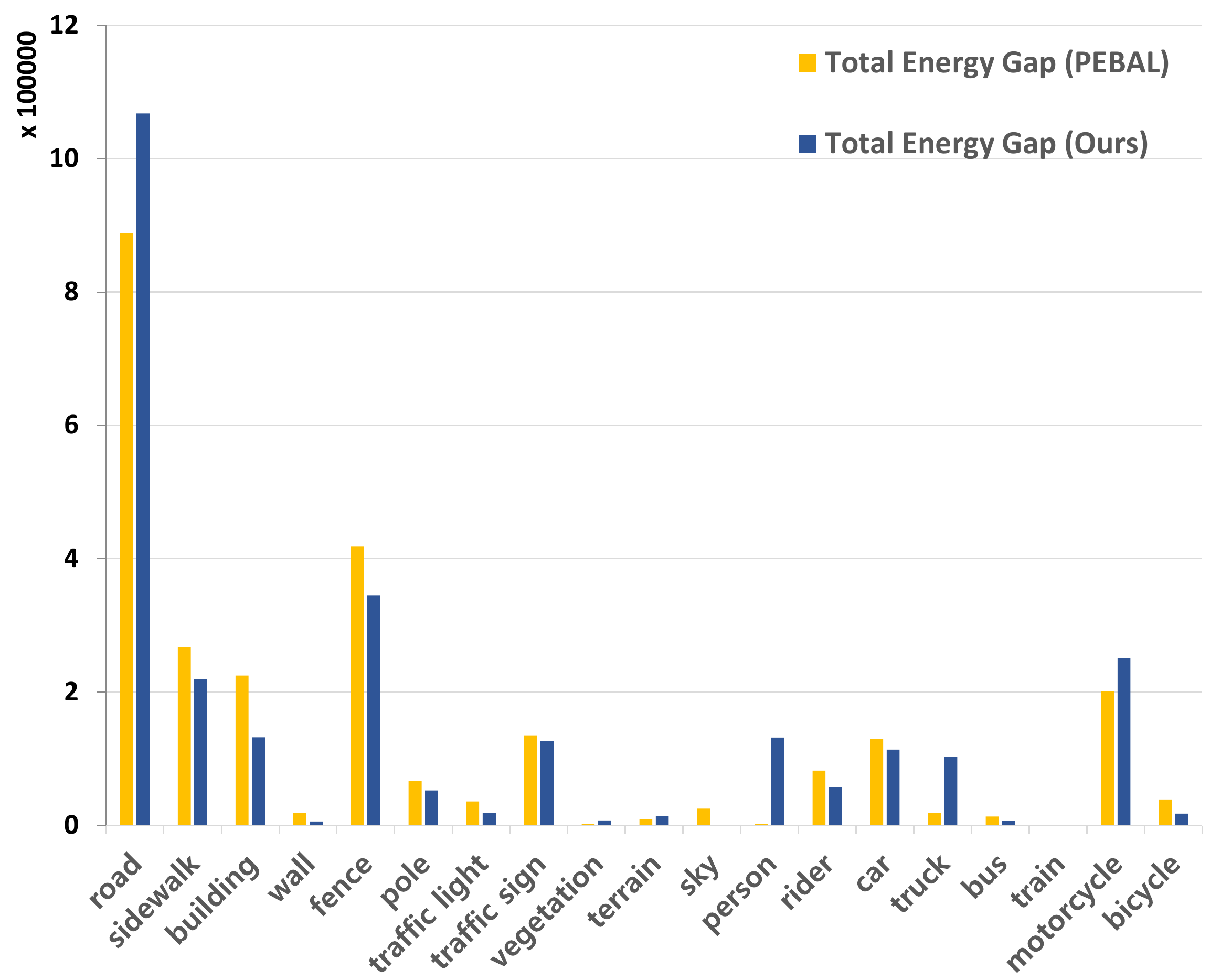}
\label{anal_fig_top}

}

\vfill
\centering
\subfloat[]{\includegraphics[width=0.85\columnwidth]{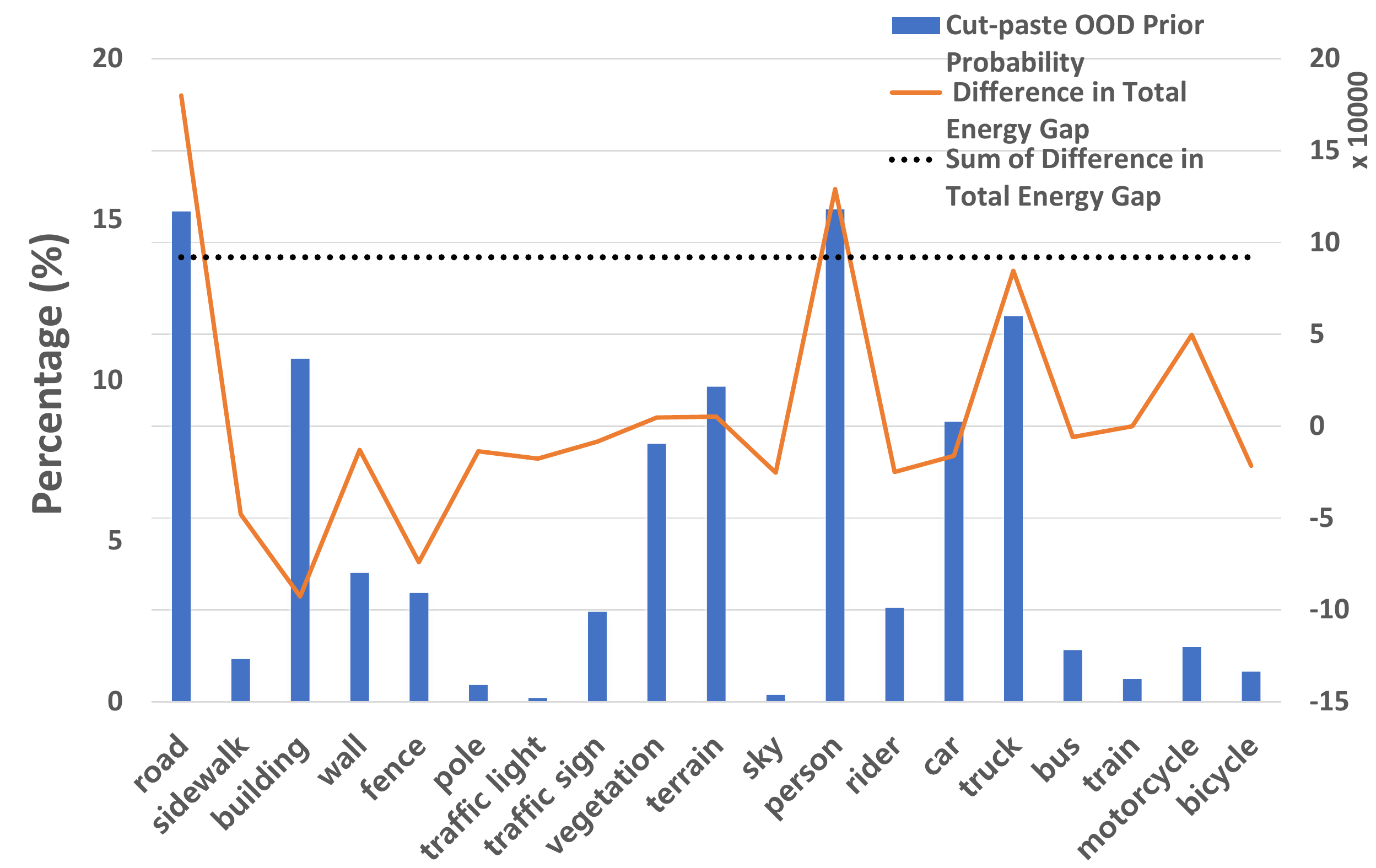}

\label{anal_fig_down}
}
\vspace{-0.2cm}
\caption{Analysis results of our method in the semantic segmentation task (Fishyscapes Lost\&Found validation set) 
(a): Comparison of the class-wise total energy gap of PEBAL and our method; (b): class-wise total energy gap difference between PEBAL and our method.}
\end{figure}
\vspace{-0.1cm}

We define the novel term Energy Gap as $\mathbb{E}[E(x_{in})]-\mathbb{E}[E(x_{out})]$, which measures the average energy gap between ID data and OOD data. We can measure it class-wise and $i$ th class Energy Gap is $\mathbb{E}[E(x_{in,i})]-\mathbb{E}[E(x_{out,i})]$. Finally, the class-wise Total Energy Gap is defined as $(\mathbb{E}[E(x_{in,i})]-\mathbb{E}[E(x_{out,i})]) \cdot N_{out,i}$ by multiplying frequency of class $i$ OOD data. Intuitively, the larger the Total Energy Gap, the larger the energy gap between ID and OOD, and the better the OOD detection performance. Figure~\ref{anal_fig_top} compares the results of our method and PEBAL with regard to the class-wise Total Energy Gap. We see that this gap is elevated in the majority class, like the road.

Figure~\ref{anal_fig_down} shows the difference in Total Energy Gap between PEBAL and ours. 
As shown in Figure~\ref{anal_fig_down}, class-wise difference distribution shown in orange is similar to the prior probability shown in blue, which infers that our balanced energy regularization effectively works. Furthermore, by improving the gap on the majority class effectively, the sum of the class-wise Total Energy Gap increases (black dotted line is over 0). Thus, our method improves the OOD detection performance compared to PEBAL.
\vspace{-0.1cm}
\subsection{Ablation Study}
\vspace{-0.1cm}
\subsubsection{Hyperparameter Analysis}
\vspace{-0.1cm}
\begin{table}[t!]
\centering
\scriptsize
\captionsetup{font=footnotesize}
\caption{ Evaluation result in semantic segmentation task depending on $\gamma$ : OOD detection performance(AUROC,AP,FPR) and accuracy(MIOU for Cityscapes validation) on the Fishyscapes validation sets}
\vspace{-0.2cm}
\label{table_seg_gamma}
\begin{adjustbox}{width=0.5\textwidth}
\begin{tabular}{cc|c|ccc|ccc}
\hline
\multicolumn{2}{c|}{Method}                                                                                       & City           & \multicolumn{3}{c|}{FS Lost \& Found}                                                     & \multicolumn{3}{c}{FS Static}                                                             \\ \hline
\multicolumn{1}{c|}{Name}                                                                          &  $\gamma$       & MIOU$\uparrow$           & \multicolumn{1}{c|}{AUC$\uparrow$}            & \multicolumn{1}{c|}{AP$\uparrow$}             & FPR$\downarrow$           & \multicolumn{1}{c|}{AUC$\uparrow$}            & \multicolumn{1}{c|}{AP$\uparrow$}             & FPR$\downarrow$           \\ \hline
\multicolumn{1}{c|}{EnergyOE}                                                                      & 0.0          & 89.07          & \multicolumn{1}{c|}{98.14}          & \multicolumn{1}{c|}{45.61}          & 8.21          & \multicolumn{1}{c|}{99.32}          & \multicolumn{1}{c|}{89.12}          & 2.62          \\ \hline
\multicolumn{1}{c|}{\multirow{4}{*}{\begin{tabular}[c]{@{}c@{}}Balanced \\ EnergyOE\end{tabular}}} & 1.0          & 89.04          & \multicolumn{1}{c|}{98.01}          & \multicolumn{1}{c|}{50.42}          & 8.87          & \multicolumn{1}{c|}{99.34}          & \multicolumn{1}{c|}{89.28}          & 2.33          \\
\multicolumn{1}{c|}{}                                                                              & 2.0          & \textbf{89.83} & \multicolumn{1}{c|}{98.72}          & \multicolumn{1}{c|}{53.36}          & 6.57          & \multicolumn{1}{c|}{99.42}          & \multicolumn{1}{c|}{90.32}          & 2.31          \\
\multicolumn{1}{c|}{}                                                                              & \textbf{3.0} & 88.91          & \multicolumn{1}{c|}{98.42}          & \multicolumn{1}{c|}{\textbf{54.58}} & 6.70          & \multicolumn{1}{c|}{\textbf{99.43}} & \multicolumn{1}{c|}{\textbf{91.77}} & \textbf{1.63} \\
\multicolumn{1}{c|}{}                                                                              & 4.0          & 88.53          & \multicolumn{1}{c|}{\textbf{98.81}} & \multicolumn{1}{c|}{53.27}          & \textbf{5.18} & \multicolumn{1}{c|}{99.34}          & \multicolumn{1}{c|}{89.98}          & 2.39          \\ \hline
\multicolumn{1}{c|}{\begin{tabular}[c]{@{}c@{}}Inverse \\ Balanced\\  EnergyOE\end{tabular}}       & -3.0         & 84.28          & \multicolumn{1}{c|}{95.49}          & \multicolumn{1}{c|}{43.94}          & 31.28         & \multicolumn{1}{c|}{98.45}          & \multicolumn{1}{c|}{81.32}          & 5.66          \\ \hline
\end{tabular}
\end{adjustbox}
\end{table}

\begin{table}[t!]
\centering
\scriptsize
\captionsetup{font=footnotesize}
\caption{Evaluation result on CIFAR10-LT depending on $\gamma$ :average (over 6 datasets) OOD detection performance (AUROC,AP, FPR) and classification accuracy (ACC) with model ResNet18; Mean over six random runs are reported.}
\vspace{-0.2cm}
\label{table_LT_cls_10_gamma}
\begin{tabular}{cc|c|ccc}
\hline
\multicolumn{2}{c|}{Method}            & \multirow{2}{*}{ACC$\uparrow$} & \multicolumn{3}{c}{\begin{tabular}[c]{@{}c@{}}Average\\ (total 6)\end{tabular}}            \\ \cline{1-2} \cline{4-6} 
\multicolumn{1}{c|}{Name}      & $\gamma$ &                      & \multicolumn{1}{c|}{AUC$\uparrow$}            & \multicolumn{1}{c|}{AP$\uparrow$}             & FPR$\downarrow$            \\ \hline
\multicolumn{1}{c|}{Energy OE} & 0.00  & 74.53                & \multicolumn{1}{c|}{91.92}          & \multicolumn{1}{c|}{91.97}          & 33.79          \\ \hline
\multicolumn{1}{c|}{\multirow{6}{*}{\begin{tabular}[c]{@{}c@{}}Balanced \\ Energy OE\end{tabular}}} &
  0.10 &
  75.03 &
  \multicolumn{1}{c|}{92.01} &
  \multicolumn{1}{c|}{91.57} &
  32.80 \\
\multicolumn{1}{c|}{}          & 0.25  & 75.23                & \multicolumn{1}{c|}{92.16}          & \multicolumn{1}{c|}{91.36}          & 31.83          \\
\multicolumn{1}{c|}{}          & 0.50  & 75.92                & \multicolumn{1}{c|}{92.44}          & \multicolumn{1}{c|}{91.67}          & 30.81          \\
\multicolumn{1}{c|}{}          & 0.75  & \textbf{76.22}       & \multicolumn{1}{c|}{\textbf{92.56}} & \multicolumn{1}{c|}{91.94}          & \textbf{30.60} \\
\multicolumn{1}{c|}{}          & 1.00  & 74.85                & \multicolumn{1}{c|}{92.45}          & \multicolumn{1}{c|}{\textbf{92.03}} & 31.86          \\
\multicolumn{1}{c|}{}          & 1.25  & 72.38                & \multicolumn{1}{c|}{92.33}          & \multicolumn{1}{c|}{\textbf{92.03}} & 32.60          \\ \hline
\multicolumn{1}{c|}{\begin{tabular}[c]{@{}c@{}}Inv-Balanced\\  Energy OE\end{tabular}} &
  -0.75 &
  64.24 &
  \multicolumn{1}{c|}{90.75} &
  \multicolumn{1}{c|}{90.83} &
  39.82 \\ \hline
\end{tabular}
\end{table}

As discussed in Section~\ref{section:regularization_loss}, the hyperparameter $\gamma$ controls the degree of prior difference between classes. We fix $\alpha$ as in Table~\ref{table_hyper} and find the best $\gamma$ starting from base case of $\gamma=0$. For accurate implementation of baseline EnergyOE, we only set $\alpha=0$, when $\gamma=0$.
For semantic segmentation task and long-tailed image classification task as in Table~\ref{table_seg_gamma} and Table~\ref{table_LT_cls_10_gamma}, we obtain a common tendency for $\gamma$. First, as $\gamma$ becomes larger than 0, the OOD detection performance and accuracy improves, then obtain an optimal value and decrease again.
Second, in the inverse case where $\gamma$ becomes smaller than 0, both OOD detection performance and accuracy are worse than the baseline. This would be evidence of the efficiency of prior probability.
\vspace{-0.1cm}
\subsubsection{Loss Component Analysis}
\vspace{-0.1cm}
\vspace{-0.3cm}
\begin{table}[ht!]
\centering
\scriptsize
\captionsetup{font=footnotesize}
\caption{Results of the loss component ablation on Fishyscapes validation sets in semantic segmentation task}
\vspace{-0.2cm}
\label{table_seg_elem}
\begin{tabular}{cc|ccc|ccc}
\hline
\multicolumn{2}{c|}{Loss Component} & \multicolumn{3}{c|}{FS Lost \& Found}                                   & \multicolumn{3}{c}{FS Static}                                  \\ \hline
Margin &
  Weight &
  \multicolumn{1}{c|}{AUC$\uparrow$} &
  \multicolumn{1}{c|}{AP$\uparrow$} &
  FPR$\downarrow$ &
  \multicolumn{1}{c|}{AUC$\uparrow$} &
  \multicolumn{1}{c|}{AP$\uparrow$} &
  FPR$\downarrow$ \\ \hline
\ding{55}           & \ding{55}             & \multicolumn{1}{c|}{98.14}          & \multicolumn{1}{c|}{45.61} & 8.21 & \multicolumn{1}{c|}{99.32} & \multicolumn{1}{c|}{89.12} & 2.62 \\
\ding{55}             & \ding{51}            & \multicolumn{1}{c|}{98.31}          & \multicolumn{1}{c|}{47.10} & 7.04 & \multicolumn{1}{c|}{99.34} & \multicolumn{1}{c|}{89.73} & 2.31 \\
\ding{51}            & \ding{55}             & \multicolumn{1}{c|}{\textbf{98.46}} & \multicolumn{1}{c|}{51.65} & 7.17 & \multicolumn{1}{c|}{99.41} & \multicolumn{1}{c|}{90.14} & 2.23 \\
\ding{51} &
  \ding{51} &
  \multicolumn{1}{c|}{98.42} &
  \multicolumn{1}{c|}{\textbf{54.58}} &
  \textbf{6.70} &
  \multicolumn{1}{c|}{\textbf{99.43}} &
  \multicolumn{1}{c|}{\textbf{91.77}} &
  \textbf{1.63} \\ \hline
\end{tabular}
\end{table}
\vspace{-0.3cm}

\begin{table}[ht!]
\centering
\scriptsize
\captionsetup{font=footnotesize}
\caption{Results of the loss component ablation on both CIFAR10-LT and CIFAR100-LT in long-tailed image classification task}
\vspace{-0.2cm}
\label{table_LT_cls_elem}
\begin{tabular}{cc|ccc|ccc}
\hline
\multicolumn{2}{c|}{Loss Component} &
  \multicolumn{3}{c|}{\begin{tabular}[c]{@{}c@{}}CIFAR10-LT\\ Average\end{tabular}} &
  \multicolumn{3}{c}{\begin{tabular}[c]{@{}c@{}}CIFAR100-LT\\ Average\end{tabular}} \\ \hline
Margin &
  Weight &
  \multicolumn{1}{c|}{AUC$\uparrow$} &
  \multicolumn{1}{c|}{AP$\uparrow$} &
  FPR$\downarrow$ &
  \multicolumn{1}{c|}{AUC$\uparrow$} &
  \multicolumn{1}{c|}{AP$\uparrow$} &
  FPR$\downarrow$ \\ \hline
\ding{55}  & \ding{55}  & \multicolumn{1}{c|}{91.92} & \multicolumn{1}{c|}{91.97}          & 33.79 & \multicolumn{1}{c|}{76.40} & \multicolumn{1}{c|}{72.24} & 64.54 \\
\ding{55}  & \ding{51}  & \multicolumn{1}{c|}{92.32} & \multicolumn{1}{c|}{91.39}          & 31.00 & \multicolumn{1}{c|}{77.28} & \multicolumn{1}{c|}{72.20} & 61.92 \\
\ding{51}  & \ding{55} & \multicolumn{1}{c|}{92.32} & \multicolumn{1}{c|}{\textbf{92.27}} & 32.44 & \multicolumn{1}{c|}{77.17} & \multicolumn{1}{c|}{72.99} & 63.23 \\
\ding{51}  &
  \ding{51}  &
  \multicolumn{1}{c|}{\textbf{92.56}} &
  \multicolumn{1}{c|}{91.94} &
  \textbf{30.60} &
  \multicolumn{1}{c|}{\textbf{77.75}} &
  \multicolumn{1}{c|}{\textbf{73.10}} &
  \textbf{61.15} \\ \hline
\end{tabular}
\end{table}

As defined in Section~\ref{section:regularization_loss}, our balanced energy regularization loss has two adaptive loss components that depend on $Z_{\gamma}$. For the semantic segmentation task and long-tailed image classification task as in Table~\ref{table_seg_elem} and Table~\ref{table_LT_cls_elem}, ablation results on two loss components show that we can achieve the best performance when using both adaptive loss margin and adaptive loss weight.    
\vspace{-0.2cm}
\subsubsection{Network Analysis}
\vspace{-0.2cm}

\vspace{-0.3cm}

\begin{table}[ht!]
\captionsetup{font=footnotesize}
\caption{ Evaluation result on long-tailed CIFAR using WideResNet(WRN-40-2); average (over 6 datasets) OOD detection performance (AUROC,AP, FPR) and classification accuracy (ACC) (a): Result on CIFAR10-LT (b): Result on CIFAR100-LT}
\vspace{-0.2cm}
\begin{subtable}{1\linewidth}
\centering
\scriptsize
\caption{}
\label{wideres_cifar10}
\begin{tabular}{c|c|c|c|c|c}
\hline
Dataset                  & Method          & AUC$\uparrow$    & AP$\uparrow$      & FPR$\downarrow$        & ACC$\uparrow$               \\ \hline
\multirow{5}{*}{Average} & MSP~\cite{hendrycks2016baseline}(ST)        & 74.59          & 72.62          & 63.96          & 72.32          \\
                         & Energy~\cite{liu2020energy}(ST)     & 80.23          & 77.67          & 58.44          & 72.32          \\
                         & OE~\cite{hendrycks2018deep}(tune)       & 83.96          & 83.70          & 54.60          & 69.31          \\
                         & EnergyOE~\cite{liu2020energy}(tune) & 91.44          & \textbf{91.01} & 34.02          & 75.02          \\ \cline{2-6} 
                         &  \textbf{Ours}            & \textbf{91.85} & 90.48          & \textbf{31.03} & \textbf{76.14} \\ \hline
\end{tabular}

\vspace{0.1cm}
\caption{}
\begin{tabular}{c|c|c|c|c|c}
\hline
Dataset                  & Method          & AUC$\uparrow$    & AP$\uparrow$      & FPR$\downarrow$        & ACC$\uparrow$              \\ \hline
\multirow{5}{*}{Average} & MSP~\cite{hendrycks2016baseline}(ST)        & 60.24          & 57.22          & 83.52          & 40.96          \\
                         & Energy~\cite{liu2020energy}(ST)     & 63.31          & 59.44          & 81.95          & 40.96          \\
                         & OE~\cite{hendrycks2018deep}(tune)       & 68.52          & 65.18          & 76.88          & \textbf{41.73} \\
                         & EnergyOE~\cite{liu2020energy}(tune) & 76.45          & 72.75          & 65.70          & 39.95          \\ \cline{2-6} 
                         &  \textbf{Ours}            & \textbf{77.41} & \textbf{73.10} & \textbf{62.84} & 39.44          \\ \hline
\end{tabular}
\label{wideres_cifar100}
\end{subtable}
\end{table}
\vspace{-0.3cm}

To show the generality of our method, we perform a long-tailed image classification experiment on WideResNet (WRN-40-2)~\cite{zagoruyko2016wide} instead of ResNet18~\cite{he2016deep}. 
When $\gamma$ is 0.5 ($\alpha$ as in Table~\ref{table_hyper}), we get optimal performance in both CIFAR10-LT and CIFAR100-LT, and the results are as summarized in Table~\ref{wideres_cifar10} and Table~\ref{wideres_cifar100}, respectively. Our approach outperforms the current baselines (OE, EnergyOE) on both CIFAR10-LT and CIFAR100-LT, similar to the case with ResNet18.
\vspace{-0.3cm}

\section{Conclusion}
\vspace{-0.1cm}
To solve the OOD detection issue in various tasks, we propose a new balanced energy regularization loss. The main idea of our loss is to apply large regularization to auxiliary samples of majority classes, compared to those of minority. We show the effectiveness of our novel loss through extensive experiments on semantic segmentation, long-tailed image classification, and image classification datasets. {\bf Limitations and potential negative social impacts} are provided in the supplement. 
\vspace{-0.1cm}
\section*{Acknowledgements}
\vspace{-0.1cm}
\addcontentsline{toc}{section}{Acknowledgements}
\footnotesize This work was supported by IITP/MSIT [No.B0101-15-0266, Development of High Performance Visual BigData Discovery Platform for Large-Scale Realtime Data Analysis; No.2021-0-01343, Artificial Intelligence Graduate School Program (Seoul National University)], and KAIA/MOLIT [No.1615012984, Development of Automated Driving System (Lv.4/4+) based Car-Sharing Service Technologies].


{\small
\bibliographystyle{ieee_fullname}

}
\newpage
\appendix
\onecolumn

\renewcommand{\appendixname}{S}
\renewcommand{\thesection}{S}
\renewcommand\thefigure{\thesection.\arabic{figure}}
\renewcommand\thetable{\thesection.\arabic{table}}

\setcounter{figure}{0}
\setcounter{table}{0}

\renewcommand*{\thepage}{S\arabic{page}}
\setcounter{page}{1}
\begin{center}
\section*{Supplementary Material}
\end{center}

\subsection{Limitations}

Although our method is a simple and competitive approach to various OOD detection tasks, additional auxiliary data is still required.
OOD detection without additional data would be an ideal model. However, approaches that use auxiliary data still show superior performance to the approaches without using auxiliary data.
Instead, we can consider a way to utilize a small amount of auxiliary data as little as possible with competitive performance in future work.

\subsection{Potential Negative Societal Impacts}
OOD detection is a branch of anomaly detection.
In vision tasks, anomaly detection typically plays a positive role in strengthening security in surveillance systems.
However, indiscriminate abuse of automated surveillance systems can increase the surveillance of workers.
Therefore, when our method is applied to a real environment, we ensure ours to be used only for the purpose of enhancing human safety without infringing on human privacy.

\subsection{Further details of implementation}

     For Section~\ref{section:Implementation}, we add a bit more to our implementation, 
     In long-tailed image classification, fine-tuning-based methods  OE~\cite{hendrycks2018deep}, OECC~\cite{papadopoulos2021outlier}, EnergyOE~\cite{liu2020energy}, and ours are trained for 20 epochs on both CIFAR10-LT and CIFAR100-LT.
     For OECC, we set the coefficients to $\lambda_1 = \lambda_2 = 0.03$. 


    In image classification, the number of training epochs mostly follows the conventional method. Therefore, in OE it is set to 10 for both CIFAR10 and CIFAR100, and in OECC it is set to 15 for both.
    For OECC, we set the coefficients to $\lambda_1=0.07$ and $\lambda_2=0.05$.
    On the other hand, we found that the convergence of training is fast when using our balanced energy regularization loss, unlike the conventional method. Therefore, we set the number of training epochs to 8 for CIFAR10 and the number of training epochs to 4 for CIFAR100 to get the optimal OOD detection performance. 
    Therefore, we also report the results of EnergyOE with the same number of epochs as 8 for CIFAR10 and 4 for CIFAR100 for fair comparison with our loss.
    For reference, results of EnergyOE for training 10 epochs as in the original work are 96.56, 96.37, 15.27, and 93.29 for CIFAR10 in the order of AUROC, AP, FPR, and ACC.
    For CIFAR100, we get 85.56, 80.68, 43.64, and 74.92.
    Still, the comparison in Table~\ref{balanced_merge_table} confirms the superiority of our method.

\subsection{Details on the setting of hyperparameter $\alpha$}
    As discussed in Section~\ref{section:regularization_loss},
    we set the adaptive loss margin as $\alpha$$\cdot$$Z_{\gamma}$.
    More specifically, to standardize $\alpha$, we redesign $\alpha$ as a relative offset from the baseline which the case of $\gamma$=$0$.
    Thus, we set  $\alpha\approx\beta$$\cdot$$K$$\cdot$ ($m_{out}-m_{in}$), where $\beta$ is proportionality constant, $K$ is a number of classes and $m_{out}-m_{in}$ is energy distance between in-distribution and OOD, respectively. When $\gamma$=$0$, adaptive loss margin $\alpha$$\cdot$$Z_{\gamma}$ become $\beta$$\cdot$ ($m_{out}-m_{in}$), which is the relative offset proportional to $\beta$. 
     As $\beta$ increases, the sensitivity to $\gamma$ increases, and even a small change in $\gamma$ greatly changes the adaptive loss margin.
    Conversely, if we set $\beta$$\approx$$0$, the sensitivity to $\gamma$ becomes small, and even if $\gamma$ increases, the adaptive margin approaches 0 regardless of this, so there is no difference from the original energy regularization loss $L_{out,hinge}$.
    Finally, we choose to use the appropriate $\beta=0.05$ with corresponding $\alpha$ and adjust the hyperparameter $\gamma$.
    As seen in Table~\ref{alpha_AUROC_table} and Table~\ref{alpha_ACC_table}, AUROC and accuracy also show similar tendencies as discussed above.

\begin{table}[h]
\centering
\footnotesize
\caption{Average AUROC (over 6 datasets) on CIFAR10-LT using ResNet18 depending on hyperparameter $\gamma$ and $\alpha$}
\label{alpha_AUROC_table}
\begin{tabular}{|cccccccc|}
\hline
\multicolumn{8}{|c|}{Average AUROC} \\ \hline
\multicolumn{1}{|c|}{\multirow{4}{*}{Hyperparameter $\gamma$}} &
  \multicolumn{7}{c|}{Hyperparameter $\beta$} \\ \cline{2-8} 
\multicolumn{1}{|c|}{} &
  \multicolumn{1}{c|}{0.0025} &
  \multicolumn{1}{c|}{0.01} &
  \multicolumn{1}{c|}{0.025} &
  \multicolumn{1}{c|}{0.05} &
  \multicolumn{1}{c|}{0.1} &
  \multicolumn{1}{c|}{0.25} &
  0.5 \\ \cline{2-8} 
\multicolumn{1}{|c|}{} &
  \multicolumn{7}{c|}{Hyperparameter $\alpha$} \\ \cline{2-8} 
\multicolumn{1}{|c|}{} &
  \multicolumn{1}{c|}{0.5} &
  \multicolumn{1}{c|}{2} &
  \multicolumn{1}{c|}{5} &
  \multicolumn{1}{c|}{10} &
  \multicolumn{1}{c|}{20} &
  \multicolumn{1}{c|}{50} &
  100 \\ \hline
\multicolumn{1}{|c|}{0.1} &
  \multicolumn{1}{c|}{91.92} &
  \multicolumn{1}{c|}{91.93} &
  \multicolumn{1}{c|}{91.98} &
  \multicolumn{1}{c|}{92.01} &
  \multicolumn{1}{c|}{92.12} &
  \multicolumn{1}{c|}{92.02} &
  91.52 \\ \hline
\multicolumn{1}{|c|}{0.25} &
  \multicolumn{1}{c|}{91.96} &
  \multicolumn{1}{c|}{92.00} &
  \multicolumn{1}{c|}{92.07} &
  \multicolumn{1}{c|}{92.16} &
  \multicolumn{1}{c|}{92.28} &
  \multicolumn{1}{c|}{92.42} &
  91.69 \\ \hline
\multicolumn{1}{|c|}{0.5} &
  \multicolumn{1}{c|}{92.17} &
  \multicolumn{1}{c|}{92.20} &
  \multicolumn{1}{c|}{92.29} &
  \multicolumn{1}{c|}{92.44} &
  \multicolumn{1}{c|}{92.55} &
  \multicolumn{1}{c|}{92.11} &
  91.85 \\ \hline
\multicolumn{1}{|c|}{0.75} &
  \multicolumn{1}{c|}{92.34} &
  \multicolumn{1}{c|}{92.39} &
  \multicolumn{1}{c|}{92.48} &
  \multicolumn{1}{c|}{\textbf{92.56}} &
  \multicolumn{1}{c|}{92.50} &
  \multicolumn{1}{c|}{92.00} &
  91.82 \\ \hline
\multicolumn{1}{|c|}{1} &
  \multicolumn{1}{c|}{92.41} &
  \multicolumn{1}{c|}{92.47} &
  \multicolumn{1}{c|}{92.52} &
  \multicolumn{1}{c|}{92.45} &
  \multicolumn{1}{c|}{92.44} &
  \multicolumn{1}{c|}{91.91} &
  91.79 \\ \hline
\multicolumn{1}{|c|}{1.25} &
  \multicolumn{1}{c|}{92.39} &
  \multicolumn{1}{c|}{92.41} &
  \multicolumn{1}{c|}{92.37} &
  \multicolumn{1}{c|}{92.33} &
  \multicolumn{1}{c|}{92.27} &
  \multicolumn{1}{c|}{91.94} &
  91.61 \\ \hline
\end{tabular}
\end{table}

\begin{table}[h]
\centering
\footnotesize
\caption{Accuracy on CIFAR10-LT using ResNet18 depending on hyperparameter $\gamma$ and $\alpha$}
\label{alpha_ACC_table}
\begin{tabular}{|cccccccc|}
\hline
\multicolumn{8}{|c|}{Accuracy(ACC)} \\ \hline
\multicolumn{1}{|c|}{\multirow{4}{*}{Hyperparameter $\gamma$}} &
  \multicolumn{7}{c|}{Hyperparameter $\beta$} \\ \cline{2-8} 
\multicolumn{1}{|c|}{} &
  \multicolumn{1}{c|}{0.0025} &
  \multicolumn{1}{c|}{0.01} &
  \multicolumn{1}{c|}{0.025} &
  \multicolumn{1}{c|}{0.05} &
  \multicolumn{1}{c|}{0.1} &
  \multicolumn{1}{c|}{0.25} &
  0.5 \\ \cline{2-8} 
\multicolumn{1}{|c|}{} &
  \multicolumn{7}{c|}{Hyperparameter $\alpha$} \\ \cline{2-8} 
\multicolumn{1}{|c|}{} &
  \multicolumn{1}{c|}{0.5} &
  \multicolumn{1}{c|}{2} &
  \multicolumn{1}{c|}{5} &
  \multicolumn{1}{c|}{10} &
  \multicolumn{1}{c|}{20} &
  \multicolumn{1}{c|}{50} &
  100 \\ \hline
\multicolumn{1}{|c|}{0.1} &
  \multicolumn{1}{c|}{74.53} &
  \multicolumn{1}{c|}{74.60} &
  \multicolumn{1}{c|}{74.84} &
  \multicolumn{1}{c|}{75.03} &
  \multicolumn{1}{c|}{75.45} &
  \multicolumn{1}{c|}{73.71} &
  70.18 \\ \hline
\multicolumn{1}{|c|}{0.25} &
  \multicolumn{1}{c|}{74.57} &
  \multicolumn{1}{c|}{74.64} &
  \multicolumn{1}{c|}{74.86} &
  \multicolumn{1}{c|}{75.23} &
  \multicolumn{1}{c|}{75.95} &
  \multicolumn{1}{c|}{76.13} &
  63.15 \\ \hline
\multicolumn{1}{|c|}{0.5} &
  \multicolumn{1}{c|}{75.15} &
  \multicolumn{1}{c|}{75.23} &
  \multicolumn{1}{c|}{75.61} &
  \multicolumn{1}{c|}{75.92} &
  \multicolumn{1}{c|}{\textbf{76.60}} &
  \multicolumn{1}{c|}{68.03} &
  59.78 \\ \hline
\multicolumn{1}{|c|}{0.75} &
  \multicolumn{1}{c|}{75.88} &
  \multicolumn{1}{c|}{75.87} &
  \multicolumn{1}{c|}{76.29} &
  \multicolumn{1}{c|}{76.22} &
  \multicolumn{1}{c|}{74.87} &
  \multicolumn{1}{c|}{65.64} &
  56.92 \\ \hline
\multicolumn{1}{|c|}{1} &
  \multicolumn{1}{c|}{76.15} &
  \multicolumn{1}{c|}{75.92} &
  \multicolumn{1}{c|}{75.98} &
  \multicolumn{1}{c|}{74.85} &
  \multicolumn{1}{c|}{72.40} &
  \multicolumn{1}{c|}{62.86} &
  54.66 \\ \hline
\multicolumn{1}{|c|}{1.25} &
  \multicolumn{1}{c|}{75.94} &
  \multicolumn{1}{c|}{75.66} &
  \multicolumn{1}{c|}{74.61} &
  \multicolumn{1}{c|}{72.38} &
  \multicolumn{1}{c|}{69.96} &
  \multicolumn{1}{c|}{60.85} &
  52.43 \\ \hline
\end{tabular}
\end{table}

\newpage
\subsection{OOD prior probability for all tasks}

\subsubsection{Semantic Segmentation}
\begin{table}[h!]
\centering
\scriptsize
\captionsetup{font=footnotesize}
\caption{OOD prior probability in semantic segmentation task: cut-pasted OOD pixel inference result collected from 10000 synthesized scene images}
\label{table_prior_seg}
\begin{tabular}{|c|c|c|c|c|c|c|c|c|c|c|}
\hline
class index & 0   & 1      & 2     & 3   & 4     & 5   & 6     & 7          & 8       & 9 \\ \hline
class name  & road     & sidewalk & building & wall     & fence    & pole     & traffic light & traffic sign & vegetation & terrain  \\ \hline
probability & 1.53E-01 & 1.33E-02 & 1.07E-01 & 4.01E-02 & 3.39E-02 & 5.26E-03 & 1.06E-03      & 2.80E-02     & 8.02E-02   & 9.80E-02 \\ \hline
class index & 10  & 11     & 12    & 13  & 14    & 15  & 16    & 17         & 18      &   \\ \hline
class name  & sky & person & rider & car & truck & bus & train & motorcycle & bicycle &   \\ \hline
probability & 2.17E-03 & 1.53E-01 & 2.92E-02 & 8.71E-02 & 1.20E-01 & 1.60E-02 & 7.07E-03      & 1.70E-02     & 9.39E-03   &          \\ \hline
\end{tabular}
\end{table}
\subsubsection{Long-tailed Image Classification}
\begin{table}[h!]
\centering
\scriptsize
\captionsetup{font=footnotesize}
\caption{OOD prior probability in long-tailed image classification task: TinyImages 300K inference result on CIFAR10-LT model using ResNet18 : }
\label{table_prior_cifar10_LT}
\begin{tabular}{|c|c|c|c|c|c|c|c|c|c|c|}
\hline
class index & 0        & 1          & 2        & 3        & 4        & 5        & 6        & 7        & 8        & 9        \\ \hline
class name  & airplane & automobile & bird     & cat      & deer     & dog      & frog     & horse    & ship     & truck    \\ \hline
probability & 2.50E-01 & 7.96E-02   & 1.63E-01 & 2.49E-01 & 6.15E-02 & 8.50E-02 & 3.05E-02 & 2.32E-02 & 2.39E-02 & 3.45E-02 \\ \hline
\end{tabular}
\end{table}
\begin{table}[h!]
\centering
\scriptsize
\captionsetup{font=footnotesize}
\caption{OOD prior probability in long-tailed image classification task: TinyImages 300K inference result on CIFAR100-LT model using ResNet18}
\label{table_prior_cifar100_LT}
\begin{tabular}{|c|c|c|c|c|c|c|c|c|c|c|}
\hline
class index & 0                            & 1              & 2         & 3             & 4         & 5        & 6            & 7           & 8             & 9          \\ \hline
class name  & {\color[HTML]{292929} apple} & aquarium\_fish & baby      & bear          & beaver    & bed      & bee          & beetle      & bicycle       & bottle     \\ \hline
probability & 7.58E-03                     & 1.69E-02       & 6.16E-02  & 2.54E-02      & 1.78E-02  & 3.67E-02 & 2.81E-02     & 1.02E-02    & 1.87E-02      & 3.70E-02   \\ \hline
class index & 10                           & 11             & 12        & 13            & 14        & 15       & 16           & 17          & 18            & 19         \\ \hline
class name  & bowl                         & boy            & bridge    & bus           & butterfly & camel    & can          & castle      & caterpillar   & cattle     \\ \hline
probability & 2.98E-02                     & 6.17E-02       & 3.48E-02  & 4.20E-02      & 1.69E-02  & 1.61E-02 & 6.56E-02     & 2.40E-02    & 2.52E-02      & 1.70E-02   \\ \hline
class index & 20                           & 21             & 22        & 23            & 24        & 25       & 26           & 27          & 28            & 29         \\ \hline
class name  & chair                        & chimpanzee     & clock     & cloud         & cockroach & couch    & crab         & crocodile   & cup           & dinosaur   \\ \hline
probability & 7.81E-03                     & 7.30E-03       & 2.55E-02  & 1.84E-02      & 3.27E-03  & 2.78E-02 & 9.09E-03     & 1.17E-02    & 7.72E-03      & 9.21E-03   \\ \hline
class index & 30                           & 31             & 32        & 33            & 34        & 35       & 36           & 37          & 38            & 39         \\ \hline
class name  & dolphin                      & elephant       & flatfish  & forest        & fox       & girl     & hamster      & house       & kangaroo      & keyboard   \\ \hline
probability & 8.08E-03                     & 7.00E-03       & 9.30E-03  & 2.38E-02      & 3.79E-03  & 1.99E-02 & 4.01E-03     & 1.64E-02    & 6.19E-03      & 1.74E-02   \\ \hline
class index & 40                           & 41             & 42        & 43            & 44        & 45       & 46           & 47          & 48            & 49         \\ \hline
class name  & lamp                         & lawn\_mower    & leopard   & lion          & lizard    & lobster  & man          & maple\_tree & motorcycle    & mountain   \\ \hline
probability & 1.59E-02                     & 6.44E-03       & 7.91E-03  & 2.53E-03      & 6.93E-03  & 7.94E-03 & 1.97E-02     & 4.40E-03    & 3.35E-03      & 8.74E-03   \\ \hline
class index & 50                           & 51             & 52        & 53            & 54        & 55       & 56           & 57          & 58            & 59         \\ \hline
class name  & mouse                        & mushroom       & oak\_tree & orange        & orchid    & otter    & palm\_tree   & pear        & pickup\_truck & pine\_tree \\ \hline
probability & 2.52E-03                     & 4.93E-03       & 1.88E-03  & 2.78E-03      & 5.78E-03  & 1.05E-03 & 2.04E-03     & 1.31E-03    & 2.52E-03      & 2.38E-03   \\ \hline
class index & 60                           & 61             & 62        & 63            & 64        & 65       & 66           & 67          & 68            & 69         \\ \hline
class name  & plain                        & plate          & poppy     & porcupine     & possum    & rabbit   & raccoon      & ray         & road          & rocket     \\ \hline
probability & 1.14E-02                     & 1.46E-02       & 3.08E-03  & 1.42E-03      & 4.73E-03  & 7.11E-03 & 3.10E-03     & 5.36E-03    & 4.87E-03      & 9.77E-03   \\ \hline
class index & 70                           & 71             & 72        & 73            & 74        & 75       & 76           & 77          & 78            & 79         \\ \hline
class name  & rose                         & sea            & seal      & shark         & shrew     & skunk    & skyscraper   & snail       & snake         & spider     \\ \hline
probability & 1.94E-03                     & 3.22E-03       & 4.47E-04  & 9.43E-04      & 2.63E-04  & 3.23E-04 & 2.14E-03     & 1.33E-04    & 1.59E-03      & 8.27E-04   \\ \hline
class index & 80                           & 81             & 82        & 83            & 84        & 85       & 86           & 87          & 88            & 89         \\ \hline
class name  & squirrel                     & streetcar      & sunflower & sweet\_pepper & table     & tank     & telephone    & television  & tiger         & tractor    \\ \hline
probability & 2.10E-04                     & 1.89E-03       & 1.05E-03  & 4.87E-04      & 1.35E-03  & 3.57E-04 & 1.10E-03     & 3.32E-03    & 4.07E-04      & 2.84E-03   \\ \hline
class index & 90                           & 91             & 92        & 93            & 94        & 95       & 96           & 97          & 98            & 99         \\ \hline
class name  & train                        & trout          & tulip     & turtle        & wardrobe  & whale    & willow\_tree & wolf        & woman         & worm       \\ \hline
probability & 7.27E-04                     & 7.00E-05       & 9.33E-05  & 2.00E-05      & 2.87E-03  & 1.60E-04 & 2.43E-04     & 5.33E-05    & 4.03E-04      & 9.00E-05   \\ \hline
\end{tabular}
\end{table}
\newpage
\subsubsection{Image Classification}

\begin{table}[h!]
\centering
\scriptsize
\captionsetup{font=footnotesize}
\caption{OOD prior probability in image classification task: TinyImages 300K inference result on CIFAR10 model using ResNet18 }
\label{table_prior_cifar10}
\begin{tabular}{|c|c|c|c|c|c|c|c|c|c|c|}
\hline
class index & 0        & 1          & 2        & 3        & 4        & 5        & 6        & 7        & 8        & 9        \\ \hline
class name  & airplane & automobile & bird     & cat      & deer     & dog      & frog     & horse    & ship     & truck    \\ \hline
probability & 1.15E-01 & 3.62E-02   & 9.94E-02 & 2.04E-01 & 3.58E-02 & 1.19E-01 & 7.51E-02 & 5.35E-02 & 9.91E-02 & 1.62E-01 \\ \hline
\end{tabular}
\end{table}
\begin{table}[h!]
\centering
\scriptsize
\captionsetup{font=footnotesize}
\caption{OOD prior probability in image classification task: TinyImages 300K inference result on CIFAR100 model using ResNet18.}
\label{table_prior_cifar100}
\begin{tabular}{|c|c|c|c|c|c|c|c|c|c|c|}
\hline
class index & 0                            & 1              & 2         & 3             & 4         & 5        & 6            & 7           & 8             & 9          \\ \hline
class name  & {\color[HTML]{292929} apple} & aquarium\_fish & baby      & bear          & beaver    & bed      & bee          & beetle      & bicycle       & bottle     \\ \hline
probability & 1.39E-03                     & 1.60E-03       & 1.50E-02  & 6.34E-03      & 2.66E-03  & 1.15E-02 & 3.01E-03     & 2.50E-03    & 5.32E-03      & 1.88E-02   \\ \hline
class index & 10                           & 11             & 12        & 13            & 14        & 15       & 16           & 17          & 18            & 19         \\ \hline
class name  & bowl                         & boy            & bridge    & bus           & butterfly & camel    & can          & castle      & caterpillar   & cattle     \\ \hline
probability & 1.98E-02                     & 1.97E-02       & 1.69E-02  & 1.36E-02      & 4.82E-03  & 5.97E-03 & 4.14E-02     & 1.24E-02    & 9.66E-03      & 1.01E-02   \\ \hline
class index & 20                           & 21             & 22        & 23            & 24        & 25       & 26           & 27          & 28            & 29         \\ \hline
class name  & chair                        & chimpanzee     & clock     & cloud         & cockroach & couch    & crab         & crocodile   & cup           & dinosaur   \\ \hline
probability & 7.11E-03                     & 4.63E-03       & 2.11E-02  & 1.93E-02      & 2.88E-03  & 2.62E-02 & 4.44E-03     & 4.88E-03    & 7.81E-03      & 3.48E-03   \\ \hline
class index & 30                           & 31             & 32        & 33            & 34        & 35       & 36           & 37          & 38            & 39         \\ \hline
class name  & dolphin                      & elephant       & flatfish  & forest        & fox       & girl     & hamster      & house       & kangaroo      & keyboard   \\ \hline
probability & 1.46E-03                     & 5.00E-03       & 1.37E-02  & 2.69E-02      & 2.08E-03  & 1.68E-02 & 1.25E-03     & 2.83E-02    & 4.45E-03      & 2.18E-02   \\ \hline
class index & 40                           & 41             & 42        & 43            & 44        & 45       & 46           & 47          & 48            & 49         \\ \hline
class name  & lamp                         & lawn\_mower    & leopard   & lion          & lizard    & lobster  & man          & maple\_tree & motorcycle    & mountain   \\ \hline
probability & 2.38E-02                     & 5.93E-03       & 3.82E-03  & 9.67E-04      & 9.30E-03  & 1.61E-02 & 7.35E-02     & 2.05E-03    & 8.32E-03      & 6.84E-03   \\ \hline
class index & 50                           & 51             & 52        & 53            & 54        & 55       & 56           & 57          & 58            & 59         \\ \hline
class name  & mouse                        & mushroom       & oak\_tree & orange        & orchid    & otter    & palm\_tree   & pear        & pickup\_truck & pine\_tree \\ \hline
probability & 3.60E-03                     & 6.78E-03       & 8.63E-04  & 3.11E-03      & 1.05E-02  & 5.41E-03 & 3.04E-03     & 5.48E-03    & 6.31E-03      & 9.94E-03   \\ \hline
class index & 60                           & 61             & 62        & 63            & 64        & 65       & 66           & 67          & 68            & 69         \\ \hline
class name  & plain                        & plate          & poppy     & porcupine     & possum    & rabbit   & raccoon      & ray         & road          & rocket     \\ \hline
probability & 1.14E-02                     & 1.46E-02       & 3.08E-03  & 1.42E-03      & 4.73E-03  & 7.11E-03 & 3.10E-03     & 5.36E-03    & 4.87E-03      & 9.77E-03   \\ \hline
class index & 70                           & 71             & 72        & 73            & 74        & 75       & 76           & 77          & 78            & 79         \\ \hline
class name  & rose                         & sea            & seal      & shark         & shrew     & skunk    & skyscraper   & snail       & snake         & spider     \\ \hline
probability & 4.86E-03                     & 1.47E-02       & 4.44E-03  & 5.88E-03      & 1.82E-03  & 2.13E-03 & 8.50E-03     & 2.21E-03    & 8.53E-03      & 5.92E-03   \\ \hline
class index & 80                           & 81             & 82        & 83            & 84        & 85       & 86           & 87          & 88            & 89         \\ \hline
class name  & squirrel                     & streetcar      & sunflower & sweet\_pepper & table     & tank     & telephone    & television  & tiger         & tractor    \\ \hline
probability & 4.27E-03                     & 1.23E-02       & 3.71E-03  & 9.63E-03      & 3.25E-02  & 1.06E-02 & 1.38E-02     & 2.93E-02    & 3.31E-03      & 1.12E-02   \\ \hline
class index & 90                           & 91             & 92        & 93            & 94        & 95       & 96           & 97          & 98            & 99         \\ \hline
class name  & train                        & trout          & tulip     & turtle        & wardrobe  & whale    & willow\_tree & wolf        & woman         & worm       \\ \hline
probability & 8.04E-03                     & 2.99E-03       & 6.48E-03  & 2.52E-03      & 2.10E-02  & 2.42E-03 & 9.76E-03     & 3.52E-03    & 2.64E-02      & 1.00E-02   \\ \hline
\end{tabular}
\end{table}

\subsection{Detailed Experiment Results}
\subsubsection{Semantic Segmentation}
\begin{table}[h!]
\centering
\footnotesize
\captionsetup{font=footnotesize}
\caption{ Detailed evaluation result in semantic segmentation task depending on $\gamma$ : OOD detection performance(AUROC,AP,FPR) and accuracy(MIOU for Cityscapes validation) on the Fishyscapes validation sets and Road Anomaly test set}
\label{table_seg_gamma_for_all}
\begin{tabular}{cc|c|ccc|ccc|ccc}
\hline
\multicolumn{2}{c|}{Method} &
  City &
  \multicolumn{3}{c|}{FS Lost \& Found} &
  \multicolumn{3}{c|}{FS Static} &
  \multicolumn{3}{c}{Road Anomaly} \\ \hline
\multicolumn{1}{c|}{Name} &
  $\gamma$ &
  MIOU$\uparrow$ &
  \multicolumn{1}{c|}{AUC$\uparrow$} &
  \multicolumn{1}{c|}{AP$\uparrow$} &
  FPR$\downarrow$ &
  \multicolumn{1}{c|}{AUC$\uparrow$} &
  \multicolumn{1}{c|}{AP$\uparrow$} &
  FPR$\downarrow$ &
  \multicolumn{1}{c|}{AUC$\uparrow$} &
  \multicolumn{1}{c|}{AP$\uparrow$} &
  FPR$\downarrow$ \\ \hline
\multicolumn{1}{c|}{EnergyOE} &
  0.0 &
  89.07 &
  \multicolumn{1}{c|}{98.14} &
  \multicolumn{1}{c|}{45.61} &
  8.21 &
  \multicolumn{1}{c|}{99.32} &
  \multicolumn{1}{c|}{89.12} &
  2.62 &
  \multicolumn{1}{c|}{83.32} &
  \multicolumn{1}{c|}{32.59} &
  53.01 \\ \hline
\multicolumn{1}{c|}{\multirow{4}{*}{\begin{tabular}[c]{@{}c@{}}Balanced \\ EnergyOE\end{tabular}}} &
  1.0 &
  89.04 &
  \multicolumn{1}{c|}{98.01} &
  \multicolumn{1}{c|}{50.42} &
  8.87 &
  \multicolumn{1}{c|}{99.34} &
  \multicolumn{1}{c|}{89.28} &
  2.33 &
  \multicolumn{1}{c|}{83.61} &
  \multicolumn{1}{c|}{30.73} &
  49.96 \\
\multicolumn{1}{c|}{} &
  2.0 &
  \textbf{89.83} &
  \multicolumn{1}{c|}{98.72} &
  \multicolumn{1}{c|}{53.36} &
  6.57 &
  \multicolumn{1}{c|}{99.42} &
  \multicolumn{1}{c|}{90.32} &
  2.31 &
  \multicolumn{1}{c|}{84.90} &
  \multicolumn{1}{c|}{33.72} &
  \textbf{46.39} \\
\multicolumn{1}{c|}{} &
  3.0 &
  88.91 &
  \multicolumn{1}{c|}{98.42} &
  \multicolumn{1}{c|}{\textbf{54.58}} &
  6.70 &
  \multicolumn{1}{c|}{\textbf{99.43}} &
  \multicolumn{1}{c|}{\textbf{91.77}} &
  \textbf{1.63} &
  \multicolumn{1}{c|}{\textbf{85.50}} &
  \multicolumn{1}{c|}{\textbf{34.90}} &
  46.60 \\
\multicolumn{1}{c|}{} &
  4.0 &
  88.53 &
  \multicolumn{1}{c|}{\textbf{98.81}} &
  \multicolumn{1}{c|}{53.27} &
  \textbf{5.18} &
  \multicolumn{1}{c|}{99.34} &
  \multicolumn{1}{c|}{89.98} &
  2.39 &
  \multicolumn{1}{c|}{83.53} &
  \multicolumn{1}{c|}{29.51} &
  46.91 \\ \hline
\multicolumn{1}{c|}{\begin{tabular}[c]{@{}c@{}}Inverse \\ Balanced\\  EnergyOE\end{tabular}} &
  -3.0 &
  84.28 &
  \multicolumn{1}{c|}{95.49} &
  \multicolumn{1}{c|}{43.94} &
  31.28 &
  \multicolumn{1}{c|}{98.45} &
  \multicolumn{1}{c|}{81.32} &
  5.66 &
  \multicolumn{1}{c|}{84.92} &
  \multicolumn{1}{c|}{41.17} &
  55.57 \\ \hline
\end{tabular}
\end{table}
\newpage

\subsubsection{Long-tailed Image Classification}
{\bf\noindent ResNet18 model on CIFAR10-LT:}
\begin{table}[h!]
\captionsetup{font=footnotesize}
\caption{Detailed evaluation result on CIFAR10-LT depending on $\gamma$ : OOD detection performance (AUROC,AP, FPR) and classification accuracy (ACC) with model ResNet18; Mean over six random runs are reported; (a): ACC and result on Texture, SVHN, and CIFAR100; (b): Total average result and result on Tiny Imagenet, LSUN, and Place365}
\vspace{-0.2cm}
\begin{subtable}{1\linewidth}
\centering
\scriptsize
\caption{}
\begin{tabular}{cc|c|ccc|ccc|ccc}
\hline
\multicolumn{2}{c|}{Method} &
  \multirow{2}{*}{ACC$\uparrow$} &
  \multicolumn{3}{c|}{Texture} &
  \multicolumn{3}{c|}{SVHN} &
  \multicolumn{3}{c}{CIFAR 100} \\ \cline{1-2} \cline{4-12} 
\multicolumn{1}{c|}{Name} &
  $\gamma$ &
   &
  \multicolumn{1}{c|}{AUC$\uparrow$} &
  \multicolumn{1}{c|}{AP$\uparrow$} &
  FPR$\downarrow$ &
  \multicolumn{1}{c|}{AUC$\uparrow$} &
  \multicolumn{1}{c|}{AP$\uparrow$} &
  FPR$\downarrow$ &
  \multicolumn{1}{c|}{AUC$\uparrow$} &
  \multicolumn{1}{c|}{AP$\uparrow$} &
  FPR$\downarrow$ \\ \hline
\multicolumn{1}{c|}{Energy OE} &
  0.00 &
  74.53 &
  \multicolumn{1}{c|}{95.53} &
  \multicolumn{1}{c|}{92.93} &
  23.26 &
  \multicolumn{1}{c|}{96.63} &
  \multicolumn{1}{c|}{98.46} &
  14.52 &
  \multicolumn{1}{c|}{84.44} &
  \multicolumn{1}{c|}{84.63} &
  59.92 \\ \hline
\multicolumn{1}{c|}{\multirow{6}{*}{\begin{tabular}[c]{@{}c@{}}Balanced \\ Energy OE\end{tabular}}} &
  0.10 &
  75.03 &
  \multicolumn{1}{c|}{95.49} &
  \multicolumn{1}{c|}{91.94} &
  22.44 &
  \multicolumn{1}{c|}{96.68} &
  \multicolumn{1}{c|}{97.90} &
  12.64 &
  \multicolumn{1}{c|}{84.63} &
  \multicolumn{1}{c|}{84.42} &
  59.16 \\
\multicolumn{1}{c|}{} &
  0.25 &
  75.23 &
  \multicolumn{1}{c|}{95.55} &
  \multicolumn{1}{c|}{91.63} &
  22.03 &
  \multicolumn{1}{c|}{96.99} &
  \multicolumn{1}{c|}{98.06} &
  11.82 &
  \multicolumn{1}{c|}{84.89} &
  \multicolumn{1}{c|}{84.40} &
  57.80 \\
\multicolumn{1}{c|}{} &
  0.50 &
  75.92 &
  \multicolumn{1}{c|}{95.70} &
  \multicolumn{1}{c|}{92.05} &
  21.13 &
  \multicolumn{1}{c|}{97.49} &
  \multicolumn{1}{c|}{98.63} &
  10.48 &
  \multicolumn{1}{c|}{85.18} &
  \multicolumn{1}{c|}{84.78} &
  57.09 \\
\multicolumn{1}{c|}{} &
  0.75 &
  \textbf{76.22} &
  \multicolumn{1}{c|}{95.69} &
  \multicolumn{1}{c|}{92.38} &
  21.26 &
  \multicolumn{1}{c|}{97.74} &
  \multicolumn{1}{c|}{98.89} &
  9.87 &
  \multicolumn{1}{c|}{85.20} &
  \multicolumn{1}{c|}{84.98} &
  57.95 \\
\multicolumn{1}{c|}{} &
  1.00 &
  74.85 &
  \multicolumn{1}{c|}{95.56} &
  \multicolumn{1}{c|}{92.57} &
  22.57 &
  \multicolumn{1}{c|}{97.94} &
  \multicolumn{1}{c|}{99.03} &
  9.11 &
  \multicolumn{1}{c|}{84.84} &
  \multicolumn{1}{c|}{84.90} &
  61.18 \\
\multicolumn{1}{c|}{} &
  1.25 &
  72.38 &
  \multicolumn{1}{c|}{95.42} &
  \multicolumn{1}{c|}{92.67} &
  23.92 &
  \multicolumn{1}{c|}{98.10} &
  \multicolumn{1}{c|}{99.09} &
  8.86 &
  \multicolumn{1}{c|}{84.45} &
  \multicolumn{1}{c|}{84.61} &
  62.15 \\ \hline
\multicolumn{1}{c|}{\begin{tabular}[c]{@{}c@{}}Inv-Balanced\\  Energy OE\end{tabular}} &
  -0.75 &
  64.24 &
  \multicolumn{1}{c|}{95.19} &
  \multicolumn{1}{c|}{92.14} &
  27.52 &
  \multicolumn{1}{c|}{96.59} &
  \multicolumn{1}{c|}{98.48} &
  18.89 &
  \multicolumn{1}{c|}{81.76} &
  \multicolumn{1}{c|}{81.91} &
  63.45 \\ \hline
\end{tabular}

\caption{}
\begin{tabular}{c|ccc|ccc|ccc|ccc}
\hline
Method &
  \multicolumn{3}{c|}{Tiny ImageNet} &
  \multicolumn{3}{c|}{LSUN} &
  \multicolumn{3}{c|}{Place365} &
  \multicolumn{3}{c}{Average} \\ \hline
$\gamma$ &
  \multicolumn{1}{c|}{AUC$\uparrow$} &
  \multicolumn{1}{c|}{AP$\uparrow$} &
  FPR$\downarrow$ &
  \multicolumn{1}{c|}{AUC$\uparrow$} &
  \multicolumn{1}{c|}{AP$\uparrow$} &
  FPR$\downarrow$ &
  \multicolumn{1}{c|}{AUC$\uparrow$} &
  \multicolumn{1}{c|}{AP$\uparrow$} &
  FPR$\downarrow$ &
  \multicolumn{1}{c|}{AUC$\uparrow$} &
  \multicolumn{1}{c|}{AP$\uparrow$} &
  FPR$\downarrow$ \\ \hline
0.00 &
  \multicolumn{1}{c|}{88.40} &
  \multicolumn{1}{c|}{84.95} &
  45.17 &
  \multicolumn{1}{c|}{94.00} &
  \multicolumn{1}{c|}{93.70} &
  26.96 &
  \multicolumn{1}{c|}{92.51} &
  \multicolumn{1}{c|}{97.14} &
  32.88 &
  \multicolumn{1}{c|}{91.92} &
  \multicolumn{1}{c|}{91.97} &
  33.79 \\ \hline
0.10 &
  \multicolumn{1}{c|}{88.51} &
  \multicolumn{1}{c|}{84.58} &
  44.56 &
  \multicolumn{1}{c|}{94.12} &
  \multicolumn{1}{c|}{93.49} &
  26.36 &
  \multicolumn{1}{c|}{92.67} &
  \multicolumn{1}{c|}{97.10} &
  31.61 &
  \multicolumn{1}{c|}{92.01} &
  \multicolumn{1}{c|}{91.57} &
  32.80 \\
0.25 &
  \multicolumn{1}{c|}{88.63} &
  \multicolumn{1}{c|}{84.37} &
  43.29 &
  \multicolumn{1}{c|}{94.12} &
  \multicolumn{1}{c|}{92.72} &
  25.52 &
  \multicolumn{1}{c|}{92.79} &
  \multicolumn{1}{c|}{96.98} &
  30.54 &
  \multicolumn{1}{c|}{92.16} &
  \multicolumn{1}{c|}{91.36} &
  31.83 \\
0.50 &
  \multicolumn{1}{c|}{88.85} &
  \multicolumn{1}{c|}{84.71} &
  42.46 &
  \multicolumn{1}{c|}{94.30} &
  \multicolumn{1}{c|}{92.80} &
  24.62 &
  \multicolumn{1}{c|}{93.10} &
  \multicolumn{1}{c|}{97.08} &
  29.11 &
  \multicolumn{1}{c|}{92.44} &
  \multicolumn{1}{c|}{91.67} &
  30.81 \\
0.75 &
  \multicolumn{1}{c|}{88.92} &
  \multicolumn{1}{c|}{84.98} &
  42.38 &
  \multicolumn{1}{c|}{94.48} &
  \multicolumn{1}{c|}{93.15} &
  23.88 &
  \multicolumn{1}{c|}{93.35} &
  \multicolumn{1}{c|}{97.23} &
  28.25 &
  \multicolumn{1}{c|}{\textbf{92.56}} &
  \multicolumn{1}{c|}{91.94} &
  \textbf{30.60} \\
1.00 &
  \multicolumn{1}{c|}{88.74} &
  \multicolumn{1}{c|}{85.06} &
  43.11 &
  \multicolumn{1}{c|}{94.36} &
  \multicolumn{1}{c|}{93.35} &
  25.71 &
  \multicolumn{1}{c|}{93.26} &
  \multicolumn{1}{c|}{97.27} &
  29.46 &
  \multicolumn{1}{c|}{92.45} &
  \multicolumn{1}{c|}{\textbf{92.03}} &
  31.86 \\
1.25 &
  \multicolumn{1}{c|}{88.47} &
  \multicolumn{1}{c|}{84.95} &
  44.07 &
  \multicolumn{1}{c|}{94.31} &
  \multicolumn{1}{c|}{93.56} &
  26.46 &
  \multicolumn{1}{c|}{93.24} &
  \multicolumn{1}{c|}{97.30} &
  30.15 &
  \multicolumn{1}{c|}{92.33} &
  \multicolumn{1}{c|}{\textbf{92.03}} &
  32.60 \\ \hline
-0.75 &
  \multicolumn{1}{c|}{86.70} &
  \multicolumn{1}{c|}{82.96} &
  50.86 &
  \multicolumn{1}{c|}{93.10} &
  \multicolumn{1}{c|}{92.86} &
  36.08 &
  \multicolumn{1}{c|}{91.14} &
  \multicolumn{1}{c|}{96.62} &
  42.15 &
  \multicolumn{1}{c|}{90.75} &
  \multicolumn{1}{c|}{90.83} &
  39.82 \\ \hline
\end{tabular}
\end{subtable}
\end{table}

\begin{table}[h!]
\captionsetup{font=footnotesize}
\caption{Detailed evaluation result on CIFAR10-LT depending on $\gamma$ : OOD detection performance (AUROC,AP, FPR) and classification accuracy (ACC) with model ResNet18; Std over six random runs are reported; (a): ACC and result on Texture, SVHN, and CIFAR100; (b): Total average result and result on Tiny Imagenet, LSUN, and Place365}
\vspace{-0.2cm}
\begin{subtable}{1\linewidth}
\centering
\scriptsize
\caption{}
\begin{tabular}{cc|l|lll|lll|lll}
\hline
\multicolumn{2}{c|}{Method} &
  \multicolumn{1}{c|}{\multirow{2}{*}{ACC$\uparrow$}} &
  \multicolumn{3}{c|}{Texture} &
  \multicolumn{3}{c|}{SVHN} &
  \multicolumn{3}{c}{CIFAR 100} \\ \cline{1-2} \cline{4-12} 
\multicolumn{1}{c|}{Name} &
  $\gamma$ &
  \multicolumn{1}{c|}{} &
  \multicolumn{1}{c|}{AUC$\uparrow$} &
  \multicolumn{1}{c|}{AP$\uparrow$} &
  \multicolumn{1}{c|}{FPR$\downarrow$} &
  \multicolumn{1}{c|}{AUC$\uparrow$} &
  \multicolumn{1}{c|}{AP$\uparrow$} &
  \multicolumn{1}{c|}{FPR$\downarrow$} &
  \multicolumn{1}{c|}{AUC$\uparrow$} &
  \multicolumn{1}{c|}{AP$\uparrow$} &
  \multicolumn{1}{c}{FPR$\downarrow$} \\ \hline
\multicolumn{1}{c|}{Energy OE} &
  0.00 &
  3.09E-02 &
  \multicolumn{1}{l|}{3.73E-03} &
  \multicolumn{1}{l|}{8.98E-03} &
  1.43E-01 &
  \multicolumn{1}{l|}{3.82E-02} &
  \multicolumn{1}{l|}{1.89E-02} &
  1.14E-01 &
  \multicolumn{1}{l|}{5.00E-03} &
  \multicolumn{1}{l|}{5.00E-03} &
  8.22E-02 \\ \hline
\multicolumn{1}{c|}{\multirow{6}{*}{\begin{tabular}[c]{@{}c@{}}Balanced \\ Energy OE\end{tabular}}} &
  0.10 &
  2.63E-02 &
  \multicolumn{1}{l|}{5.00E-03} &
  \multicolumn{1}{l|}{1.11E-02} &
  8.76E-02 &
  \multicolumn{1}{l|}{2.97E-02} &
  \multicolumn{1}{l|}{1.89E-02} &
  1.52E-01 &
  \multicolumn{1}{l|}{7.64E-03} &
  \multicolumn{1}{l|}{9.43E-03} &
  1.59E-01 \\
\multicolumn{1}{c|}{} &
  0.25 &
  4.75E-02 &
  \multicolumn{1}{l|}{6.87E-03} &
  \multicolumn{1}{l|}{1.49E-02} &
  8.96E-02 &
  \multicolumn{1}{l|}{1.97E-02} &
  \multicolumn{1}{l|}{1.73E-02} &
  1.03E-01 &
  \multicolumn{1}{l|}{1.00E-02} &
  \multicolumn{1}{l|}{1.15E-02} &
  8.42E-02 \\
\multicolumn{1}{c|}{} &
  0.50 &
  4.83E-02 &
  \multicolumn{1}{l|}{5.77E-03} &
  \multicolumn{1}{l|}{2.13E-02} &
  5.93E-02 &
  \multicolumn{1}{l|}{1.61E-02} &
  \multicolumn{1}{l|}{8.16E-03} &
  8.55E-02 &
  \multicolumn{1}{l|}{6.87E-03} &
  \multicolumn{1}{l|}{1.29E-02} &
  1.37E-01 \\
\multicolumn{1}{c|}{} &
  0.75 &
  3.73E-02 &
  \multicolumn{1}{l|}{8.98E-03} &
  \multicolumn{1}{l|}{2.71E-02} &
  1.30E-01 &
  \multicolumn{1}{l|}{9.57E-03} &
  \multicolumn{1}{l|}{6.87E-03} &
  1.15E-01 &
  \multicolumn{1}{l|}{5.77E-03} &
  \multicolumn{1}{l|}{1.29E-02} &
  1.79E-01 \\
\multicolumn{1}{c|}{} &
  1.00 &
  5.15E-02 &
  \multicolumn{1}{l|}{6.87E-03} &
  \multicolumn{1}{l|}{2.21E-02} &
  1.51E-01 &
  \multicolumn{1}{l|}{1.41E-02} &
  \multicolumn{1}{l|}{6.87E-03} &
  9.36E-02 &
  \multicolumn{1}{l|}{7.64E-03} &
  \multicolumn{1}{l|}{8.98E-03} &
  1.22E-01 \\
\multicolumn{1}{c|}{} &
  1.25 &
  4.12E-02 &
  \multicolumn{1}{l|}{1.07E-02} &
  \multicolumn{1}{l|}{2.69E-02} &
  1.13E-01 &
  \multicolumn{1}{l|}{2.52E-02} &
  \multicolumn{1}{l|}{1.49E-02} &
  9.93E-02 &
  \multicolumn{1}{l|}{8.98E-03} &
  \multicolumn{1}{l|}{9.43E-03} &
  1.28E-01 \\ \hline
\multicolumn{1}{c|}{\begin{tabular}[c]{@{}c@{}}Inv-Balanced\\  Energy OE\end{tabular}} &
  -0.75 &
  4.89E-02 &
  \multicolumn{1}{l|}{4.71E-03} &
  \multicolumn{1}{l|}{7.45E-03} &
  1.22E-01 &
  \multicolumn{1}{l|}{2.11E-02} &
  \multicolumn{1}{l|}{1.26E-02} &
  1.39E-01 &
  \multicolumn{1}{l|}{4.71E-03} &
  \multicolumn{1}{l|}{4.71E-03} &
  9.64E-02 \\ \hline
\end{tabular}

\caption{}
\begin{tabular}{c|lll|lll|lll|lll}
\hline
Method &
  \multicolumn{3}{c|}{Tiny ImageNet} &
  \multicolumn{3}{c|}{LSUN} &
  \multicolumn{3}{c|}{Place365} &
  \multicolumn{3}{c}{Average} \\ \hline
$\gamma$ &
  \multicolumn{1}{c|}{AUC$\uparrow$} &
  \multicolumn{1}{c|}{AP$\uparrow$} &
  \multicolumn{1}{c|}{FPR$\downarrow$} &
  \multicolumn{1}{c|}{AUC$\uparrow$} &
  \multicolumn{1}{c|}{AP$\uparrow$} &
  \multicolumn{1}{c|}{FPR$\downarrow$} &
  \multicolumn{1}{c|}{AUC$\uparrow$} &
  \multicolumn{1}{c|}{AP$\uparrow$} &
  \multicolumn{1}{c|}{FPR$\downarrow$} &
  \multicolumn{1}{c|}{AUC$\uparrow$} &
  \multicolumn{1}{c|}{AP$\uparrow$} &
  \multicolumn{1}{c}{FPR$\downarrow$} \\ \hline
0.00 &
  \multicolumn{1}{l|}{5.00E-03} &
  \multicolumn{1}{l|}{3.73E-03} &
  1.03E-01 &
  \multicolumn{1}{l|}{9.43E-03} &
  \multicolumn{1}{l|}{9.57E-03} &
  5.35E-02 &
  \multicolumn{1}{l|}{3.73E-03} &
  \multicolumn{1}{l|}{0.00E+00} &
  6.47E-02 &
  \multicolumn{1}{l|}{1.08E-02} &
  \multicolumn{1}{l|}{7.70E-03} &
  9.34E-02 \\ \hline
0.10 &
  \multicolumn{1}{l|}{9.57E-03} &
  \multicolumn{1}{l|}{6.87E-03} &
  1.46E-01 &
  \multicolumn{1}{l|}{3.73E-03} &
  \multicolumn{1}{l|}{8.98E-03} &
  9.69E-02 &
  \multicolumn{1}{l|}{5.00E-03} &
  \multicolumn{1}{l|}{5.00E-03} &
  6.59E-02 &
  \multicolumn{1}{l|}{1.01E-02} &
  \multicolumn{1}{l|}{1.00E-02} &
  1.18E-01 \\
0.25 &
  \multicolumn{1}{l|}{5.00E-03} &
  \multicolumn{1}{l|}{1.34E-02} &
  3.65E-02 &
  \multicolumn{1}{l|}{1.15E-02} &
  \multicolumn{1}{l|}{1.26E-02} &
  9.09E-02 &
  \multicolumn{1}{l|}{8.16E-03} &
  \multicolumn{1}{l|}{5.00E-03} &
  7.69E-02 &
  \multicolumn{1}{l|}{1.02E-02} &
  \multicolumn{1}{l|}{1.25E-02} &
  8.01E-02 \\
0.50 &
  \multicolumn{1}{l|}{3.73E-03} &
  \multicolumn{1}{l|}{8.98E-03} &
  5.21E-02 &
  \multicolumn{1}{l|}{6.87E-03} &
  \multicolumn{1}{l|}{1.21E-02} &
  6.88E-02 &
  \multicolumn{1}{l|}{7.45E-03} &
  \multicolumn{1}{l|}{4.71E-03} &
  5.98E-02 &
  \multicolumn{1}{l|}{7.80E-03} &
  \multicolumn{1}{l|}{1.14E-02} &
  7.71E-02 \\
0.75 &
  \multicolumn{1}{l|}{1.11E-02} &
  \multicolumn{1}{l|}{1.07E-02} &
  1.16E-01 &
  \multicolumn{1}{l|}{7.64E-03} &
  \multicolumn{1}{l|}{1.34E-02} &
  7.16E-02 &
  \multicolumn{1}{l|}{4.71E-03} &
  \multicolumn{1}{l|}{5.00E-03} &
  1.38E-01 &
  \multicolumn{1}{l|}{7.96E-03} &
  \multicolumn{1}{l|}{1.27E-02} &
  1.25E-01 \\
1.00 &
  \multicolumn{1}{l|}{7.64E-03} &
  \multicolumn{1}{l|}{1.70E-02} &
  4.16E-02 &
  \multicolumn{1}{l|}{1.29E-02} &
  \multicolumn{1}{l|}{1.89E-02} &
  1.59E-01 &
  \multicolumn{1}{l|}{4.71E-03} &
  \multicolumn{1}{l|}{5.00E-03} &
  4.74E-02 &
  \multicolumn{1}{l|}{8.99E-03} &
  \multicolumn{1}{l|}{1.31E-02} &
  1.03E-01 \\
1.25 &
  \multicolumn{1}{l|}{9.43E-03} &
  \multicolumn{1}{l|}{1.11E-02} &
  6.99E-02 &
  \multicolumn{1}{l|}{5.77E-03} &
  \multicolumn{1}{l|}{8.98E-03} &
  1.84E-01 &
  \multicolumn{1}{l|}{5.00E-03} &
  \multicolumn{1}{l|}{3.73E-03} &
  5.43E-02 &
  \multicolumn{1}{l|}{1.08E-02} &
  \multicolumn{1}{l|}{1.25E-02} &
  1.08E-01 \\ \hline
-0.75 &
  \multicolumn{1}{l|}{3.73E-03} &
  \multicolumn{1}{l|}{6.87E-03} &
  1.47E-01 &
  \multicolumn{1}{l|}{3.73E-03} &
  \multicolumn{1}{l|}{5.77E-03} &
  7.29E-02 &
  \multicolumn{1}{l|}{3.73E-03} &
  \multicolumn{1}{l|}{5.00E-03} &
  7.57E-02 &
  \multicolumn{1}{l|}{6.96E-03} &
  \multicolumn{1}{l|}{7.07E-03} &
  1.09E-01 \\ \hline
\end{tabular}
\end{subtable}
\end{table}

\newpage
{\bf\noindent ResNet18 model on CIFAR100-LT:}

\begin{table}[h!]
\centering
\scriptsize
\captionsetup{font=footnotesize}
\caption{Detailed evaluation result on CIFAR100-LT depending on $\gamma$ : OOD detection performance (AUROC,AP, FPR) and classification accuracy (ACC) with model ResNet18; Mean over six random runs are reported; (a): ACC and result on Texture, SVHN, and CIFAR10; (b): Total average result and result on Tiny Imagenet, LSUN, and Place365}
\vspace{-0.2cm}
\begin{subtable}{1\linewidth}
\centering
\scriptsize
\caption{}
\begin{tabular}{cc|c|ccc|ccc|ccc}
\hline
\multicolumn{2}{c|}{Method} &
  \multirow{2}{*}{ACC$\uparrow$} &
  \multicolumn{3}{c|}{Texture} &
  \multicolumn{3}{c|}{SVHN} &
  \multicolumn{3}{c}{CIFAR 10} \\ \cline{1-2} \cline{4-12} 
\multicolumn{1}{c|}{Name} &
  $\gamma$ &
   &
  \multicolumn{1}{c|}{AUC$\uparrow$} &
  \multicolumn{1}{c|}{AP$\uparrow$} &
  FPR$\downarrow$ &
  \multicolumn{1}{c|}{AUC$\uparrow$} &
  \multicolumn{1}{c|}{AP$\uparrow$} &
  FPR$\downarrow$ &
  \multicolumn{1}{c|}{AUC$\uparrow$} &
  \multicolumn{1}{c|}{AP$\uparrow$} &
  FPR$\downarrow$ \\ \hline
\multicolumn{1}{c|}{Energy OE} &
  0.00 &
  40.65 &
  \multicolumn{1}{c|}{79.56} &
  \multicolumn{1}{c|}{70.88} &
  68.60 &
  \multicolumn{1}{c|}{86.19} &
  \multicolumn{1}{c|}{91.74} &
  42.27 &
  \multicolumn{1}{c|}{61.15} &
  \multicolumn{1}{c|}{56.66} &
  82.60 \\ \hline
\multicolumn{1}{c|}{\multirow{6}{*}{\begin{tabular}[c]{@{}c@{}}Balanced \\ Energy OE\end{tabular}}} &
  0.10 &
  40.90 &
  \multicolumn{1}{c|}{79.78} &
  \multicolumn{1}{c|}{70.80} &
  68.01 &
  \multicolumn{1}{c|}{86.65} &
  \multicolumn{1}{c|}{91.93} &
  41.01 &
  \multicolumn{1}{c|}{61.29} &
  \multicolumn{1}{c|}{56.65} &
  82.19 \\
\multicolumn{1}{c|}{} &
  0.25 &
  41.50 &
  \multicolumn{1}{c|}{80.06} &
  \multicolumn{1}{c|}{69.91} &
  66.93 &
  \multicolumn{1}{c|}{86.91} &
  \multicolumn{1}{c|}{91.99} &
  39.46 &
  \multicolumn{1}{c|}{61.24} &
  \multicolumn{1}{c|}{56.51} &
  82.23 \\
\multicolumn{1}{c|}{} &
  0.50 &
  \textbf{41.64} &
  \multicolumn{1}{c|}{81.01} &
  \multicolumn{1}{c|}{70.87} &
  65.50 &
  \multicolumn{1}{c|}{87.45} &
  \multicolumn{1}{c|}{92.15} &
  36.93 &
  \multicolumn{1}{c|}{61.38} &
  \multicolumn{1}{c|}{55.73} &
  83.49 \\
\multicolumn{1}{c|}{} &
  0.75 &
  41.05 &
  \multicolumn{1}{c|}{82.10} &
  \multicolumn{1}{c|}{73.09} &
  64.19 &
  \multicolumn{1}{c|}{88.66} &
  \multicolumn{1}{c|}{92.88} &
  33.79 &
  \multicolumn{1}{c|}{59.40} &
  \multicolumn{1}{c|}{54.97} &
  85.16 \\
\multicolumn{1}{c|}{} &
  1.00 &
  38.62 &
  \multicolumn{1}{c|}{83.28} &
  \multicolumn{1}{c|}{75.67} &
  63.54 &
  \multicolumn{1}{c|}{89.68} &
  \multicolumn{1}{c|}{93.71} &
  32.64 &
  \multicolumn{1}{c|}{58.81} &
  \multicolumn{1}{c|}{54.60} &
  86.46 \\
\multicolumn{1}{c|}{} &
  1.25 &
  36.51 &
  \multicolumn{1}{c|}{84.33} &
  \multicolumn{1}{c|}{77.49} &
  62.14 &
  \multicolumn{1}{c|}{90.71} &
  \multicolumn{1}{c|}{94.48} &
  30.90 &
  \multicolumn{1}{c|}{58.29} &
  \multicolumn{1}{c|}{54.30} &
  87.42 \\ \hline
\multicolumn{1}{c|}{\begin{tabular}[c]{@{}c@{}}Inv-Balanced\\  Energy OE\end{tabular}} &
  -0.75 &
  34.77 &
  \multicolumn{1}{c|}{82.79} &
  \multicolumn{1}{c|}{77.96} &
  70.45 &
  \multicolumn{1}{c|}{83.12} &
  \multicolumn{1}{c|}{90.07} &
  51.32 &
  \multicolumn{1}{c|}{58.34} &
  \multicolumn{1}{c|}{54.32} &
  88.17 \\ \hline
\end{tabular}

\caption{}
\begin{tabular}{c|ccc|ccc|ccc|ccc}
\hline
Method &
  \multicolumn{3}{c|}{Tiny ImageNet} &
  \multicolumn{3}{c|}{LSUN} &
  \multicolumn{3}{c|}{Place365} &
  \multicolumn{3}{c}{Average} \\ \hline
$\gamma$ &
  \multicolumn{1}{c|}{AUC$\uparrow$} &
  \multicolumn{1}{c|}{AP$\uparrow$} &
  FPR$\downarrow$ &
  \multicolumn{1}{c|}{AUC$\uparrow$} &
  \multicolumn{1}{c|}{AP$\uparrow$} &
  FPR$\downarrow$ &
  \multicolumn{1}{c|}{AUC$\uparrow$} &
  \multicolumn{1}{c|}{AP$\uparrow$} &
  FPR$\downarrow$ &
  \multicolumn{1}{c|}{AUC$\uparrow$} &
  \multicolumn{1}{c|}{AP$\uparrow$} &
  FPR$\downarrow$ \\ \hline
0.00 &
  \multicolumn{1}{c|}{70.78} &
  \multicolumn{1}{c|}{55.90} &
  74.43 &
  \multicolumn{1}{c|}{81.61} &
  \multicolumn{1}{c|}{69.16} &
  57.37 &
  \multicolumn{1}{c|}{79.12} &
  \multicolumn{1}{c|}{89.09} &
  61.96 &
  \multicolumn{1}{c|}{76.40} &
  \multicolumn{1}{c|}{72.24} &
  64.54 \\ \hline
0.10 &
  \multicolumn{1}{c|}{70.89} &
  \multicolumn{1}{c|}{55.85} &
  74.23 &
  \multicolumn{1}{c|}{81.94} &
  \multicolumn{1}{c|}{69.37} &
  56.53 &
  \multicolumn{1}{c|}{79.42} &
  \multicolumn{1}{c|}{89.18} &
  60.93 &
  \multicolumn{1}{c|}{76.66} &
  \multicolumn{1}{c|}{72.29} &
  63.82 \\
0.25 &
  \multicolumn{1}{c|}{71.10} &
  \multicolumn{1}{c|}{55.85} &
  74.12 &
  \multicolumn{1}{c|}{82.35} &
  \multicolumn{1}{c|}{69.55} &
  55.50 &
  \multicolumn{1}{c|}{79.86} &
  \multicolumn{1}{c|}{89.31} &
  59.54 &
  \multicolumn{1}{c|}{76.92} &
  \multicolumn{1}{c|}{72.18} &
  62.96 \\
0.50 &
  \multicolumn{1}{c|}{71.36} &
  \multicolumn{1}{c|}{56.15} &
  73.81 &
  \multicolumn{1}{c|}{83.25} &
  \multicolumn{1}{c|}{70.48} &
  52.87 &
  \multicolumn{1}{c|}{80.65} &
  \multicolumn{1}{c|}{89.65} &
  58.09 &
  \multicolumn{1}{c|}{77.35} &
  \multicolumn{1}{c|}{72.50} &
  61.78 \\
0.75 &
  \multicolumn{1}{c|}{71.42} &
  \multicolumn{1}{c|}{56.52} &
  74.22 &
  \multicolumn{1}{c|}{83.83} &
  \multicolumn{1}{c|}{71.23} &
  52.04 &
  \multicolumn{1}{c|}{81.10} &
  \multicolumn{1}{c|}{89.94} &
  57.52 &
  \multicolumn{1}{c|}{77.75} &
  \multicolumn{1}{c|}{73.10} &
  \textbf{61.15} \\
1.00 &
  \multicolumn{1}{c|}{71.41} &
  \multicolumn{1}{c|}{56.76} &
  75.08 &
  \multicolumn{1}{c|}{83.76} &
  \multicolumn{1}{c|}{71.40} &
  53.46 &
  \multicolumn{1}{c|}{81.12} &
  \multicolumn{1}{c|}{90.09} &
  58.79 &
  \multicolumn{1}{c|}{\textbf{78.01}} &
  \multicolumn{1}{c|}{73.70} &
  61.66 \\
1.25 &
  \multicolumn{1}{c|}{71.38} &
  \multicolumn{1}{c|}{56.81} &
  75.18 &
  \multicolumn{1}{c|}{82.85} &
  \multicolumn{1}{c|}{70.92} &
  58.68 &
  \multicolumn{1}{c|}{80.17} &
  \multicolumn{1}{c|}{89.83} &
  64.16 &
  \multicolumn{1}{c|}{77.96} &
  \multicolumn{1}{c|}{\textbf{73.97}} &
  63.08 \\ \hline
-0.75 &
  \multicolumn{1}{c|}{69.27} &
  \multicolumn{1}{c|}{55.59} &
  77.09 &
  \multicolumn{1}{c|}{80.20} &
  \multicolumn{1}{c|}{68.79} &
  66.63 &
  \multicolumn{1}{c|}{77.09} &
  \multicolumn{1}{c|}{88.64} &
  70.64 &
  \multicolumn{1}{c|}{75.14} &
  \multicolumn{1}{c|}{72.56} &
  70.71 \\ \hline
\end{tabular}
\end{subtable}
\end{table}

\begin{table}[h!]
\centering
\scriptsize
\captionsetup{font=footnotesize}
\caption{Detailed evaluation result on CIFAR100-LT depending on $\gamma$ : OOD detection performance (AUROC,AP, FPR) and classification accuracy (ACC) with model ResNet18; Std over six random runs are reported; (a): ACC and result on Texture, SVHN, and CIFAR10; (b): Total average result and result on Tiny Imagenet, LSUN, and Place365}
\vspace{-0.2cm}
\begin{subtable}{1\linewidth}
\centering
\scriptsize
\caption{}
\begin{tabular}{cc|l|lll|lll|lll}
\hline
\multicolumn{2}{c|}{Method} &
  \multicolumn{1}{c|}{\multirow{2}{*}{ACC$\uparrow$}} &
  \multicolumn{3}{c|}{Texture} &
  \multicolumn{3}{c|}{SVHN} &
  \multicolumn{3}{c}{CIFAR 10} \\ \cline{1-2} \cline{4-12} 
\multicolumn{1}{c|}{Name} &
  $\gamma$ &
  \multicolumn{1}{c|}{} &
  \multicolumn{1}{c|}{AUC$\uparrow$} &
  \multicolumn{1}{c|}{AP$\uparrow$} &
  \multicolumn{1}{c|}{FPR$\downarrow$} &
  \multicolumn{1}{c|}{AUC$\uparrow$} &
  \multicolumn{1}{c|}{AP$\uparrow$} &
  \multicolumn{1}{c|}{FPR$\downarrow$} &
  \multicolumn{1}{c|}{AUC$\uparrow$} &
  \multicolumn{1}{c|}{AP$\uparrow$} &
  \multicolumn{1}{c}{FPR$\downarrow$} \\ \hline
\multicolumn{1}{c|}{Energy OE} &
  0.00 &
  1.63E-02 &
  \multicolumn{1}{l|}{3.73E-03} &
  \multicolumn{1}{l|}{1.25E-02} &
  4.22E-02 &
  \multicolumn{1}{l|}{2.03E-02} &
  \multicolumn{1}{l|}{1.41E-02} &
  4.07E-02 &
  \multicolumn{1}{l|}{4.71E-03} &
  \multicolumn{1}{l|}{3.73E-03} &
  2.93E-02 \\ \hline
\multicolumn{1}{c|}{\multirow{6}{*}{\begin{tabular}[c]{@{}c@{}}Balanced \\ Energy OE\end{tabular}}} &
  0.10 &
  2.48E-02 &
  \multicolumn{1}{l|}{1.12E-02} &
  \multicolumn{1}{l|}{2.29E-02} &
  6.19E-02 &
  \multicolumn{1}{l|}{1.53E-02} &
  \multicolumn{1}{l|}{1.11E-02} &
  4.23E-02 &
  \multicolumn{1}{l|}{0.00E+00} &
  \multicolumn{1}{l|}{4.71E-03} &
  2.92E-02 \\
\multicolumn{1}{c|}{} &
  0.25 &
  2.91E-02 &
  \multicolumn{1}{l|}{3.73E-03} &
  \multicolumn{1}{l|}{6.87E-03} &
  6.47E-02 &
  \multicolumn{1}{l|}{1.86E-02} &
  \multicolumn{1}{l|}{1.77E-02} &
  8.08E-02 &
  \multicolumn{1}{l|}{4.71E-03} &
  \multicolumn{1}{l|}{5.00E-03} &
  2.99E-02 \\
\multicolumn{1}{c|}{} &
  0.50 &
  2.48E-02 &
  \multicolumn{1}{l|}{9.57E-03} &
  \multicolumn{1}{l|}{1.71E-02} &
  5.16E-02 &
  \multicolumn{1}{l|}{2.27E-02} &
  \multicolumn{1}{l|}{1.61E-02} &
  4.81E-02 &
  \multicolumn{1}{l|}{4.71E-03} &
  \multicolumn{1}{l|}{5.00E-03} &
  4.36E-02 \\
\multicolumn{1}{c|}{} &
  0.75 &
  3.64E-02 &
  \multicolumn{1}{l|}{7.45E-03} &
  \multicolumn{1}{l|}{1.25E-02} &
  1.09E-01 &
  \multicolumn{1}{l|}{1.95E-02} &
  \multicolumn{1}{l|}{1.41E-02} &
  2.11E-02 &
  \multicolumn{1}{l|}{3.73E-03} &
  \multicolumn{1}{l|}{3.73E-03} &
  3.09E-02 \\
\multicolumn{1}{c|}{} &
  1.00 &
  3.35E-02 &
  \multicolumn{1}{l|}{9.43E-03} &
  \multicolumn{1}{l|}{8.98E-03} &
  1.06E-01 &
  \multicolumn{1}{l|}{2.21E-02} &
  \multicolumn{1}{l|}{1.38E-02} &
  1.36E-01 &
  \multicolumn{1}{l|}{5.77E-03} &
  \multicolumn{1}{l|}{3.73E-03} &
  4.35E-02 \\
\multicolumn{1}{c|}{} &
  1.25 &
  3.83E-02 &
  \multicolumn{1}{l|}{9.43E-03} &
  \multicolumn{1}{l|}{1.15E-02} &
  4.99E-02 &
  \multicolumn{1}{l|}{2.13E-02} &
  \multicolumn{1}{l|}{1.89E-02} &
  8.65E-02 &
  \multicolumn{1}{l|}{7.64E-03} &
  \multicolumn{1}{l|}{9.57E-03} &
  4.35E-02 \\ \hline
\multicolumn{1}{c|}{\begin{tabular}[c]{@{}c@{}}Inv-Balanced\\  Energy OE\end{tabular}} &
  -0.75 &
  2.31E-02 &
  \multicolumn{1}{l|}{4.71E-03} &
  \multicolumn{1}{l|}{4.71E-03} &
  2.81E-02 &
  \multicolumn{1}{l|}{2.00E-02} &
  \multicolumn{1}{l|}{1.26E-02} &
  7.39E-02 &
  \multicolumn{1}{l|}{7.11E-15} &
  \multicolumn{1}{l|}{4.71E-03} &
  2.08E-02 \\ \hline
\end{tabular}

\caption{}
\begin{tabular}{c|lll|lll|lll|lll}
\hline
Method &
  \multicolumn{3}{c|}{Tiny ImageNet} &
  \multicolumn{3}{c|}{LSUN} &
  \multicolumn{3}{c|}{Place365} &
  \multicolumn{3}{c}{Average} \\ \hline
$\gamma$ &
  \multicolumn{1}{c|}{AUC$\uparrow$} &
  \multicolumn{1}{c|}{AP$\uparrow$} &
  \multicolumn{1}{c|}{FPR$\downarrow$} &
  \multicolumn{1}{c|}{AUC$\uparrow$} &
  \multicolumn{1}{c|}{AP$\uparrow$} &
  \multicolumn{1}{c|}{FPR$\downarrow$} &
  \multicolumn{1}{c|}{AUC$\uparrow$} &
  \multicolumn{1}{c|}{AP$\uparrow$} &
  \multicolumn{1}{c|}{FPR$\downarrow$} &
  \multicolumn{1}{c|}{AUC$\uparrow$} &
  \multicolumn{1}{c|}{AP$\uparrow$} &
  \multicolumn{1}{c}{FPR$\downarrow$} \\ \hline
0.00 &
  \multicolumn{1}{l|}{1.42E-14} &
  \multicolumn{1}{l|}{4.71E-03} &
  4.41E-02 &
  \multicolumn{1}{l|}{0.00E+00} &
  \multicolumn{1}{l|}{4.71E-03} &
  9.09E-02 &
  \multicolumn{1}{l|}{0.00E+00} &
  \multicolumn{1}{l|}{1.42E-14} &
  4.12E-02 &
  \multicolumn{1}{l|}{4.80E-03} &
  \multicolumn{1}{l|}{6.63E-03} &
  4.81E-02 \\ \hline
0.10 &
  \multicolumn{1}{l|}{0.00E+00} &
  \multicolumn{1}{l|}{3.73E-03} &
  3.59E-02 &
  \multicolumn{1}{l|}{5.00E-03} &
  \multicolumn{1}{l|}{4.71E-03} &
  8.35E-02 &
  \multicolumn{1}{l|}{0.00E+00} &
  \multicolumn{1}{l|}{5.00E-03} &
  3.53E-02 &
  \multicolumn{1}{l|}{5.24E-03} &
  \multicolumn{1}{l|}{8.68E-03} &
  4.80E-02 \\
0.25 &
  \multicolumn{1}{l|}{4.71E-03} &
  \multicolumn{1}{l|}{4.71E-03} &
  4.34E-02 &
  \multicolumn{1}{l|}{3.73E-03} &
  \multicolumn{1}{l|}{3.73E-03} &
  4.14E-02 &
  \multicolumn{1}{l|}{4.71E-03} &
  \multicolumn{1}{l|}{3.73E-03} &
  1.63E-02 &
  \multicolumn{1}{l|}{6.70E-03} &
  \multicolumn{1}{l|}{6.96E-03} &
  4.61E-02 \\
0.50 &
  \multicolumn{1}{l|}{0.00E+00} &
  \multicolumn{1}{l|}{4.71E-03} &
  5.58E-02 &
  \multicolumn{1}{l|}{4.71E-03} &
  \multicolumn{1}{l|}{3.73E-03} &
  6.15E-02 &
  \multicolumn{1}{l|}{5.00E-03} &
  \multicolumn{1}{l|}{3.73E-03} &
  6.74E-02 &
  \multicolumn{1}{l|}{7.78E-03} &
  \multicolumn{1}{l|}{8.39E-03} &
  5.47E-02 \\
0.75 &
  \multicolumn{1}{l|}{3.73E-03} &
  \multicolumn{1}{l|}{6.87E-03} &
  6.52E-02 &
  \multicolumn{1}{l|}{0.00E+00} &
  \multicolumn{1}{l|}{5.00E-03} &
  7.91E-02 &
  \multicolumn{1}{l|}{4.71E-03} &
  \multicolumn{1}{l|}{5.00E-03} &
  4.26E-02 &
  \multicolumn{1}{l|}{6.52E-03} &
  \multicolumn{1}{l|}{7.87E-03} &
  5.80E-02 \\
1.00 &
  \multicolumn{1}{l|}{6.87E-03} &
  \multicolumn{1}{l|}{1.07E-02} &
  3.50E-02 &
  \multicolumn{1}{l|}{5.77E-03} &
  \multicolumn{1}{l|}{6.87E-03} &
  4.78E-02 &
  \multicolumn{1}{l|}{3.73E-03} &
  \multicolumn{1}{l|}{4.71E-03} &
  5.89E-02 &
  \multicolumn{1}{l|}{8.95E-03} &
  \multicolumn{1}{l|}{8.13E-03} &
  7.11E-02 \\
1.25 &
  \multicolumn{1}{l|}{4.71E-03} &
  \multicolumn{1}{l|}{4.71E-03} &
  6.08E-02 &
  \multicolumn{1}{l|}{1.07E-02} &
  \multicolumn{1}{l|}{1.11E-02} &
  3.73E-02 &
  \multicolumn{1}{l|}{7.45E-03} &
  \multicolumn{1}{l|}{0.00E+00} &
  5.55E-02 &
  \multicolumn{1}{l|}{1.02E-02} &
  \multicolumn{1}{l|}{9.30E-03} &
  5.56E-02 \\ \hline
-0.75 &
  \multicolumn{1}{l|}{0.00E+00} &
  \multicolumn{1}{l|}{4.71E-03} &
  2.67E-02 &
  \multicolumn{1}{l|}{5.77E-03} &
  \multicolumn{1}{l|}{6.87E-03} &
  2.63E-02 &
  \multicolumn{1}{l|}{4.71E-03} &
  \multicolumn{1}{l|}{0.00E+00} &
  3.89E-02 &
  \multicolumn{1}{l|}{5.87E-03} &
  \multicolumn{1}{l|}{5.60E-03} &
  3.58E-02 \\ \hline
\end{tabular}
\end{subtable}
\end{table}

\newpage

{\bf\noindent WideResNet model on CIFAR10-LT:}

\begin{table}[h!]
\centering
\scriptsize
\captionsetup{font=footnotesize}
\caption{Detailed evaluation result on CIFAR10-LT depending on $\gamma$ : OOD detection performance (AUROC,AP, FPR) and classification accuracy (ACC) with model WideResNet(WRN-40-2); Mean over six random runs are reported; (a): ACC and result on Texture, SVHN, and CIFAR100; (b): Total average result and result on Tiny Imagenet, LSUN, and Place365}
\vspace{-0.2cm}
\begin{subtable}{1\linewidth}
\centering
\scriptsize
\caption{}
\begin{tabular}{cc|c|ccc|ccc|ccc}
\hline
\multicolumn{2}{c|}{Method} &
  \multirow{2}{*}{ACC$\uparrow$} &
  \multicolumn{3}{c|}{Texture} &
  \multicolumn{3}{c|}{SVHN} &
  \multicolumn{3}{c}{CIFAR 100} \\ \cline{1-2} \cline{4-12} 
\multicolumn{1}{c|}{Name} &
  $\gamma$ &
   &
  \multicolumn{1}{c|}{AUC$\uparrow$} &
  \multicolumn{1}{c|}{AP$\uparrow$} &
  FPR$\downarrow$ &
  \multicolumn{1}{c|}{AUC$\uparrow$} &
  \multicolumn{1}{c|}{AP$\uparrow$} &
  FPR$\downarrow$ &
  \multicolumn{1}{c|}{AUC$\uparrow$} &
  \multicolumn{1}{c|}{AP$\uparrow$} &
  FPR$\downarrow$ \\ \hline
\multicolumn{1}{c|}{Energy OE} &
  0.00 &
  75.03 &
  \multicolumn{1}{c|}{94.25} &
  \multicolumn{1}{c|}{89.94} &
  26.79 &
  \multicolumn{1}{c|}{95.07} &
  \multicolumn{1}{c|}{96.66} &
  16.08 &
  \multicolumn{1}{c|}{84.20} &
  \multicolumn{1}{c|}{84.00} &
  57.71 \\ \hline
\multicolumn{1}{c|}{\multirow{6}{*}{\begin{tabular}[c]{@{}c@{}}Balanced \\ Energy OE\end{tabular}}} &
  0.10 &
  75.28 &
  \multicolumn{1}{c|}{94.39} &
  \multicolumn{1}{c|}{89.27} &
  25.82 &
  \multicolumn{1}{c|}{95.22} &
  \multicolumn{1}{c|}{96.34} &
  14.76 &
  \multicolumn{1}{c|}{84.39} &
  \multicolumn{1}{c|}{83.81} &
  56.63 \\
\multicolumn{1}{c|}{} &
  0.25 &
  75.74 &
  \multicolumn{1}{c|}{94.32} &
  \multicolumn{1}{c|}{88.57} &
  25.55 &
  \multicolumn{1}{c|}{95.43} &
  \multicolumn{1}{c|}{96.33} &
  13.87 &
  \multicolumn{1}{c|}{84.60} &
  \multicolumn{1}{c|}{83.64} &
  55.22 \\
\multicolumn{1}{c|}{} &
  0.50 &
  \textbf{76.14} &
  \multicolumn{1}{c|}{94.31} &
  \multicolumn{1}{c|}{88.39} &
  26.08 &
  \multicolumn{1}{c|}{95.83} &
  \multicolumn{1}{c|}{96.73} &
  12.75 &
  \multicolumn{1}{c|}{84.80} &
  \multicolumn{1}{c|}{83.81} &
  54.68 \\
\multicolumn{1}{c|}{} &
  0.75 &
  74.76 &
  \multicolumn{1}{c|}{94.23} &
  \multicolumn{1}{c|}{88.75} &
  26.51 &
  \multicolumn{1}{c|}{96.23} &
  \multicolumn{1}{c|}{97.12} &
  12.24 &
  \multicolumn{1}{c|}{84.50} &
  \multicolumn{1}{c|}{83.66} &
  56.59 \\
\multicolumn{1}{c|}{} &
  1.00 &
  71.20 &
  \multicolumn{1}{c|}{94.14} &
  \multicolumn{1}{c|}{89.77} &
  29.40 &
  \multicolumn{1}{c|}{96.51} &
  \multicolumn{1}{c|}{97.74} &
  12.56 &
  \multicolumn{1}{c|}{84.00} &
  \multicolumn{1}{c|}{83.50} &
  58.40 \\
\multicolumn{1}{c|}{} &
  1.25 &
  68.70 &
  \multicolumn{1}{c|}{94.06} &
  \multicolumn{1}{c|}{90.32} &
  31.19 &
  \multicolumn{1}{c|}{96.63} &
  \multicolumn{1}{c|}{98.11} &
  13.97 &
  \multicolumn{1}{c|}{83.60} &
  \multicolumn{1}{c|}{83.33} &
  59.46 \\ \hline
\multicolumn{1}{c|}{\begin{tabular}[c]{@{}c@{}}Inv-Balanced\\  Energy OE\end{tabular}} &
  -0.75 &
  62.91 &
  \multicolumn{1}{c|}{93.15} &
  \multicolumn{1}{c|}{90.49} &
  39.41 &
  \multicolumn{1}{c|}{94.22} &
  \multicolumn{1}{c|}{97.22} &
  24.63 &
  \multicolumn{1}{c|}{80.54} &
  \multicolumn{1}{c|}{80.87} &
  65.46 \\ \hline
\end{tabular}

\caption{}
\begin{tabular}{c|c|ccc|ccc|ccc|ccc}
\hline
Method &
  \multirow{2}{*}{ACC$\uparrow$} &
  \multicolumn{3}{c|}{Tiny ImageNet} &
  \multicolumn{3}{c|}{LSUN} &
  \multicolumn{3}{c|}{Place365} &
  \multicolumn{3}{c}{Average} \\ \cline{1-1} \cline{3-14} 
$\gamma$ &
   &
  \multicolumn{1}{c|}{AUC$\uparrow$} &
  \multicolumn{1}{c|}{AP$\uparrow$} &
  FPR$\downarrow$ &
  \multicolumn{1}{c|}{AUC$\uparrow$} &
  \multicolumn{1}{c|}{AP$\uparrow$} &
  FPR$\downarrow$ &
  \multicolumn{1}{c|}{AUC$\uparrow$} &
  \multicolumn{1}{c|}{AP$\uparrow$} &
  FPR$\downarrow$ &
  \multicolumn{1}{c|}{AUC$\uparrow$} &
  \multicolumn{1}{c|}{AP$\uparrow$} &
  FPR$\downarrow$ \\ \hline
0.00 &
  75.03 &
  \multicolumn{1}{c|}{87.63} &
  \multicolumn{1}{c|}{83.86} &
  47.77 &
  \multicolumn{1}{c|}{94.80} &
  \multicolumn{1}{c|}{94.41} &
  24.17 &
  \multicolumn{1}{c|}{92.72} &
  \multicolumn{1}{c|}{97.17} &
  31.58 &
  \multicolumn{1}{c|}{91.44} &
  \multicolumn{1}{c|}{91.01} &
  34.02 \\ \hline
0.10 &
  75.28 &
  \multicolumn{1}{c|}{87.82} &
  \multicolumn{1}{c|}{83.75} &
  46.80 &
  \multicolumn{1}{c|}{94.78} &
  \multicolumn{1}{c|}{93.81} &
  22.68 &
  \multicolumn{1}{c|}{92.84} &
  \multicolumn{1}{c|}{97.06} &
  29.98 &
  \multicolumn{1}{c|}{91.57} &
  \multicolumn{1}{c|}{90.67} &
  32.78 \\
0.25 &
  75.74 &
  \multicolumn{1}{c|}{87.97} &
  \multicolumn{1}{c|}{83.61} &
  45.47 &
  \multicolumn{1}{c|}{94.72} &
  \multicolumn{1}{c|}{93.03} &
  21.51 &
  \multicolumn{1}{c|}{92.89} &
  \multicolumn{1}{c|}{96.89} &
  28.90 &
  \multicolumn{1}{c|}{91.65} &
  \multicolumn{1}{c|}{90.34} &
  31.75 \\
0.50 &
  \textbf{76.14} &
  \multicolumn{1}{c|}{88.12} &
  \multicolumn{1}{c|}{83.76} &
  44.61 &
  \multicolumn{1}{c|}{94.90} &
  \multicolumn{1}{c|}{93.23} &
  20.45 &
  \multicolumn{1}{c|}{93.16} &
  \multicolumn{1}{c|}{96.96} &
  27.61 &
  \multicolumn{1}{c|}{\textbf{91.85}} &
  \multicolumn{1}{c|}{90.48} &
  \textbf{31.03} \\
0.75 &
  74.76 &
  \multicolumn{1}{c|}{87.91} &
  \multicolumn{1}{c|}{83.71} &
  45.68 &
  \multicolumn{1}{c|}{94.94} &
  \multicolumn{1}{c|}{93.67} &
  20.91 &
  \multicolumn{1}{c|}{93.19} &
  \multicolumn{1}{c|}{97.05} &
  28.25 &
  \multicolumn{1}{c|}{91.83} &
  \multicolumn{1}{c|}{90.66} &
  31.69 \\
1.00 &
  71.20 &
  \multicolumn{1}{c|}{87.59} &
  \multicolumn{1}{c|}{83.56} &
  47.68 &
  \multicolumn{1}{c|}{94.91} &
  \multicolumn{1}{c|}{94.25} &
  22.63 &
  \multicolumn{1}{c|}{93.19} &
  \multicolumn{1}{c|}{97.21} &
  29.08 &
  \multicolumn{1}{c|}{91.72} &
  \multicolumn{1}{c|}{91.01} &
  33.29 \\
1.25 &
  68.70 &
  \multicolumn{1}{c|}{87.22} &
  \multicolumn{1}{c|}{83.42} &
  48.87 &
  \multicolumn{1}{c|}{94.81} &
  \multicolumn{1}{c|}{92.28} &
  23.16 &
  \multicolumn{1}{c|}{93.06} &
  \multicolumn{1}{c|}{97.22} &
  30.14 &
  \multicolumn{1}{c|}{91.56} &
  \multicolumn{1}{c|}{\textbf{91.11}} &
  34.46 \\ \hline
-0.75 &
  62.91 &
  \multicolumn{1}{c|}{84.81} &
  \multicolumn{1}{c|}{81.26} &
  56.31 &
  \multicolumn{1}{c|}{92.93} &
  \multicolumn{1}{c|}{92.66} &
  36.62 &
  \multicolumn{1}{c|}{90.38} &
  \multicolumn{1}{c|}{96.35} &
  44.05 &
  \multicolumn{1}{c|}{89.34} &
  \multicolumn{1}{c|}{89.81} &
  44.41 \\ \hline
\end{tabular}
\end{subtable}
\end{table}

\begin{table}[h!]
\centering
\tiny
\captionsetup{font=footnotesize}
\caption{Detailed evaluation result on CIFAR10-LT depending on $\gamma$ : OOD detection performance (AUROC,AP, FPR) and classification accuracy (ACC) with model WideResNet(WRN-40-2); Std over six random runs are reported; (a): ACC and result on Texture, SVHN, and CIFAR100; (b): Total average result and result on Tiny Imagenet, LSUN, and Place365}
\vspace{-0.2cm}
\begin{subtable}{1\linewidth}
\centering
\scriptsize
\caption{}
\begin{tabular}{cc|l|lll|lll|lll}
\hline
\multicolumn{2}{c|}{Method} &
  \multicolumn{1}{c|}{\multirow{2}{*}{ACC$\uparrow$}} &
  \multicolumn{3}{c|}{Texture} &
  \multicolumn{3}{c|}{SVHN} &
  \multicolumn{3}{c}{CIFAR 100} \\ \cline{1-2} \cline{4-12} 
\multicolumn{1}{c|}{Name} &
  $\gamma$ &
  \multicolumn{1}{c|}{} &
  \multicolumn{1}{c|}{AUC$\uparrow$} &
  \multicolumn{1}{c|}{AP$\uparrow$} &
  \multicolumn{1}{c|}{FPR$\downarrow$} &
  \multicolumn{1}{c|}{AUC$\uparrow$} &
  \multicolumn{1}{c|}{AP$\uparrow$} &
  \multicolumn{1}{c|}{FPR$\downarrow$} &
  \multicolumn{1}{c|}{AUC$\uparrow$} &
  \multicolumn{1}{c|}{AP$\uparrow$} &
  \multicolumn{1}{c}{FPR$\downarrow$} \\ \hline
\multicolumn{1}{c|}{Energy OE} &
  0.00 &
  4.92E-02 &
  \multicolumn{1}{l|}{6.87E-03} &
  \multicolumn{1}{l|}{1.26E-02} &
  7.27E-02 &
  \multicolumn{1}{l|}{1.49E-02} &
  \multicolumn{1}{l|}{1.25E-02} &
  5.34E-02 &
  \multicolumn{1}{l|}{6.87E-03} &
  \multicolumn{1}{l|}{6.87E-03} &
  1.36E-01 \\ \hline
\multicolumn{1}{c|}{\multirow{6}{*}{\begin{tabular}[c]{@{}c@{}}Balanced \\ Energy OE\end{tabular}}} &
  0.10 &
  3.80E-02 &
  \multicolumn{1}{l|}{5.77E-03} &
  \multicolumn{1}{l|}{1.63E-02} &
  1.27E-01 &
  \multicolumn{1}{l|}{1.34E-02} &
  \multicolumn{1}{l|}{9.57E-03} &
  4.35E-02 &
  \multicolumn{1}{l|}{3.73E-03} &
  \multicolumn{1}{l|}{3.73E-03} &
  6.90E-02 \\
\multicolumn{1}{c|}{} &
  0.25 &
  3.72E-02 &
  \multicolumn{1}{l|}{9.43E-03} &
  \multicolumn{1}{l|}{2.67E-02} &
  1.34E-01 &
  \multicolumn{1}{l|}{1.71E-02} &
  \multicolumn{1}{l|}{1.73E-02} &
  4.76E-02 &
  \multicolumn{1}{l|}{5.77E-03} &
  \multicolumn{1}{l|}{1.11E-02} &
  7.86E-02 \\
\multicolumn{1}{c|}{} &
  0.50 &
  4.74E-02 &
  \multicolumn{1}{l|}{6.87E-03} &
  \multicolumn{1}{l|}{4.20E-02} &
  9.52E-02 &
  \multicolumn{1}{l|}{1.80E-02} &
  \multicolumn{1}{l|}{1.67E-02} &
  7.24E-02 &
  \multicolumn{1}{l|}{4.71E-03} &
  \multicolumn{1}{l|}{1.61E-02} &
  9.99E-02 \\
\multicolumn{1}{c|}{} &
  0.75 &
  5.70E-02 &
  \multicolumn{1}{l|}{9.43E-03} &
  \multicolumn{1}{l|}{5.27E-02} &
  1.56E-01 &
  \multicolumn{1}{l|}{1.57E-02} &
  \multicolumn{1}{l|}{1.57E-02} &
  1.25E-01 &
  \multicolumn{1}{l|}{1.11E-02} &
  \multicolumn{1}{l|}{2.06E-02} &
  2.15E-01 \\
\multicolumn{1}{c|}{} &
  1.00 &
  6.52E-02 &
  \multicolumn{1}{l|}{1.12E-02} &
  \multicolumn{1}{l|}{4.56E-02} &
  1.39E-01 &
  \multicolumn{1}{l|}{1.71E-02} &
  \multicolumn{1}{l|}{2.81E-02} &
  1.66E-01 &
  \multicolumn{1}{l|}{6.87E-03} &
  \multicolumn{1}{l|}{1.95E-02} &
  2.36E-01 \\
\multicolumn{1}{c|}{} &
  1.25 &
  8.08E-02 &
  \multicolumn{1}{l|}{2.48E-02} &
  \multicolumn{1}{l|}{4.15E-02} &
  1.65E-01 &
  \multicolumn{1}{l|}{5.19E-02} &
  \multicolumn{1}{l|}{4.11E-02} &
  2.43E-01 &
  \multicolumn{1}{l|}{1.70E-02} &
  \multicolumn{1}{l|}{1.80E-02} &
  1.47E-01 \\ \hline
\multicolumn{1}{c|}{\begin{tabular}[c]{@{}c@{}}Inv-Balanced\\  Energy OE\end{tabular}} &
  -0.75 &
  2.49E-02 &
  \multicolumn{1}{l|}{1.34E-02} &
  \multicolumn{1}{l|}{2.29E-02} &
  1.34E-01 &
  \multicolumn{1}{l|}{3.02E-02} &
  \multicolumn{1}{l|}{1.50E-02} &
  1.26E-01 &
  \multicolumn{1}{l|}{0.00E+00} &
  \multicolumn{1}{l|}{6.87E-03} &
  1.05E-01 \\ \hline
\end{tabular}

\caption{}
\begin{tabular}{c|lll|lll|lll|lll}
\hline
Method &
  \multicolumn{3}{c|}{Tiny ImageNet} &
  \multicolumn{3}{c|}{LSUN} &
  \multicolumn{3}{c|}{Place365} &
  \multicolumn{3}{c}{Average} \\ \hline
$\gamma$ &
  \multicolumn{1}{c|}{AUC$\uparrow$} &
  \multicolumn{1}{c|}{AP$\uparrow$} &
  \multicolumn{1}{c|}{FPR$\downarrow$} &
  \multicolumn{1}{c|}{AUC$\uparrow$} &
  \multicolumn{1}{c|}{AP$\uparrow$} &
  \multicolumn{1}{c|}{FPR$\downarrow$} &
  \multicolumn{1}{c|}{AUC$\uparrow$} &
  \multicolumn{1}{c|}{AP$\uparrow$} &
  \multicolumn{1}{c|}{FPR$\downarrow$} &
  \multicolumn{1}{c|}{AUC$\uparrow$} &
  \multicolumn{1}{c|}{AP$\uparrow$} &
  \multicolumn{1}{c}{FPR$\downarrow$} \\ \hline
0.00 &
  \multicolumn{1}{l|}{5.00E-03} &
  \multicolumn{1}{l|}{7.45E-03} &
  5.89E-02 &
  \multicolumn{1}{l|}{6.87E-03} &
  \multicolumn{1}{l|}{6.87E-03} &
  6.09E-02 &
  \multicolumn{1}{l|}{5.77E-03} &
  \multicolumn{1}{l|}{3.73E-03} &
  6.59E-02 &
  \multicolumn{1}{l|}{7.72E-03} &
  \multicolumn{1}{l|}{8.33E-03} &
  7.46E-02 \\ \hline
0.10 &
  \multicolumn{1}{l|}{7.45E-03} &
  \multicolumn{1}{l|}{6.87E-03} &
  1.10E-01 &
  \multicolumn{1}{l|}{3.73E-03} &
  \multicolumn{1}{l|}{7.45E-03} &
  8.56E-02 &
  \multicolumn{1}{l|}{4.71E-03} &
  \multicolumn{1}{l|}{3.73E-03} &
  8.64E-02 &
  \multicolumn{1}{l|}{6.47E-03} &
  \multicolumn{1}{l|}{7.95E-03} &
  8.69E-02 \\
0.25 &
  \multicolumn{1}{l|}{3.73E-03} &
  \multicolumn{1}{l|}{8.98E-03} &
  7.39E-02 &
  \multicolumn{1}{l|}{6.87E-03} &
  \multicolumn{1}{l|}{1.60E-02} &
  5.44E-02 &
  \multicolumn{1}{l|}{0.00E+00} &
  \multicolumn{1}{l|}{5.00E-03} &
  4.53E-02 &
  \multicolumn{1}{l|}{7.15E-03} &
  \multicolumn{1}{l|}{1.42E-02} &
  7.22E-02 \\
0.50 &
  \multicolumn{1}{l|}{7.64E-03} &
  \multicolumn{1}{l|}{2.13E-02} &
  5.25E-02 &
  \multicolumn{1}{l|}{3.73E-03} &
  \multicolumn{1}{l|}{1.71E-02} &
  5.87E-02 &
  \multicolumn{1}{l|}{6.87E-03} &
  \multicolumn{1}{l|}{7.45E-03} &
  6.38E-02 &
  \multicolumn{1}{l|}{7.97E-03} &
  \multicolumn{1}{l|}{2.01E-02} &
  7.38E-02 \\
0.75 &
  \multicolumn{1}{l|}{4.71E-03} &
  \multicolumn{1}{l|}{6.87E-03} &
  1.30E-01 &
  \multicolumn{1}{l|}{6.87E-03} &
  \multicolumn{1}{l|}{1.53E-02} &
  7.74E-02 &
  \multicolumn{1}{l|}{6.87E-03} &
  \multicolumn{1}{l|}{4.71E-03} &
  1.13E-01 &
  \multicolumn{1}{l|}{9.11E-03} &
  \multicolumn{1}{l|}{1.93E-02} &
  1.36E-01 \\
1.00 &
  \multicolumn{1}{l|}{1.26E-02} &
  \multicolumn{1}{l|}{2.49E-02} &
  1.65E-01 &
  \multicolumn{1}{l|}{1.07E-02} &
  \multicolumn{1}{l|}{2.11E-02} &
  1.08E-01 &
  \multicolumn{1}{l|}{1.11E-02} &
  \multicolumn{1}{l|}{9.43E-03} &
  7.04E-02 &
  \multicolumn{1}{l|}{1.16E-02} &
  \multicolumn{1}{l|}{2.48E-02} &
  1.48E-01 \\
1.25 &
  \multicolumn{1}{l|}{1.25E-02} &
  \multicolumn{1}{l|}{1.29E-02} &
  1.93E-01 &
  \multicolumn{1}{l|}{1.37E-02} &
  \multicolumn{1}{l|}{2.54E-02} &
  9.66E-02 &
  \multicolumn{1}{l|}{3.73E-03} &
  \multicolumn{1}{l|}{5.00E-03} &
  9.37E-02 &
  \multicolumn{1}{l|}{2.06E-02} &
  \multicolumn{1}{l|}{2.40E-02} &
  1.56E-01 \\ \hline
-0.75 &
  \multicolumn{1}{l|}{5.77E-03} &
  \multicolumn{1}{l|}{6.87E-03} &
  7.06E-02 &
  \multicolumn{1}{l|}{7.45E-03} &
  \multicolumn{1}{l|}{1.86E-02} &
  1.05E-01 &
  \multicolumn{1}{l|}{5.77E-03} &
  \multicolumn{1}{l|}{5.00E-03} &
  5.93E-02 &
  \multicolumn{1}{l|}{1.04E-02} &
  \multicolumn{1}{l|}{1.25E-02} &
  1.00E-01 \\ \hline
\end{tabular}
\end{subtable}
\end{table}

\newpage

{\bf\noindent WideResNet model on CIFAR100-LT:}

\begin{table}[h!]
\centering
\scriptsize
\captionsetup{font=footnotesize}
\caption{Detailed evaluation result on CIFAR100-LT depending on $\gamma$ : OOD detection performance (AUROC,AP, FPR) and classification accuracy (ACC) with model WideResNet(WRN-40-2); Mean over six random runs are reported; (a): ACC and result on Texture, SVHN, and CIFAR10; (b): Total average result and result on Tiny Imagenet, LSUN, and Place365; N/A: not available because of  unstable neural networks training}
\vspace{-0.2cm}
\begin{subtable}{1\linewidth}
\centering
\scriptsize
\caption{}
\begin{tabular}{cc|c|ccc|ccc|ccc}
\hline
\multicolumn{2}{c|}{Method} &
  \multirow{2}{*}{ACC$\uparrow$} &
  \multicolumn{3}{c|}{Texture} &
  \multicolumn{3}{c|}{SVHN} &
  \multicolumn{3}{c}{CIFAR 10} \\ \cline{1-2} \cline{4-12} 
\multicolumn{1}{c|}{Name} &
  $\gamma$ &
   &
  \multicolumn{1}{c|}{AUC$\uparrow$} &
  \multicolumn{1}{c|}{AP$\uparrow$} &
  FPR$\downarrow$ &
  \multicolumn{1}{c|}{AUC$\uparrow$} &
  \multicolumn{1}{c|}{AP$\uparrow$} &
  FPR$\downarrow$ &
  \multicolumn{1}{c|}{AUC$\uparrow$} &
  \multicolumn{1}{c|}{AP$\uparrow$} &
  FPR$\downarrow$ \\ \hline
\multicolumn{1}{c|}{Energy OE} &
  0.00 &
  39.95 &
  \multicolumn{1}{c|}{80.50} &
  \multicolumn{1}{c|}{70.51} &
  65.56 &
  \multicolumn{1}{c|}{86.48} &
  \multicolumn{1}{c|}{92.68} &
  45.49 &
  \multicolumn{1}{c|}{60.52} &
  \multicolumn{1}{c|}{56.89} &
  82.90 \\ \hline
\multicolumn{1}{c|}{\multirow{6}{*}{\begin{tabular}[c]{@{}c@{}}Balanced \\ Energy OE\end{tabular}}} &
  0.10 &
  40.30 &
  \multicolumn{1}{c|}{80.57} &
  \multicolumn{1}{c|}{69.89} &
  64.59 &
  \multicolumn{1}{c|}{86.99} &
  \multicolumn{1}{c|}{92.93} &
  43.79 &
  \multicolumn{1}{c|}{60.46} &
  \multicolumn{1}{c|}{56.72} &
  82.97 \\
\multicolumn{1}{c|}{} &
  0.25 &
  \textbf{40.43} &
  \multicolumn{1}{c|}{80.92} &
  \multicolumn{1}{c|}{70.02} &
  63.84 &
  \multicolumn{1}{c|}{87.58} &
  \multicolumn{1}{c|}{93.32} &
  41.94 &
  \multicolumn{1}{c|}{60.12} &
  \multicolumn{1}{c|}{56.36} &
  83.46 \\
\multicolumn{1}{c|}{} &
  0.50 &
  39.44 &
  \multicolumn{1}{c|}{81.72} &
  \multicolumn{1}{c|}{71.40} &
  63.81 &
  \multicolumn{1}{c|}{88.23} &
  \multicolumn{1}{c|}{93.91} &
  41.00 &
  \multicolumn{1}{c|}{59.17} &
  \multicolumn{1}{c|}{55.44} &
  85.03 \\
\multicolumn{1}{c|}{} &
  0.75 &
  35.60 &
  \multicolumn{1}{c|}{82.86} &
  \multicolumn{1}{c|}{73.64} &
  64.46 &
  \multicolumn{1}{c|}{89.27} &
  \multicolumn{1}{c|}{94.50} &
  38.52 &
  \multicolumn{1}{c|}{58.45} &
  \multicolumn{1}{c|}{54.74} &
  85.60 \\
\multicolumn{1}{c|}{} &
  1.00 &
  32.45 &
  \multicolumn{1}{c|}{84.06} &
  \multicolumn{1}{c|}{76.45} &
  64.64 &
  \multicolumn{1}{c|}{90.26} &
  \multicolumn{1}{c|}{95.04} &
  37.27 &
  \multicolumn{1}{c|}{57.50} &
  \multicolumn{1}{c|}{53.65} &
  85.37 \\
\multicolumn{1}{c|}{} &
  1.25 &
  30.38 &
  \multicolumn{1}{c|}{84.70} &
  \multicolumn{1}{c|}{78.12} &
  65.06 &
  \multicolumn{1}{c|}{90.70} &
  \multicolumn{1}{c|}{95.40} &
  36.96 &
  \multicolumn{1}{c|}{56.71} &
  \multicolumn{1}{c|}{52.98} &
  86.00 \\ \hline
\multicolumn{1}{c|}{\begin{tabular}[c]{@{}c@{}}Inv-Balanced\\  Energy OE\end{tabular}} &
  -0.75 &
  N/A &
  \multicolumn{1}{c|}{N/A} &
  \multicolumn{1}{c|}{N/A} &
  N/A &
  \multicolumn{1}{c|}{N/A} &
  \multicolumn{1}{c|}{N/A} &
  N/A &
  \multicolumn{1}{c|}{N/A} &
  \multicolumn{1}{c|}{N/A} &
  N/A \\ \hline
\end{tabular}

\caption{}
\begin{tabular}{c|ccc|ccc|ccc|ccc}
\hline
Method &
  \multicolumn{3}{c|}{Tiny ImageNet} &
  \multicolumn{3}{c|}{LSUN} &
  \multicolumn{3}{c|}{Place365} &
  \multicolumn{3}{c}{Average} \\ \hline
$\gamma$ &
  \multicolumn{1}{c|}{AUC$\uparrow$} &
  \multicolumn{1}{c|}{AP$\uparrow$} &
  FPR$\downarrow$ &
  \multicolumn{1}{c|}{AUC$\uparrow$} &
  \multicolumn{1}{c|}{AP$\uparrow$} &
  FPR$\downarrow$ &
  \multicolumn{1}{c|}{AUC$\uparrow$} &
  \multicolumn{1}{c|}{AP$\uparrow$} &
  FPR$\downarrow$ &
  \multicolumn{1}{c|}{AUC$\uparrow$} &
  \multicolumn{1}{c|}{AP$\uparrow$} &
  FPR$\downarrow$ \\ \hline
0.00 &
  \multicolumn{1}{c|}{70.32} &
  \multicolumn{1}{c|}{56.65} &
  77.08 &
  \multicolumn{1}{c|}{80.95} &
  \multicolumn{1}{c|}{6.93} &
  61.56 &
  \multicolumn{1}{c|}{79.95} &
  \multicolumn{1}{c|}{89.83} &
  61.62 &
  \multicolumn{1}{c|}{76.45} &
  \multicolumn{1}{c|}{72.75} &
  65.70 \\ \hline
0.10 &
  \multicolumn{1}{c|}{70.40} &
  \multicolumn{1}{c|}{56.49} &
  76.79 &
  \multicolumn{1}{c|}{81.53} &
  \multicolumn{1}{c|}{70.21} &
  59.55 &
  \multicolumn{1}{c|}{80.36} &
  \multicolumn{1}{c|}{89.99} &
  60.28 &
  \multicolumn{1}{c|}{76.72} &
  \multicolumn{1}{c|}{72.70} &
  64.66 \\
0.25 &
  \multicolumn{1}{c|}{70.44} &
  \multicolumn{1}{c|}{56.15} &
  76.60 &
  \multicolumn{1}{c|}{82.29} &
  \multicolumn{1}{c|}{70.54} &
  56.84 &
  \multicolumn{1}{c|}{80.84} &
  \multicolumn{1}{c|}{90.17} &
  59.00 &
  \multicolumn{1}{c|}{77.03} &
  \multicolumn{1}{c|}{72.76} &
  63.61 \\
0.50 &
  \multicolumn{1}{c|}{70.46} &
  \multicolumn{1}{c|}{56.07} &
  76.66 &
  \multicolumn{1}{c|}{83.44} &
  \multicolumn{1}{c|}{71.42} &
  53.31 &
  \multicolumn{1}{c|}{81.45} &
  \multicolumn{1}{c|}{90.36} &
  57.25 &
  \multicolumn{1}{c|}{77.41} &
  \multicolumn{1}{c|}{73.10} &
  \textbf{62.84} \\
0.75 &
  \multicolumn{1}{c|}{70.59} &
  \multicolumn{1}{c|}{56.41} &
  76.97 &
  \multicolumn{1}{c|}{83.42} &
  \multicolumn{1}{c|}{71.63} &
  54.27 &
  \multicolumn{1}{c|}{80.90} &
  \multicolumn{1}{c|}{90.20} &
  60.85 &
  \multicolumn{1}{c|}{\textbf{77.58}} &
  \multicolumn{1}{c|}{\textbf{73.52}} &
  63.44 \\
1.00 &
  \multicolumn{1}{c|}{70.47} &
  \multicolumn{1}{c|}{56.46} &
  77.02 &
  \multicolumn{1}{c|}{80.78} &
  \multicolumn{1}{c|}{69.44} &
  65.38 &
  \multicolumn{1}{c|}{78.77} &
  \multicolumn{1}{c|}{89.36} &
  67.34 &
  \multicolumn{1}{c|}{76.97} &
  \multicolumn{1}{c|}{73.40} &
  66.17 \\
1.25 &
  \multicolumn{1}{c|}{70.24} &
  \multicolumn{1}{c|}{56.41} &
  76.32 &
  \multicolumn{1}{c|}{79.00} &
  \multicolumn{1}{c|}{67.9} &
  69.88 &
  \multicolumn{1}{c|}{77.10} &
  \multicolumn{1}{c|}{88.72} &
  70.86 &
  \multicolumn{1}{c|}{76.41} &
  \multicolumn{1}{c|}{73.25} &
  67.51 \\ \hline
-0.75 &
  \multicolumn{1}{c|}{N/A} &
  \multicolumn{1}{c|}{N/A} &
  N/A &
  \multicolumn{1}{c|}{N/A} &
  \multicolumn{1}{c|}{N/A} &
  N/A &
  \multicolumn{1}{c|}{N/A} &
  \multicolumn{1}{c|}{N/A} &
  N/A &
  \multicolumn{1}{c|}{N/A} &
  \multicolumn{1}{c|}{N/A} &
  N/A \\ \hline
\end{tabular}
\end{subtable}
\end{table}

\begin{table}[h!]
\centering
\scriptsize
\captionsetup{font=footnotesize}
\caption{Detailed evaluation result on CIFAR100-LT depending on $\gamma$ : OOD detection performance (AUROC,AP, FPR) and classification accuracy (ACC) with model WideResNet(WRN-40-2); Std over six random runs are reported; (a): ACC and result on Texture, SVHN, and CIFAR10; (b): Total average result and result on Tiny Imagenet, LSUN, and Place365; N/A: not available because of  unstable neural networks training}
\vspace{-0.2cm}
\begin{subtable}{1\linewidth}
\centering
\scriptsize
\caption{}
\begin{tabular}{cc|l|lll|lll|lll}
\hline
\multicolumn{2}{c|}{Method} &
  \multicolumn{1}{c|}{\multirow{2}{*}{ACC$\uparrow$}} &
  \multicolumn{3}{c|}{Texture} &
  \multicolumn{3}{c|}{SVHN} &
  \multicolumn{3}{c}{CIFAR 10} \\ \cline{1-2} \cline{4-12} 
\multicolumn{1}{c|}{Name} &
 $\gamma$ &
  \multicolumn{1}{c|}{} &
  \multicolumn{1}{c|}{AUC$\uparrow$} &
  \multicolumn{1}{c|}{AP$\uparrow$} &
  \multicolumn{1}{c|}{FPR$\downarrow$} &
  \multicolumn{1}{c|}{AUC$\uparrow$} &
  \multicolumn{1}{c|}{AP$\uparrow$} &
  \multicolumn{1}{c|}{FPR$\downarrow$} &
  \multicolumn{1}{c|}{AUC$\uparrow$} &
  \multicolumn{1}{c|}{AP$\uparrow$} &
  \multicolumn{1}{c}{FPR$\downarrow$} \\ \hline
\multicolumn{1}{c|}{Energy OE} &
  0.00 &
  2.81E-02 &
  \multicolumn{1}{l|}{1.77E-02} &
  \multicolumn{1}{l|}{3.13E-02} &
  8.84E-02 &
  \multicolumn{1}{l|}{1.80E-02} &
  \multicolumn{1}{l|}{1.50E-02} &
  8.82E-02 &
  \multicolumn{1}{l|}{1.07E-02} &
  \multicolumn{1}{l|}{6.87E-03} &
  2.92E-02 \\ \hline
\multicolumn{1}{c|}{\multirow{6}{*}{\begin{tabular}[c]{@{}c@{}}Balanced \\ Energy OE\end{tabular}}} &
  0.10 &
  3.18E-02 &
  \multicolumn{1}{l|}{1.07E-02} &
  \multicolumn{1}{l|}{2.41E-02} &
  1.38E-01 &
  \multicolumn{1}{l|}{4.57E-02} &
  \multicolumn{1}{l|}{3.27E-02} &
  1.55E-01 &
  \multicolumn{1}{l|}{6.87E-03} &
  \multicolumn{1}{l|}{5.00E-03} &
  1.89E-02 \\
\multicolumn{1}{c|}{} &
  0.25 &
  3.70E-02 &
  \multicolumn{1}{l|}{1.80E-02} &
  \multicolumn{1}{l|}{3.64E-02} &
  1.15E-01 &
  \multicolumn{1}{l|}{2.98E-02} &
  \multicolumn{1}{l|}{2.13E-02} &
  9.89E-02 &
  \multicolumn{1}{l|}{8.98E-03} &
  \multicolumn{1}{l|}{1.07E-02} &
  4.46E-02 \\
\multicolumn{1}{c|}{} &
  0.50 &
  2.21E-02 &
  \multicolumn{1}{l|}{6.87E-03} &
  \multicolumn{1}{l|}{1.57E-02} &
  8.86E-02 &
  \multicolumn{1}{l|}{5.50E-02} &
  \multicolumn{1}{l|}{3.20E-02} &
  9.32E-02 &
  \multicolumn{1}{l|}{9.57E-03} &
  \multicolumn{1}{l|}{8.98E-03} &
  1.00E-01 \\
\multicolumn{1}{c|}{} &
  0.75 &
  3.59E-02 &
  \multicolumn{1}{l|}{9.43E-03} &
  \multicolumn{1}{l|}{1.77E-02} &
  1.64E-01 &
  \multicolumn{1}{l|}{5.15E-02} &
  \multicolumn{1}{l|}{2.63E-02} &
  1.48E-01 &
  \multicolumn{1}{l|}{8.98E-03} &
  \multicolumn{1}{l|}{6.87E-03} &
  3.40E-02 \\
\multicolumn{1}{c|}{} &
  1.00 &
  2.71E-02 &
  \multicolumn{1}{l|}{1.97E-02} &
  \multicolumn{1}{l|}{2.65E-02} &
  1.99E-01 &
  \multicolumn{1}{l|}{4.61E-02} &
  \multicolumn{1}{l|}{3.48E-02} &
  1.60E-01 &
  \multicolumn{1}{l|}{1.34E-02} &
  \multicolumn{1}{l|}{1.07E-02} &
  7.93E-02 \\
\multicolumn{1}{c|}{} &
  1.25 &
  4.03E-02 &
  \multicolumn{1}{l|}{1.37E-02} &
  \multicolumn{1}{l|}{2.75E-02} &
  1.36E-01 &
  \multicolumn{1}{l|}{6.01E-02} &
  \multicolumn{1}{l|}{2.79E-02} &
  3.19E-01 &
  \multicolumn{1}{l|}{1.25E-02} &
  \multicolumn{1}{l|}{1.12E-02} &
  7.02E-02 \\ \hline
\multicolumn{1}{c|}{\begin{tabular}[c]{@{}c@{}}Inv-Balanced\\  Energy OE\end{tabular}} &
  -0.75 &
  \multicolumn{1}{c|}{N/A} &
  \multicolumn{1}{c|}{N/A} &
  \multicolumn{1}{c|}{N/A} &
  \multicolumn{1}{c|}{N/A} &
  \multicolumn{1}{c|}{N/A} &
  \multicolumn{1}{c|}{N/A} &
  \multicolumn{1}{c|}{N/A} &
  \multicolumn{1}{c|}{N/A} &
  \multicolumn{1}{c|}{N/A} &
  \multicolumn{1}{c}{N/A} \\ \hline
\end{tabular}

\caption{}
\begin{tabular}{c|lll|lll|lll|lll}
\hline
Method &
  \multicolumn{3}{c|}{Tiny ImageNet} &
  \multicolumn{3}{c|}{LSUN} &
  \multicolumn{3}{c|}{Place365} &
  \multicolumn{3}{c}{Average} \\ \hline
$\gamma$ &
  \multicolumn{1}{c|}{AUC$\uparrow$} &
  \multicolumn{1}{c|}{AP$\uparrow$} &
  \multicolumn{1}{c|}{FPR$\downarrow$} &
  \multicolumn{1}{c|}{AUC$\uparrow$} &
  \multicolumn{1}{c|}{AP$\uparrow$} &
  \multicolumn{1}{c|}{FPR$\downarrow$} &
  \multicolumn{1}{c|}{AUC$\uparrow$} &
  \multicolumn{1}{c|}{AP$\uparrow$} &
  \multicolumn{1}{c|}{FPR$\downarrow$} &
  \multicolumn{1}{c|}{AUC$\uparrow$} &
  \multicolumn{1}{c|}{AP$\uparrow$} &
  \multicolumn{1}{c}{FPR$\downarrow$} \\ \hline
0.00 &
  \multicolumn{1}{l|}{3.73E-03} &
  \multicolumn{1}{l|}{3.73E-03} &
  6.58E-02 &
  \multicolumn{1}{l|}{6.87E-03} &
  \multicolumn{1}{l|}{9.57E-03} &
  5.66E-02 &
  \multicolumn{1}{l|}{3.73E-03} &
  \multicolumn{1}{l|}{3.73E-03} &
  3.13E-02 &
  \multicolumn{1}{l|}{1.01E-02} &
  \multicolumn{1}{l|}{1.17E-02} &
  5.99E-02 \\ \hline
0.10 &
  \multicolumn{1}{l|}{3.73E-03} &
  \multicolumn{1}{l|}{7.45E-03} &
  8.28E-02 &
  \multicolumn{1}{l|}{9.43E-03} &
  \multicolumn{1}{l|}{1.46E-02} &
  9.81E-02 &
  \multicolumn{1}{l|}{3.73E-03} &
  \multicolumn{1}{l|}{4.71E-03} &
  5.80E-02 &
  \multicolumn{1}{l|}{1.34E-02} &
  \multicolumn{1}{l|}{1.48E-02} &
  9.18E-02 \\
0.25 &
  \multicolumn{1}{l|}{6.87E-03} &
  \multicolumn{1}{l|}{1.07E-02} &
  4.92E-02 &
  \multicolumn{1}{l|}{7.64E-03} &
  \multicolumn{1}{l|}{1.38E-02} &
  1.57E-01 &
  \multicolumn{1}{l|}{5.77E-03} &
  \multicolumn{1}{l|}{3.73E-03} &
  6.34E-02 &
  \multicolumn{1}{l|}{1.28E-02} &
  \multicolumn{1}{l|}{1.61E-02} &
  8.79E-02 \\
0.50 &
  \multicolumn{1}{l|}{7.64E-03} &
  \multicolumn{1}{l|}{1.07E-02} &
  2.87E-02 &
  \multicolumn{1}{l|}{6.87E-03} &
  \multicolumn{1}{l|}{1.80E-02} &
  1.03E-01 &
  \multicolumn{1}{l|}{5.00E-03} &
  \multicolumn{1}{l|}{4.71E-03} &
  8.75E-02 &
  \multicolumn{1}{l|}{1.52E-02} &
  \multicolumn{1}{l|}{1.50E-02} &
  8.36E-02 \\
0.75 &
  \multicolumn{1}{l|}{7.64E-03} &
  \multicolumn{1}{l|}{1.25E-02} &
  1.81E-01 &
  \multicolumn{1}{l|}{6.87E-03} &
  \multicolumn{1}{l|}{1.07E-02} &
  1.33E-01 &
  \multicolumn{1}{l|}{9.43E-03} &
  \multicolumn{1}{l|}{4.71E-03} &
  4.71E-02 &
  \multicolumn{1}{l|}{1.56E-02} &
  \multicolumn{1}{l|}{1.31E-02} &
  1.18E-01 \\
1.00 &
  \multicolumn{1}{l|}{8.98E-03} &
  \multicolumn{1}{l|}{9.43E-03} &
  9.67E-02 &
  \multicolumn{1}{l|}{8.98E-03} &
  \multicolumn{1}{l|}{1.15E-02} &
  8.15E-02 &
  \multicolumn{1}{l|}{1.00E-02} &
  \multicolumn{1}{l|}{6.87E-03} &
  7.19E-02 &
  \multicolumn{1}{l|}{1.79E-02} &
  \multicolumn{1}{l|}{1.66E-02} &
  1.15E-01 \\
1.25 &
  \multicolumn{1}{l|}{7.45E-03} &
  \multicolumn{1}{l|}{8.16E-03} &
  6.64E-02 &
  \multicolumn{1}{l|}{1.25E-02} &
  \multicolumn{1}{l|}{1.29E-02} &
  9.99E-02 &
  \multicolumn{1}{l|}{5.00E-03} &
  \multicolumn{1}{l|}{1.42E-14} &
  5.68E-02 &
  \multicolumn{1}{l|}{1.85E-02} &
  \multicolumn{1}{l|}{1.46E-02} &
  1.25E-01 \\ \hline
-0.75 &
  \multicolumn{1}{c|}{N/A} &
  \multicolumn{1}{c|}{N/A} &
  \multicolumn{1}{c|}{N/A} &
  \multicolumn{1}{c|}{N/A} &
  \multicolumn{1}{c|}{N/A} &
  \multicolumn{1}{c|}{N/A} &
  \multicolumn{1}{c|}{N/A} &
  \multicolumn{1}{c|}{N/A} &
  \multicolumn{1}{c|}{N/A} &
  \multicolumn{1}{c|}{N/A} &
  \multicolumn{1}{c|}{N/A} &
  \multicolumn{1}{c}{N/A} \\ \hline
\end{tabular}
\end{subtable}
\end{table}

\newpage

\subsubsection{Image Classification}
{\bf\noindent ResNet18 model on CIFAR10:}
\begin{table}[h]
\centering
\footnotesize
\caption{Evaluation result on CIFAR10 using ResNet18 : OOD detection performance with AUROC,AP and FPR; Mean over six random runs are reported(OE,EnergyOE,Ours).}
\begin{tabular}{c|c|c|c|c}
\hline
Dataset                    & Method          & AUC$\uparrow$            & AP$\uparrow$             & FPR$\downarrow$            \\ \hline
\multirow{3}{*}{Texture}   & OE (tune)       & 97.90          & 96.41      & 11.67          \\
                           & EnergyOE (tune) & 98.88          &  \textbf{98.13}          & 5.94           \\ \cline{2-5} 
                           &  \textbf{Ours}            & \textbf{98.90} &98.01 & \textbf{5.65}  \\ \hline
\multirow{3}{*}{SVHN}      & OE (tune)       & 98.67          & 99.44          & 7.10           \\
                           & EnergyOE (tune) & 98.88          & 99.57          & 6.51           \\ \cline{2-5} 
                           &  \textbf{Ours}            & \textbf{98.97} & \textbf{99.60} & \textbf{5.68}  \\ \hline
\multirow{3}{*}{CIFAR100}  & OE (tune)       & 92.07          & 91.39          & \textbf{31.05}          \\
                           & EnergyOE (tune) & \textbf{92.61}          & \textbf{92.76}          & 34.41          \\ \cline{2-5} 
                           &  \textbf{Ours}            & 92.54 & 92.64 & 34.27 \\ \hline
\multirow{3}{*}{\begin{tabular}[c]{@{}c@{}}Tiny\\ ImageNet\end{tabular}} & OE (tune) & 93.17 & 90.36 & 25.39 \\
                           & EnergyOE (tune) & \textbf{94.61}          & \textbf{92.76}          & \textbf{22.79}          \\ \cline{2-5} 
                           &  \textbf{Ours}            & 94.59 & 92.68 & 23.56 \\ \hline
\multirow{3}{*}{LSUN}      & OE (tune)       & 96.63          & 96.25          & 15.22          \\
                           & EnergyOE (tune) & 98.34          & 98.16          & 7.33           \\ \cline{2-5} 
                           & \textbf{Ours}           & \textbf{98.50} & \textbf{98.23} & \textbf{6.60}  \\ \hline
\multirow{3}{*}{Places365} & OE (tune)       & 95.63          & 98.31          & 18.80          \\
                           & EnergyOE (tune) & 97.31          & 98.98          & 11.97           \\ \cline{2-5} 
                           & \textbf{Ours}            & \textbf{97.47} & \textbf{99.01} & \textbf{11.28}  \\ \hline
\multirow{3}{*}{Average}   & OE (tune)       & 95.68          & 95.36          & 18.20          \\
                           & EnergyOE (tune) & 96.77          & \textbf{96.72}          & 14.82           \\ \cline{2-5} 
                           &  \textbf{Ours}            & \textbf{96.83} & 96.70 & \textbf{14.51}  \\ \hline
\end{tabular}
\end{table}

\begin{table}[h]
\centering
\footnotesize
\caption{Evaluation result on CIFAR10 using ResNet18 : OOD detection performance with AUROC,AP and FPR; Std over six random runs are reported(OE,EnergyOE,Ours).}
\begin{tabular}{c|c|l|l|l}
\hline
Dataset &
  Method &
  \multicolumn{1}{c|}{AUC$\uparrow$} &
  \multicolumn{1}{c|}{AP$\uparrow$} &
  \multicolumn{1}{c}{FPR$\downarrow$} \\ \hline
\multirow{3}{*}{Texture} &
  OE (tune) &
  \multicolumn{1}{c|}{1.42E-14} &
  \multicolumn{1}{c|}{1.42E-14} &
  \multicolumn{1}{c}{1.77E-02} \\
                           & EnergyOE (tune) & 4.71E-03 & 6.87E-03 & 4.08E-02 \\ \cline{2-5} 
                           & \textbf{Ours}            & 6.87E-03 & 1.00E-02 & 6.79E-02 \\ \hline
\multirow{3}{*}{SVHN}      & OE (tune)       & 0.00E+00 & 0.00E+00 & 3.73E-03 \\
                           & EnergyOE (tune) & 1.49E-02 & 5.00E-03 & 9.84E-02 \\ \cline{2-5} 
                           & \textbf{Ours}            & 2.38E-02 & 8.98E-03 & 1.69E-01 \\ \hline
\multirow{3}{*}{CIFAR100}  & OE (tune)       & 0.00E+00 & 0.00E+00 & 2.03E-02 \\
                           & EnergyOE (tune) & 8.98E-03 & 5.00E-03 & 1.23E-01 \\ \cline{2-5} 
                           & \textbf{Ours}            & 8.98E-03 & 9.57E-03 & 5.40E-02 \\ \hline
\multirow{3}{*}{\begin{tabular}[c]{@{}c@{}}Tiny\\ ImageNet\end{tabular}} &
  OE (tune) &
  0.00E+00 &
  0.00E+00 &
  7.64E-03 \\
                           & EnergyOE (tune) & 3.73E-03 & 3.73E-03 & 7.56E-02 \\ \cline{2-5} 
                           & \textbf{Ours}            & 6.87E-03 & 5.77E-03 & 7.65E-02 \\ \hline
\multirow{3}{*}{LSUN}      & OE (tune)       & 0.00E+00 & 0.00E+00 & 6.87E-03 \\
                           & EnergyOE (tune) & 7.45E-03 & 7.64E-03 & 4.28E-02 \\ \cline{2-5} 
                           & \textbf{Ours}            & 5.00E-03 & 4.71E-03 & 6.99E-02 \\ \hline
\multirow{3}{*}{Places365} & OE (tune)       & 0.00E+00 & 0.00E+00 & 7.45E-03 \\
                           & EnergyOE (tune) & 6.87E-03 & 0.00E+00 & 1.57E-02 \\ \cline{2-5} 
                           & \textbf{Ours}            & 3.73E-03 & 3.73E-03 & 4.82E-02 \\ \hline
\multirow{3}{*}{Average}   & OE (tune)       & 2.37E-15 & 8.33E-04 & 1.15E-02 \\
                           & EnergyOE (tune) & 3.44E-03 & 2.62E-03 & 3.21E-02 \\ \cline{2-5} 
                           & \textbf{Ours}            & 3.30E-03 & 3.52E-03 & 2.86E-02 \\ \hline
\end{tabular}
\end{table}
\newpage
{\bf\noindent ResNet18 model on CIFAR100:}
\begin{table}[h]
\centering
\footnotesize
\caption{Evaluation result on CIFAR100 using ResNet18 : OOD detection performance with AUROC,AP and FPR; Mean over six random runs are reported(OE,EnergyOE,Ours).}
\begin{tabular}{c|c|c|c|c}
\hline
Dataset                    & Method          & AUC$\uparrow$            & AP$\uparrow$             & FPR$\downarrow$            \\ \hline
\multirow{3}{*}{Texture}   & OE (tune)       & 83.69          & 71.62          & 50.41          \\
                           & EnergyOE (tune) & 89.03          & 80.83          & 38.02          \\ \cline{2-5} 
                           & \textbf{Ours}            & \textbf{89.11} & \textbf{80.94} & \textbf{37.78} \\ \hline
\multirow{3}{*}{SVHN}      & OE (tune)       & 85.15          & 91.45          & 45.70          \\
                           & EnergyOE (tune) & 93.14          & 96.19          & 23.22          \\ \cline{2-5} 
                           & \textbf{Ours}            & \textbf{93.41} & \textbf{96.33} & \textbf{22.48} \\ \hline
\multirow{3}{*}{CIFAR10}   & OE (tune)       & \textbf{75.52}          & \textbf{70.46}          & 63.52          \\
                           & EnergyOE (tune) & 74.96          & 69.20          & \textbf{62.83}          \\ \cline{2-5} 
                           &  \textbf{Ours}            & 74.70 & 68.74 & 63.43 \\ \hline
\multirow{3}{*}{\begin{tabular}[c]{@{}c@{}}Tiny\\ ImageNet\end{tabular}} & OE (tune) & 79.67 & 65.58 & \textbf{57.13} \\
                           & EnergyOE (tune) & \textbf{80.48}          & \textbf{67.74}          & 58.15          \\ \cline{2-5} 
                           &  \textbf{Ours}            & 80.36 & 67.56 & 58.61 \\ \hline
\multirow{3}{*}{LSUN}      & OE (tune)       & 86.11          & 75.38          & 46.25          \\
                           & EnergyOE (tune) & 88.26 & \textbf{77.72} & 38.34          \\ \cline{2-5} 
                           &  \textbf{Ours}            & \textbf{88.34}          & 77.68          & \textbf{37.89} \\ \hline
\multirow{3}{*}{Places365} & OE (tune)       & 86.41          & 93.11          & 47.31          \\
                           & EnergyOE (tune) & \textbf{89.19} & \textbf{94.27} & 37.56          \\ \cline{2-5} 
                           &  \textbf{Ours}            & 89.17          & 94.22          & \textbf{37.40} \\ \hline
\multirow{3}{*}{Average}   & OE (tune)       & 82.76          & 77.93          & 51.72          \\
                           & EnergyOE (tune) & 85.84          & \textbf{80.99}          & 43.02          \\ \cline{2-5} 
                           &  \textbf{Ours}            & \textbf{85.85} & 80.91 & \textbf{42.93} \\ \hline
\end{tabular}
\end{table}

\begin{table}[h]
\centering
\footnotesize
\caption{Evaluation result on CIFAR100 using ResNet18 : OOD detection performance with AUROC,AP and FPR; Std over six random runs are reported(OE,EnergyOE,Ours).}
\begin{tabular}{c|c|l|l|l}
\hline
Dataset                                                                  & Method    & \multicolumn{1}{c|}{AUC$\uparrow$} & \multicolumn{1}{c|}{AP$\uparrow$} & \multicolumn{1}{c}{FPR$\downarrow$} \\ \hline
\multirow{3}{*}{Texture}   & OE (tune)       & 0.00E+00 & 4.71E-03 & 9.57E-03 \\
                           & EnergyOE (tune) & 6.87E-03 & 1.49E-02 & 1.16E-01 \\ \cline{2-5} 
                           & \textbf{Ours}            & 1.21E-02 & 1.89E-02 & 1.29E-01 \\ \hline
\multirow{3}{*}{SVHN}      & OE (tune)       & 3.73E-03 & 0.00E+00 & 2.05E-02 \\
                           & EnergyOE (tune) & 8.98E-03 & 8.98E-03 & 3.13E-02 \\ \cline{2-5} 
                           & \textbf{Ours}           & 1.98E-02 & 1.46E-02 & 1.23E-01 \\ \hline
\multirow{3}{*}{CIFAR10}   & OE (tune)       & 3.73E-03 & 3.73E-03 & 1.34E-02 \\
                           & EnergyOE (tune) & 9.57E-03 & 1.07E-02 & 6.44E-02 \\ \cline{2-5} 
                           & \textbf{Ours}           & 7.64E-03 & 1.37E-02 & 6.50E-02 \\ \hline
\multirow{3}{*}{\begin{tabular}[c]{@{}c@{}}Tiny\\ ImageNet\end{tabular}} & OE (tune) & 0.00E+00                 & 4.71E-03                & 2.11E-02                \\
                           & EnergyOE (tune) & 5.77E-03 & 1.00E-02 & 4.85E-02 \\ \cline{2-5} 
                           & \textbf{Ours}           & 7.45E-03 & 9.57E-03 & 5.88E-02 \\ \hline
\multirow{3}{*}{LSUN}      & OE (tune)       & 0.00E+00 & 0.00E+00 & 1.83E-02 \\
                           & EnergyOE (tune) & 9.43E-03 & 1.25E-02  & 5.28E-02  \\ \cline{2-5} 
                           & \textbf{Ours}            & 8.98E-03  & 1.34E-02 & 6.08E-02  \\ \hline
\multirow{3}{*}{Places365} & OE (tune)       & 1.42E-14 & 0.00E+00 & 1.26E-02 \\
                           & EnergyOE (tune) & 5.00E-03 & 5.00E-03 & 1.67E-02 \\ \cline{2-5} 
                           & \textbf{Ours}            & 5.77E-03 &5.00E-03 &2.21E-02 \\ \hline
\multirow{3}{*}{Average}   & OE (tune)       & 9.62E-04 & 1.60E-03 & 6.14E-03 \\
                           & EnergyOE (tune) & 3.42E-03 & 5.56E-03 & 2.43E-02 \\ \cline{2-5} 
                           & \textbf{Ours}            & 4.04E-03 & 3.97E-03 & 1.41E-02 \\ \hline
\end{tabular}
\end{table}
\newpage

\subsection{Experiment Results on synthetic OOD dataset}
\subsubsection{Long-tailed Image Classification}
{\bf\noindent ResNet18 model on CIFAR10-LT:}
\begin{table}[h!]
\centering
\scriptsize
\captionsetup{font=footnotesize}
\caption{Synthetic OOD evaluation result on CIFAR10-LT using ResNet18 : OOD detection performance with AUROC,AP and FPR; (a): Mean over six random runs; (b): Std over six random runs(OE,OECC,EnergyOE,Ours).}
\begin{subtable}[c]{0.45\linewidth}
\caption{}
\begin{tabular}{c|c|c|c|c}
\hline

Dataset                    & Method          & AUC $\uparrow$            & AP $\uparrow$             & FPR $\downarrow$                \\ \hline
\multirow{4}{*}{Gaussian}   & OE (tune)      & 97.15          & 92.19     & 5.11       \\
 & OECC       & \textbf{99.80}           & \textbf{99.65}      & 0.50        \\
                           & EnergyOE (tune) & 99.74        & 99.33 & 0.54          \\ \cline{2-5} 
                           & \textbf{Ours}            & 99.76 & 99.16          & \textbf{0.49}  \\ \hline
\multirow{4}{*}{Rademacher}      & OE (tune)       &  \textbf{99.28}       & 98.24           & 1.57          \\
 & OECC       &  99.03       & \textbf{98.36}        & 2.51       \\
                           & EnergyOE (tune) & 99.13         & 97.16        & 1.51        \\ \cline{2-5} 
                           & \textbf{Ours}            & 99.00  & 96.26 & \textbf{1.42}  \\ \hline
\multirow{4}{*}{Blob}  & OE (tune)       & 43.80      & 43.12         & 84.15          \\
& OECC     & 59.18       & 52.12        & 73.19         \\
                           & EnergyOE (tune) & 90.16       & 85.39          & 28.79          \\ \cline{2-5} 
                           & \textbf{Ours}             & \textbf{93.18}& \textbf{89.34}  &\textbf{ 22.20 } \\ \hline
\multirow{4}{*}{Average}   & OE (tune)       & 80.08        & 77.85         & 30.27     \\
 & OECC       & 86.00         & 83.38         & 25.40           \\
                           & EnergyOE (tune) & 96.34         & 93.96 & 10.28          \\ \cline{2-5} 
                           & \textbf{Ours}             & \textbf{97.32} &\textbf{ 94.92 }         & \textbf{8.04} \\ \hline
\end{tabular}
\end{subtable}
\begin{subtable}[c]{0.45\linewidth}
\caption{}
\begin{tabular}{c|c|c|c|c}
\hline

Dataset                    & Method          & AUC $\uparrow$            & AP $\uparrow$             & FPR $\downarrow$                \\ \hline
\multirow{4}{*}{Gaussian}   & OE (tune)      & 1.46E-02          & 3.13E-02     & 5.25E-02       \\
 & OECC       & 3.73E-03           & 4.71E-03      & 5.00E-03        \\
                           & EnergyOE (tune) & 6.87E-03        & 3.86E-02 & 1.77E-02          \\ \cline{2-5} 
                           & \textbf{Ours}            & 7.45E-03 & 1.25E-02          & 3.73E-03  \\ \hline
\multirow{4}{*}{Rademacher}      & OE (tune)       &  3.73E-03       & 1.29E-02           & 1.37E-02          \\
 & OECC       &  3.73E-03       & 1.34E-02        & 1.25E-02       \\
                           & EnergyOE (tune) & 1.71E-02         & 4.73E-02        & 2.54E-02        \\ \cline{2-5} 
                           & \textbf{Ours}            & 1.86E-02  & 6.37E-02 & 3.54E-02  \\ \hline
\multirow{4}{*}{Blob}  & OE (tune)       & 1.16E-01      & 7.42E-02         & 1.88E-01          \\
& OECC     & 2.70E-01       & 1.60E-01        & 3.05E-01         \\
                           & EnergyOE (tune) & 6.57E-02       & 9.63E-02          & 3.64E-01          \\ \cline{2-5} 
                           & \textbf{Ours}             & 1.14E-01 & 1.63E-01  & 3.56E-01  \\ \hline
\multirow{4}{*}{Average}   & OE (tune)       & 3.62E-02       & 2.10E-02         & 7.26E-02     \\
 & OECC       & 9.00E-02        & 5.68E-02         & 1.02E-01           \\
                           & EnergyOE (tune) & 2.62E-02         & 5.10E-02 & 1.20E-01          \\ \cline{2-5} 
                           & \textbf{Ours}             & 3.65E-02 & 5.26E-02          & 1.26E-01 \\ \hline
\end{tabular}
\end{subtable}

\end{table}

{\bf\noindent ResNet18 model on CIFAR100-LT:}
\begin{table}[h!]
\centering
\scriptsize
\captionsetup{font=footnotesize}
\caption{Synthetic OOD evaluation result on CIFAR100-LT using ResNet18 : OOD detection performance with AUROC,AP and FPR; (a): Mean over six random runs; (b): Std over six random runs(OE,OECC,EnergyOE,Ours).}
\begin{subtable}[c]{0.45\linewidth}
\caption{}
\begin{tabular}{c|c|c|c|c}
\hline

Dataset                    & Method          & AUC $\uparrow$            & AP $\uparrow$             & FPR $\downarrow$                \\ \hline
\multirow{4}{*}{Gaussian}   & OE (tune)      & 56.27          & 48.35      & 69.79       \\
 & OECC       & \textbf{95.59}          & 87.82      & \textbf{6.58}         \\
                           & EnergyOE (tune) & 88.13         & 79.36 & 27.89          \\ \cline{2-5} 
                           & \textbf{Ours}            & 94.56  & \textbf{88.32}         & 13.22  \\ \hline
\multirow{4}{*}{Rademacher}      & OE (tune)       &  48.47       & 44.24          & 75.03         \\
 & OECC       & \textbf{92.11}       & 80.92        & \textbf{11.78}       \\
                           & EnergyOE (tune) & 83.43         & 71.56       & 32.76         \\ \cline{2-5} 
                           & \textbf{Ours}            & 91.48  & \textbf{82.87} & 18.42  \\ \hline
\multirow{4}{*}{Blob}  & OE (tune)       & 70.43      & 66.45         & 73.39          \\
& OECC     & 95.42       & 90.91        & \textbf{12.25}         \\
                           & EnergyOE (tune) & 88.48       & 85.37          & 36.28         \\ \cline{2-5} 
                           & \textbf{Ours}             & \textbf{96.42} & \textbf{94.92}  & 13.30  \\ \hline
\multirow{4}{*}{Average}   & OE (tune)       & 58.39         & 53.01          & 72.74      \\
 & OECC       & \textbf{94.37}         & 86.55         & \textbf{10.20}           \\
                           & EnergyOE (tune) & 86.68         & 78.76 & 32.31         \\ \cline{2-5} 
                           & \textbf{Ours}             & 94.15 &  \textbf{88.70}         & 14.98 \\ \hline
\end{tabular}
\end{subtable}
\begin{subtable}[c]{0.45\linewidth}
\caption{}
\begin{tabular}{c|c|c|c|c}
\hline

Dataset                    & Method          & AUC $\uparrow$            & AP $\uparrow$             & FPR $\downarrow$                \\ \hline
\multirow{4}{*}{Gaussian}   & OE (tune)      & 2.05E-01          & 1.22E-01      & 2.16E-01       \\
 & OECC       & 4.34E-02          & 1.16E-01      & 3.53E-02         \\
                           & EnergyOE (tune) & 1.59E-01         & 2.77E-01 & 2.54E-01          \\ \cline{2-5} 
                           & \textbf{Ours}            & 1.10E-01  & 1.91E-01         & 2.09E-01  \\ \hline
\multirow{4}{*}{Rademacher}      & OE (tune)       &  1.28E-01       & 6.03E-02          & 1.72E-01         \\
 & OECC       & 2.62E-02      & 5.16E-02        & 5.21E-02       \\
                           & EnergyOE (tune) & 2.05E-01        & 2.77E-01       & 3.80E-01         \\ \cline{2-5} 
                           & \textbf{Ours}            & 1.67E-01  & 2.48E-01 & 3.22E-01  \\ \hline
\multirow{4}{*}{Blob}  & OE (tune)       & 9.75E-02      & 1.26E-01         & 2.06E-01          \\
& OECC     & 2.31E-02       & 7.13E-02        & 1.53E-01         \\
                           & EnergyOE (tune) & 8.50E-02       & 8.30E-02          & 4.39E-01         \\ \cline{2-5} 
                           & \textbf{Ours}             & 5.88E-02 & 7.65E-02  & 3.79E-01  \\ \hline
\multirow{4}{*}{Average}   & OE (tune)       & 7.97E-02         & 6.57E-02          & 1.16E-01      \\
 & OECC       & 2.17E-02         & 5.23E-02         & 6.18E-02           \\
                           & EnergyOE (tune) & 1.02E-01         & 1.68E-01 & 6.82E-02         \\ \cline{2-5} 
                           & \textbf{Ours}             & 9.25E-02 &  1.43E-01         & 1.39E-01 \\ \hline
\end{tabular}
\end{subtable}
\end{table}

\newpage

\subsubsection{Image Classification}
{\bf\noindent ResNet18 model on CIFAR10:}
\begin{table}[h!]
\centering
\scriptsize
\captionsetup{font=footnotesize}
\caption{Synthetic OOD evaluation result on CIFAR10 using ResNet18 : OOD detection performance with AUROC,AP and FPR; (a): Mean over six random runs; (b): Std over six random runs(OE,OECC,EnergyOE,Ours).}
\begin{subtable}[c]{0.45\linewidth}
\caption{}
\begin{tabular}{c|c|c|c|c}
\hline

Dataset                    & Method          & AUC $\uparrow$            & AP $\uparrow$             & FPR $\downarrow$                \\ \hline
\multirow{4}{*}{Gaussian}   & OE (tune)      & 99.74          & \textbf{99.28}      & 0.57       \\
 & OECC       & 99.43           & 98.20      & 0.80         \\
                           & EnergyOE (tune) & 99.76         & 99.19 & 0.49          \\ \cline{2-5} 
                           & \textbf{Ours}            & \textbf{99.82}  & 99.20          & \textbf{0.28}  \\ \hline
\multirow{4}{*}{Rademacher}      & OE (tune)       &  99.63       & 98.91           & 0.80          \\
 & OECC       &  99.56       & 98.47        & 0.57       \\
                           & EnergyOE (tune) & 99.59         & 98.57        & 0.72         \\ \cline{2-5} 
                           & \textbf{Ours}            & \textbf{99.78}  & \textbf{98.93} & \textbf{0.33}  \\ \hline
\multirow{4}{*}{Blob}  & OE (tune)       & 97.71       & 96.71         & 8.84          \\
& OECC     & 99.40       & 98.69        & 1.50         \\
                           & EnergyOE (tune) & 99.42       & 99.24          & 2.10          \\ \cline{2-5} 
                           & \textbf{Ours}             & \textbf{99.63}& \textbf{99.47}  & \textbf{1.44}  \\ \hline
\multirow{4}{*}{Average}   & OE (tune)       & 99.03         & 98.30          & 3.41      \\
 & OECC       & 99.46         & 98.45         & 0.96           \\
                           & EnergyOE (tune) & 99.59         & 99.00 & 1.10          \\ \cline{2-5} 
                           & \textbf{Ours}             & \textbf{99.74}& \textbf{99.20}          & \textbf{0.69} \\ \hline
\end{tabular}
\end{subtable}
\begin{subtable}[c]{0.45\linewidth}
\caption{}
\begin{tabular}{c|c|c|c|c}
\hline

Dataset                    & Method          & AUC $\uparrow$            & AP $\uparrow$             & FPR $\downarrow$                \\ \hline
\multirow{4}{*}{Gaussian}   & OE (tune)      & 0.00E+00          & 4.71E-03     & 0.00E+00       \\
 & OECC       & 2.40E-01           & 1.02E+00      & 2.33E-01        \\
                           & EnergyOE (tune) & 5.77E-03         & 1.77E-02 & 9.57E-03          \\ \cline{2-5} 
                           & \textbf{Ours}            & 3.73E-03  & 1.80E-02          & 7.45E-03  \\ \hline
\multirow{4}{*}{Rademacher}      & OE (tune)       &  4.71E-03       & 7.45E-03           & 1.21E-02          \\
 & OECC       &  2.79E-01       & 1.17E+00        & 2.90E-01       \\
                           & EnergyOE (tune) & 1.61E-02         & 6.67E-02        & 3.16E-02         \\ \cline{2-5} 
                           & \textbf{Ours}            & 4.71E-03  & 1.70E-02 & 8.16E-03   \\ \hline
\multirow{4}{*}{Blob}  & OE (tune)       & 2.75E-02       & 3.04E-02         & 2.49E-01          \\
& OECC     & 1.11E-01       & 4.37E-01        & 8.98E-02         \\
                           & EnergyOE (tune) & 1.57E-02       & 1.57E-02          & 6.47E-02          \\ \cline{2-5} 
                           & \textbf{Ours}             & 1.80E-02 & 1.73E-02  & 5.30E-02  \\ \hline
\multirow{4}{*}{Average}   & OE (tune)       & 8.61E-03         & 9.57E-03          & 8.08E-02      \\
 & OECC       & 1.95E-01         & 8.50E-01         & 1.96E-01           \\
                           & EnergyOE (tune) & 1.10E-02         & 3.07E-02 & 2.94E-02          \\ \cline{2-5} 
                           & \textbf{Ours}             & 6.31E-03& 1.23E-02          & 1.91E-02 \\ \hline
\end{tabular}
\end{subtable}

\end{table}

{\bf\noindent ResNet18 model on CIFAR100:}
\begin{table}[h!]
\centering
\scriptsize
\captionsetup{font=footnotesize}
\caption{Synthetic OOD evaluation result on CIFAR100 using ResNet18 : OOD detection performance with AUROC,AP and FPR; (a): Mean over six random runs; (b): Std over six random runs(OE,OECC,EnergyOE,Ours).}
\begin{subtable}[c]{0.45\linewidth}
\caption{}
\begin{tabular}{c|c|c|c|c}
\hline

Dataset                    & Method          & AUC $\uparrow$            & AP $\uparrow$             & FPR $\downarrow$                \\ \hline
\multirow{4}{*}{Gaussian}   & OE (tune)      & 81.41          & 66.42      & 25.99       \\
 & OECC       &\textbf{ 99.31 }          &\textbf{ 98.59}      &\textbf{ 1.82}         \\
                           & EnergyOE (tune) & 92.66         & 83.22 & 12.77          \\ \cline{2-5} 
                           & \textbf{Ours}            & 93.92  & 85.64          & 11.01  \\ \hline
\multirow{4}{*}{Rademacher}      & OE (tune)       &  99.87       & 99.72           & 0.33          \\
 & OECC       &  96.44       & 89.06        & 4.58       \\
                           & EnergyOE (tune) & \textbf{99.91}         &\textbf{ 99.83 }       & \textbf{0.21}         \\ \cline{2-5} 
                           & \textbf{Ours}            & \textbf{99.91}  & \textbf{99.83 } & \textbf{0.21}  \\ \hline
\multirow{4}{*}{Blob}  & OE (tune)       & 98.31       & 97.81         & 6.52          \\
& OECC     & 94.26       & 89.72        & 14.67         \\
                           & EnergyOE (tune) & 98.56       & 98.07          & 6.01          \\ \cline{2-5} 
                           & \textbf{Ours}             &\textbf{ 98.75}  & \textbf{98.34}  &\textbf{ 5.24 } \\ \hline
\multirow{4}{*}{Average}   & OE (tune)       & 93.20         & 87.98          & 10.94      \\
 & OECC       & 96.67         & 92.46         & 7.02           \\
                           & EnergyOE (tune) & 97.04         & 93.71 & 6.33          \\ \cline{2-5} 
                           & \textbf{Ours}             & \textbf{ 97.53} & \textbf{94.60}          & \textbf{5.49} \\ \hline
\end{tabular}
\end{subtable}
\begin{subtable}[c]{0.45\linewidth}
\caption{}
\begin{tabular}{c|c|c|c|c}
\hline

Dataset                    & Method          & AUC $\uparrow$            & AP $\uparrow$             & FPR $\downarrow$                \\ \hline
\multirow{4}{*}{Gaussian}   & OE (tune)      & 3.27E-02          & 3.44E-02      & 1.18E-01       \\
 & OECC       & 5.06E-02           & 1.46E-01      & 1.09E-01         \\
                           & EnergyOE (tune) & 1.17E-01         & 2.20E-01 & 1.27E-01          \\ \cline{2-5} 
                           & \textbf{Ours}            & 7.08E-02  & 1.42E-01          & 1.08E-01  \\ \hline
\multirow{4}{*}{Rademacher}      & OE (tune)       &  0.00E+00       & 3.73E-03           & 4.71E-03          \\
 & OECC       &  1.35E-01       & 3.89E-01        & 1.14E-01       \\
                           & EnergyOE (tune) & 3.73E-03        & 9.43E-03        & 6.87E-03         \\ \cline{2-5} 
                           & \textbf{Ours}            & 4.71E-03  & 6.87E-03  & 7.64E-03  \\ \hline
\multirow{4}{*}{Blob}  & OE (tune)       & 2.92E-02      & 3.98E-02         & 1.55E-01          \\
& OECC     & 1.99E-01       & 3.20E-01        & 6.57E-01         \\
                           & EnergyOE (tune) & 3.40E-02       & 3.98E-02          & 1.78E-01          \\ \cline{2-5} 
                           & \textbf{Ours}             & 1.57E-02  & 2.27E-02  & 8.03E-02  \\ \hline
\multirow{4}{*}{Average}   & OE (tune)       & 1.56E-02         & 2.03E-02          & 6.29E-02      \\
 & OECC       & 8.80E-02         & 1.79E-01         & 2.49E-01           \\
                           & EnergyOE (tune) & 4.55E-02         & 8.38E-02 & 8.44E-02          \\ \cline{2-5} 
                           & \textbf{Ours}             &  2.22E-02 & 4.53E-02          & 4.59E-02 \\ \hline
\end{tabular}
\end{subtable}

\end{table}


\end{document}